%% file: main_iclr.tex
\documentclass{article}
\usepackage{iclr2027_conference,times}
\usepackage[top=2.25cm,bottom=2.5cm,left=2.5cm,right=2.5cm,headsep=30pt]{geometry}
\input{math_commands.tex}

\usepackage{hyperref}
\usepackage{url}
\hypersetup{colorlinks=true,citecolor={blue!55!black},linkcolor=black,urlcolor={blue!55!black}}

\usepackage{amsmath}
\usepackage{amssymb}
\usepackage{graphicx}
\usepackage{caption}
\usepackage{booktabs}
\usepackage{multirow}
\usepackage{tabularx}
\usepackage{siunitx}
\usepackage{xcolor}
\usepackage{colortbl}  %
\usepackage{enumitem}  %
\usepackage{float}  %
\usepackage{placeins}  %
\usepackage{wrapfig}  %
\definecolor{arbRow}{HTML}{FF9900}
\definecolor{hilite}{RGB}{255,241,153}  %
\usepackage{tikz}
\usepackage{fontawesome5}  %
\usepackage{pifont}  %
\usepackage{tcolorbox}  %
\tcbuselibrary{skins,breakable}
\usepackage{pgfplots}
\pgfplotsset{compat=1.17}
\usepgfplotslibrary{groupplots}
\usetikzlibrary{positioning,arrows.meta,shapes.misc,shapes.geometric,fit,backgrounds,calc,decorations.pathreplacing}

\definecolor{promptframe}{RGB}{255,153,0}
\definecolor{promptback}{RGB}{255,247,235}
\newtcolorbox{promptbox}[1]{%
  breakable, enhanced, sharp corners=downhill, arc=2pt,
  colback=promptback, colframe=promptframe, boxrule=0.6pt,
  left=8pt, right=8pt, top=6pt, bottom=6pt,
  fonttitle=\bfseries\small, coltitle=white,
  colbacktitle=promptframe, attach boxed title to top left={xshift=6pt, yshift=-2pt},
  boxed title style={sharp corners=downhill, arc=1pt, boxrule=0pt},
  title={#1}}

\usepackage{listings}
\tcbuselibrary{listings,breakable,skins}
\definecolor{codeback}{HTML}{FAFAFA}
\definecolor{codeframe}{HTML}{C9C9C9}
\newtcblisting{codebox}{
  listing only, breakable, enhanced, sharp corners=downhill, arc=2pt,
  colback=codeback, colframe=codeframe, boxrule=0.5pt,
  left=7pt, right=7pt, top=5pt, bottom=5pt,
  listing options={
    basicstyle=\ttfamily\scriptsize,
    commentstyle=\color{black!45}\itshape,
    columns=fullflexible, keepspaces=true,
    breaklines=true, breakatwhitespace=false,
    showstringspaces=false, tabsize=2, numbers=none,
    language=C++,
  }}

\newcommand{\logo}[2][1em]{\raisebox{-0.18em}{\includegraphics[height=#1]{figures/logos/#2}}}

\newcommand{\hi}[1]{#1}
\providecommand{\tb}[1]{\mbox{\textbf{#1}}}
\providecommand{\su}[1]{\mbox{\underline{#1}}}

\title{Beyond the Verdict: Agentic Code Judging\\with Verifiable Feedback}

\author{{\small\bfseries Zheye Deng\textsuperscript{1,2}\enspace Sha Li\textsuperscript{2}\enspace Changlong Yu\textsuperscript{2}\enspace Xin Liu\textsuperscript{2}\enspace Qin Lu\textsuperscript{2}\enspace Priyanka Nigam\textsuperscript{2}\enspace Yangqiu Song\textsuperscript{1}}\\[3pt]
\normalfont\small \textsuperscript{1}HKUST\enspace \textsuperscript{2}Amazon\\[2pt]
\normalfont\small\texttt{zdengah@cse.ust.hk, \{slliz, changlyu\}@amazon.com}\\[2pt]
\normalfont\footnotesize\faGlobe\enspace Project Page: \url{https://horizon-llm.github.io/IOI2026/}\\[1pt]
\normalfont\footnotesize\faGithub\enspace Code \& Artifacts: \url{https://github.com/horizon-llm/IOI2026}}

\iclrfinalcopy

\makeatletter
\renewcommand{\@maketitle}{%
  \vbox{\hsize\textwidth
    {\centering\LARGE\sc\@title\par}%
    \lhead{Preprint}%
    \noindent\makebox[\hsize][c]{%
      \begin{tabular}[t]{c}
        \bf\rule{\z@}{24pt}\@author
      \end{tabular}}%
    \vskip 0.3in minus 0.1in}}
\makeatother

\begin{document}

\maketitle
\lhead{Preprint}

\begin{abstract}
Test-time scaling for large language models (LLMs) can involve generating multiple candidate solutions for selection or revising an earlier attempt.
To guide selection and revision, we need a verifier that can estimate candidate correctness and provide evidence of failure. In competitive programming, however, official judging systems provide authoritative verdicts under finite submission budgets, without revealing failure explanations or failing inputs.
To provide informative feedback without consuming official submissions, we formulate a task in which a model judges and diagnoses a candidate program using only the problem statement and source code, with generated counterexamples serving as checkable evidence.
To evaluate this task, we build \textsc{USACO-Judge} from the USACO 2026 contests, using accepted and unaccepted
programs written by human experts and multiple LLMs. We then train \textsc{Arbiter}, a 35B mixture-of-experts (MoE) agentic verifier, with reinforcement learning in a generic execution sandbox.
Training raises counterexample success on incorrect programs from 40.61\% to 62.67\% and mean binary verdict accuracy on \textsc{USACO-Judge} from 71.41\% to 87.41\%, approaching frontier verification performance.
Beyond verification, \textsc{Arbiter} outperforms baselines in both code selection and refinement. In an IOI~2026 case study,
\textsc{Arbiter}-guided refinement raises \mbox{GPT-OSS-120B}'s expected score above the \mbox{gold-medal} threshold under the official submission limit.
\end{abstract}

\input{sections/body/introduction}
\input{sections/body/related_work}
\section{Verification Task}
\label{sec:task-env}

\paragraph{A running example.}
Figure~\ref{fig:task-example} shows an incorrect candidate and its counterexample.

\begin{figure}[H]
  \centering
  \begin{tikzpicture}[
    font=\footnotesize,
    node distance=5mm,
    flowbox/.style={
      draw=blue!45!black, fill=blue!3, rounded corners=2pt,
      text width=0.91\linewidth, align=left, inner sep=7pt
    },
    checkbox/.style={
      draw=black!45, fill=black!3, rounded corners=2pt,
      text width=0.91\linewidth, align=left, inner sep=7pt
    },
    flowarrow/.style={-{Latex[length=2mm]}, thick, draw=black!65}
  ]
    \node[flowbox] (given) {%
      {\centering\small\bfseries\color{blue!60!black}
      Agentic Judge receives $(\mathcal{P},\mathcal{C})$ and can run Python/Bash\par}
      \smallskip{\color{blue!25}\hrule height 0.4pt}\smallskip
      \begin{tabularx}{\linewidth}{@{}>{\bfseries\raggedright\arraybackslash}p{0.18\linewidth}@{\hspace{0.5em}}>{\raggedright\arraybackslash}X@{}}
        Problem $(\mathcal{P})$: & read $n$ integers $a_1,a_2,\ldots,a_n$ and print their maximum
        ($1\leq n\leq1000$, $-100\leq a_i\leq100$). \\[2pt]
        Candidate $(\mathcal{C})$: & \texttt{ans = 0;\quad for each a[i]: ans = max(ans, a[i]);}
      \end{tabularx}
    };
    \node[flowbox, below=of given] (answer) {%
      {\centering\small\bfseries\color{blue!60!black}
      Agentic Judge calls \textit{submit}$(v,d,g)$\par}
      \smallskip{\color{blue!25}\hrule height 0.4pt}\smallskip
      \begin{tabularx}{\linewidth}{@{}>{\bfseries\raggedright\arraybackslash}p{0.18\linewidth}@{\hspace{0.5em}}>{\raggedright\arraybackslash}X@{}}
        Verdict $(v)$: & \textit{Wrong Answer}. \\[2pt]
        Diagnosis $(d)$: & initialization to zero fails on all-negative inputs. \\[2pt]
        Generator $(g)$: & \texttt{print("3\textbackslash n-7 -2 -9")}
      \end{tabularx}
    };
    \node[checkbox, below=of answer] (judge) {%
      {\centering\small\bfseries\color{blue!60!black}
      External judge evaluates the submission\par}
      \smallskip{\color{black!20}\hrule height 0.4pt}\smallskip
      \begin{tabularx}{\linewidth}{@{}>{\bfseries\raggedright\arraybackslash}p{0.22\linewidth}@{\hspace{0.7em}}>{\raggedright\arraybackslash}X@{}}
        \ding{202}\enspace Official tests: & run $\mathcal{C}$ to obtain $v^\star$. \\
        \ding{203}\enspace Generate input: & $x=g()$ gives $n=3$, $(-7,-2,-9)$. \\
        \ding{204}\enspace Validate input: & $x$ is \textit{legal}. \\
        \ding{205}\enspace Execute: & $\mathcal{C}(x)=0$; reference output $=-2$. \\
        \ding{206}\enspace Checker: & \textit{mismatch} $\Rightarrow$ verified failure.
      \end{tabularx}
    };
    \draw[flowarrow] (given.south) -- (answer.north);
    \draw[flowarrow] (answer.south) -- (judge.north);
  \end{tikzpicture}
  \caption{A verification instance. The verifier's answer and the judge-side
  evaluation are shown in order; judge-only results are not
  revealed during the verifier's episode.}
  \label{fig:task-example}
\end{figure}

\paragraph{Verifier Interface.}
\label{sec:env}
Given a problem $\mathcal{P}$ and candidate program $\mathcal{C}$, the verifier
can run Python code and Bash commands in a generic sandbox to construct and run
its own tests. It cannot access the official tests, reference solution, input
validator, output checker, or the result of
checking its final counterexample. To end the episode, it calls
\textit{submit}$(v,d,g)$, where $v$ is a verdict, $d$ a natural-language
diagnosis, and $g$ a deterministic Python 3 generator that prints one input.
This call does not submit $\mathcal{C}$ to the official judge. Possible verdicts are
\textit{Accepted}, \textit{Wrong Answer}, \textit{Time Limit Exceeded},
\textit{Runtime Error}, \textit{Memory Limit Exceeded}, and \textit{Compile
Error}. The main decision is whether to accept $\mathcal{C}$. The other verdicts
specify a suspected failure type. The diagnosis identifies the suspected
failure for an unaccepted program. The generator $g$ is empty for
\textit{Accepted} and \textit{Compile Error}.

\paragraph{External Judge Evaluation.}
After \textit{submit}, an external judge runs $\mathcal{C}$ on official tests to
obtain the reference verdict $v^\star$. If $g$ is supplied, it generates an
input $x$, validates $x$, and tests $\mathcal{C}$ against a reference solution
under the task's judging rules. A legal $x$ that breaks $\mathcal{C}$
establishes a failure even when official tests miss it; failing to find one
does not establish acceptance. The private checks supply training rewards and
evaluation labels but are withheld from the verifier. We do not score $d$
directly; counterexample success is a checkable proxy for diagnosis quality,
not proof of every claim in $d$. Appendix~\ref{app:infra} specifies the full
protocol.

\input{sections/body/training}

\input{sections/body/benchmarks}

\input{sections/body/experiments}
\input{sections/body/conclusion}

\clearpage
\bibliography{iclr2027_conference}
\bibliographystyle{iclr2027_conference}

\appendix

\section{Evaluation Infrastructure and Protocols}
\label{app:infra}

\paragraph{Verifier interface and judge-side evaluation.}
The verifier can use \textit{execute\_code} to run Python and
\textit{execute\_command} to run shell commands in a generic sandbox. It may
construct and run its own tests, but cannot access the official test data,
reference solution, input validator, output checker, or the result of checking
its submitted counterexample. A final \textit{submit} call commits $(v,d,g)$.
We use an online judge based on the open-source Universal Online Judge
(UOJ)\footnote{\url{https://github.com/UniversalOJ/UOJ-System}}. The judge
compiles and runs $\mathcal{C}$ on the official hidden tests under the problem's
resource limits to obtain $v^\star$. Verdict accuracy compares $v$ and
$v^\star$ after mapping both to \textit{Accepted} or \textit{unaccepted}.
Separately, when $g$ is supplied, the judge runs it to obtain $x$ and checks
$x$ with the problem's input validator. For a legal $x$, it runs $\mathcal{C}$
and a reference solution; the output checker compares normally terminating
outputs, while compilation, runtime, time-limit, and memory-limit failures are
recorded directly. The counterexample succeeds only when $x$ is legal and
$\mathcal{C}$ fails on it. These checks supply training labels and rewards and
evaluate the benchmarks, without becoming observations available to the
verifier during its episode.

\paragraph{Hardware and deployment.} Our UOJ deployment runs on one server
with a 32-core / 64-thread Intel Xeon Platinum 8488C and 128 GB of RAM. The
server runs Amazon Linux~2023 with Linux~6.12. The web service and database use
separate Docker containers, alongside a pool of up to 30 judger containers. The
pool processes submissions and counterexamples in parallel. Each candidate runs
in an isolated sandbox under the time and memory limits specified for the
problem.

\paragraph{Language and compilation.} Candidate solutions, reference solutions,
validators, and checkers use C++ unless the source benchmark requires otherwise.
We compile submissions with \textbf{C++17} and \texttt{-O2}, matching the source
contests. A submission that fails under this configuration receives
\emph{Compile Error}. Counterexample generators may use Python. They produce
input files and are not judged as contest solutions.

\section{Data and Benchmark Construction}
\label{app:benchmark-construction}

\subsection{Training Data Composition}
\label{app:traindata}

Table~\ref{tab:traindata} groups the training pool by sampling score band and
judge verdict. Score bands set the sampling quotas. Training labels come from the online judge: \emph{Accepted} (AC), \emph{Wrong
Answer} (WA), \emph{Time Limit Exceeded} (TLE), \emph{Memory Limit
Exceeded} (MLE), \emph{Runtime Error} (RE), and \emph{Compile Error} (CE). The 76--99 near-miss band contains WA, TLE, MLE, and RE
programs. A time-limit failure in that band is labeled TLE.

\begin{table}[h]
\centering
\newcommand{\trainDash}{--}
\newcommand{\trainpc}[1]{{\scriptsize #1\%}}
\small
\begin{tabular*}{\linewidth}{@{\extracolsep{\fill}}l r r r r r r r r@{}}
\toprule
& \multicolumn{6}{c}{\textbf{Judge verdict}} & \multicolumn{2}{c}{} \\
\cmidrule(r){2-7}
\textbf{Score band} & AC & WA & TLE & MLE & RE & CE & \multicolumn{2}{c}{\textbf{Total}} \\
\midrule
100 (accepted)       & 3{,}000 & \trainDash   & \trainDash   & \trainDash & \trainDash & \trainDash & 3{,}000 & \trainpc{15.0} \\
76--99 (near-miss) & \trainDash   & 4{,}460 & 3{,}107 & 220   & 163   & \trainDash & \textbf{7{,}950} & \trainpc{39.8} \\
1--75 (partial)    & \trainDash   & 4{,}600 & 1{,}600 & 400   & 600   & \trainDash & 7{,}200 & \trainpc{36.0} \\
0                    & \trainDash   & 1{,}000 & \trainDash   & \trainDash & \trainDash & \trainDash & 1{,}000 & \trainpc{5.0} \\
Compile Error        & \trainDash   & \trainDash   & \trainDash   & \trainDash & \trainDash & 850   & 850 & \trainpc{4.3} \\
\midrule
\textbf{Total} & 3{,}000 & 10{,}060 & 4{,}707 & 620 & 763 & 850 & \textbf{20{,}000} & \\
& \trainpc{15.0} & \trainpc{50.3} & \trainpc{23.5} & \trainpc{3.1} & \trainpc{3.8} & \trainpc{4.3} & & \\
\bottomrule
\end{tabular*}
\caption{Composition of the training pool by sampling score band and judge verdict. Margins report the share of the pool for each row and column.}
\label{tab:traindata}
\end{table}

\subsection{\textsc{UOJ-Bench} Re-hosting}
\label{app:uojbench}

\textsc{UOJ-Bench}~\citep{uojbench2026} divides its hacking data into Easy
submissions, which score at least 60 but fail the standard tests, and Hard
submissions, which pass those tests before a successful community hack exposes
the bug. We use every released candidate from both splits for evaluation.
Neither split enters training.

We run candidate scoring and hack evaluation on UOJ, using the same
infrastructure as the original benchmark. Each target problem is re-hosted with
its validator, reference solution, and checker. Replaying the released
candidates reproduces the outcomes reported in the original paper. For a
generated hack, UOJ first checks the input with the problem validator. If the
input is valid, UOJ runs the reference and candidate solutions and passes both
outputs to the checker. The hack succeeds when the candidate fails during
execution or the checker rejects its output.

\subsection{\textsc{USACO-Judge} Validators and Custom Checkers}
\label{app:validators-checkers}

USACO supplies the problem statements, test data, and reference solutions used by
our UOJ deployment, but its release omits the validators needed for arbitrary
model-generated inputs. We wrote a C++ validator for each of the 39 problems
using \texttt{testlib}, an open source C++ library for writing contest input validators and output
checkers.\footnote{\url{https://github.com/MikeMirzayanov/testlib}}
Each proposed hack must pass this validator before UOJ runs the reference and
candidate solutions.

For 33 problems, the expected output is a unique sequence of tokens, which UOJ
checks with a built-in \texttt{ncmp} or \texttt{wcmp} comparator. Six require
custom checking. They accept multiple correct outputs or ask for a witness, so
we wrote a custom output checker for each one
(Table~\ref{tab:custom-checkers}).

\begin{table}[h]
\centering
\renewcommand{\arraystretch}{1.15}
\setlength{\tabcolsep}{5pt}
\resizebox{\linewidth}{!}{%
\begin{tabular}{@{}r >{\raggedright\arraybackslash}p{17.0cm}@{}}
\toprule
\textbf{CPID} & \textbf{What the checker validates} \tabularnewline
\midrule
\href{https://usaco.org/index.php?page=viewproblem2\&cpid=1540}{1540}
& Verifies the reported optimum and checks that the supplied position assignment reconstructs valid square subsequences. \tabularnewline
\href{https://usaco.org/index.php?page=viewproblem2\&cpid=1563}{1563}
& Simulates the submitted keystrokes and checks that they reconstruct the target string. \tabularnewline
\href{https://usaco.org/index.php?page=viewproblem2\&cpid=1566}{1566}
& Tests the submitted permutation and truth-teller/liar assignment against every directional claim. \tabularnewline
\href{https://usaco.org/index.php?page=viewproblem2\&cpid=1589}{1589}
& Replays the submitted swaps, enforces the operation limit, and checks the final string. \tabularnewline
\href{https://usaco.org/index.php?page=viewproblem2\&cpid=1597}{1597}
& Confirms the minimum attack count, replays the submitted attacks, and verifies that every target reaches zero HP. \tabularnewline
\href{https://usaco.org/index.php?page=viewproblem2\&cpid=1598}{1598}
& Validates infeasibility claims or tests the submitted array against every min, max, sum, and xor query. \tabularnewline
\bottomrule
\end{tabular}%
}
\caption{Custom output checkers in \textsc{USACO-Judge}, indexed by official USACO
\textbf{CPID}. The other 33 problems use UOJ built-in token comparators.}
\label{tab:custom-checkers}
\end{table}

\subsection{\textsc{USACO-Judge} Construction Pipeline}
\label{app:usaco-judge-construction}

We build \textsc{USACO-Judge} problem by problem from C++17 programs
generated by five LLMs and written by five human competitive programmers. We assign each
problem a quota split by score and verdict, remove duplicates, and select
diverse programs within each pool. Every retained program is rejudged on UOJ
before entering the 900-candidate benchmark.

\paragraph{Per-problem quotas.}
The 21 hard problems from the Gold, Platinum, and Open divisions each contribute
10 accepted and 20 unaccepted candidates, giving 630 programs. The 18 easy
problems from Bronze and Silver each contribute 5 accepted and 10 unaccepted
candidates, giving 270. Both groups have a 1:2 ratio of accepted to unaccepted
programs.

For every hard problem, the 20 unaccepted candidates are divided evenly across
score bands 1--19, 20--39, 40--59, 60--79, and 80--99. The easy group contains
36, 36, 36, 51, and 21 candidates in these bands. Bronze and Silver have few
high-scoring unaccepted submissions, so 15 slots move from the 80--99 band to
60--79.

\paragraph{Within-band sampling.}
Each problem and score band defines a candidate pool using scores returned by
our UOJ deployment. Reference accounts are excluded. We give preference to
higher scores within a band, then apply the diversity procedure below.

\paragraph{Deduplication and diversity.}
For duplicate detection only, we derive a canonical representation of each
program by removing comments, normalizing whitespace and include order,
replacing identifiers with positional placeholders, and expanding common
contest macros. This representation is used only for comparison. The submitted
source code remains unchanged. Exact canonical duplicates and repeated
submissions from the same author are removed before diversity selection.

For two remaining programs $x$ and $y$ in the same pool, their distance combines
four signals:
\begin{equation*}
\small
  D(x,y) = w_1\bigl(1 - \mathrm{MinHash}_J\bigr)
         + w_2\bigl(1 - \mathrm{AST}_J\bigr)
         + w_3\,\mathrm{feat}_H
         + w_4\bigl(1 - \mathrm{time}_{\cos}\bigr).
\end{equation*}
Here, $\mathrm{MinHash}_J$ estimates token 5-gram Jaccard similarity,
$\mathrm{AST}_J$ compares depth-bounded AST subtrees, and
$\mathrm{feat}_H$ is the normalized Hamming distance between program feature
vectors. These vectors record data structures, recursion, memoization, loop
depth, and function count. The final term compares per-test runtime
vectors by cosine similarity. We use
$(w_1,w_2,w_3,w_4)=(0.25,0.30,0.20,0.25)$. If a parse or timing record is
missing, the corresponding term is dropped and the remaining weights are
rescaled to sum to one.

Selection follows a greedy farthest-first pass. The first program is the fastest
candidate when timing is available and the lowest-ID candidate otherwise. Each
later choice maximizes its minimum distance from the selected set. A candidate
from an author already represented in the set receives a multiplicative
distance penalty.

\paragraph{Candidate verification.}
All selected submissions are already written in C++17 and are evaluated
without source transformation. We rejudge each program on UOJ and retain it
only when its score matches the score used during sampling. Accepted candidates
are judged several times to catch verdict changes caused by runtime variation.
Every released label comes from these UOJ runs.

\section{Training Details}
\label{app:training}

\paragraph{Optimization.}
Each GRPO update uses a batch of 128 prompts, with $n=16$ trajectories sampled
per prompt (2,048 trajectories in total).
We use Adam with a learning rate of $1\times10^{-6}$, a constant learning-rate
schedule, $\beta_1=0.9$, $\beta_2=0.98$, weight decay $0.1$, and gradient
clipping at $1.0$. We optimize a per-token objective with PPO clipping
$\epsilon_{\mathrm{low}}=0.2$ and $\epsilon_{\mathrm{high}}=0.28$. The entropy
and KL coefficients are both zero, and group advantages are centered without
standard-deviation normalization.

\paragraph{Rollouts.}
We sample with temperature $1.0$ and top-$p=1.0$. The context window is capped
at 65,536 tokens, including at most 32,768 response tokens. Each episode allows
at most 30 tool calls and 1,800 seconds. Calls to \textit{execute\_code} and
\textit{execute\_command} have time limits of 120 and 60 seconds, respectively.
We evaluate the held-out validation set every 10 training steps and select the
checkpoint with the best validation performance.

\paragraph{Length penalty.}
We schedule the length penalty manually from the observed reward curve,
enabling it when reward plateaus. It is active during steps $[90,110)$,
$[160,180)$, and $[230,250)$, and disabled at all other steps.

\section{Verification Analysis}
\label{app:full-metrics}

This section examines metric behavior, inference cost, and output patterns
behind the verification results in \S\ref{sec:exp-verification}.

\subsection{Hack Success and Verdict Accuracy Are Complementary}
\label{app:two-channels}

Hack success and verdict accuracy diverge during the unpenalized run in
Figure~\ref{fig:train-reward}. Between steps 79 and 179, hack success changes
only from 59.8\% to 60.9\%, while verdict accuracy falls from 74.8\% to
57.8\%. The stable hack rate masks two opposing changes among unaccepted
candidates. The share skipped after an \emph{Accepted} prediction falls from
17.7\% to 11.3\%, giving the model more opportunities to produce a
counterexample. Over the same interval, the share lost to rollout aborts rises
from 4.5\% to 17.7\% as responses grow longer. These shifts largely cancel in
the aggregate hack metric.

A verifier can therefore attempt counterexamples more broadly while becoming
less reliable in its binary decisions. Hack success records the final attack outcome, and verdict accuracy checks the
preceding decision. The
length-controlled run limits the rise in aborts and retains higher verdict
accuracy, so we report the two metrics together. Appendix~\ref{app:length-penalty}
gives the response-length and timeout diagnostics.

\subsection{Output Tokens and Estimated Cost}
\label{app:cost}

Table~\ref{tab:cost} compares output length with estimated output-token cost.
\textsc{Arbiter} averages 35.0K--46.9K output tokens per candidate across the
four benchmark splits, 22\%--33\% fewer than its agentic Qwen3.6-35B-A3B
baseline, while overall hack success rises from 40.61\% to 62.67\%.
At the listed prices, its mean output cost is \$0.0045 per verification,
compared with \$0.0063 for the agentic Qwen3.6-35B-A3B baseline and \$0.0298 for
agentic DeepSeek-V4-Flash-Preview. At this operating point, \textsc{Arbiter}
has higher overall hack success and lower estimated output-token cost than both
agentic baselines in the table.

We use official provider prices when available and otherwise use prices listed
by OpenRouter.\footnote{\url{https://openrouter.ai/}} These estimates cover
generated output only. Input tokens, caching effects, tool execution, and
self-hosted infrastructure costs are excluded.

\begin{table}[h]
\centering
\renewcommand{\arraystretch}{1.1}
\newcolumntype{R}{>{\centering\arraybackslash}p{1.6cm}}
\resizebox{\linewidth}{!}{%
\begin{tabular}{@{}l !{\color{gray!50}\vrule} r !{\color{gray!50}\vrule} r !{\color{gray!70}\vrule} RR!{\color{gray!70}\vrule}RR !{\color{gray!70}\vrule} r !{\color{gray!50}\vrule} r@{}}
\toprule
\multirow{2}{*}{\textbf{Model}} & \multirow{2}{*}{\textbf{Params}}
  & \multirow{2}{*}{\shortstack{\textbf{Token Price}\\\textbf{(\$/M Tokens)}}}
  & \multicolumn{2}{c!{\color{gray!70}\vrule}}{\textbf{\textsc{UOJ-Bench}}}
  & \multicolumn{2}{c!{\color{gray!70}\vrule}}{\textbf{\textsc{USACO-Judge}}}
  & \multirow{2}{*}{\shortstack{\textbf{Overall}\\\textbf{Hack Succ.}}}
  & \multirow{2}{*}{\shortstack{\textbf{Output Cost}\\\textbf{per Verification}}} \\
\cmidrule(lr){4-5}\cmidrule(lr){6-7}
& & & {\small Easy} & {\small Hard} & {\small Easy} & {\small Hard} & & \\
\midrule
\multicolumn{9}{c}{{\scriptsize\textcolor{gray}{\faBolt}}~\textit{Non-agentic}} \\
\midrule
\logo{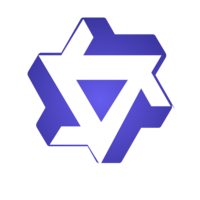}~Qwen3.6-35B-A3B & 35B & \$0.10 & \phantom{0}9.6K & 10.6K & 11.4K & 10.9K & 16.10 & \$0.0011 \\
\logo{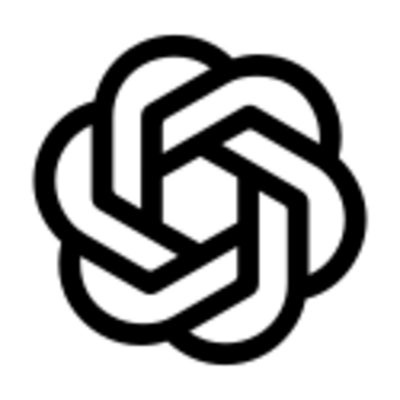}~GPT-OSS-120B & 120B & \$0.17 & \phantom{0}3.8K & \phantom{0}4.5K & \phantom{0}4.3K & \phantom{0}4.1K & 26.59 & \$0.0007 \\
\logo{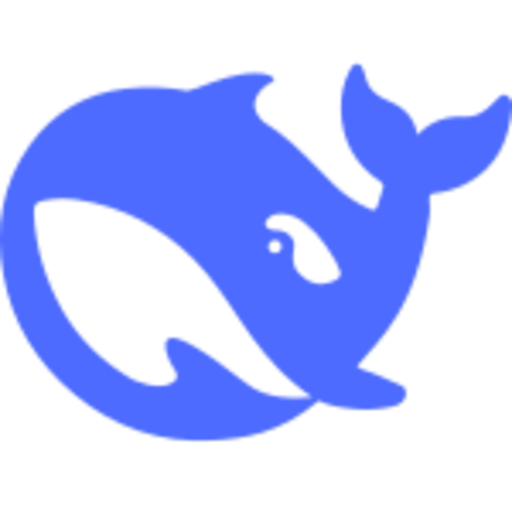}~DeepSeek-V4-Flash-Preview & 284B & \$0.66 & 16.6K & 19.4K & 18.8K & 22.0K & 31.26 & \$0.0129 \\
\logo{qwen.png}~Qwen3.5-397B-A17B & 397B & \$0.90 & 16.7K & 18.4K & 28.1K & 28.2K & 42.73 & \$0.0195 \\
\logo{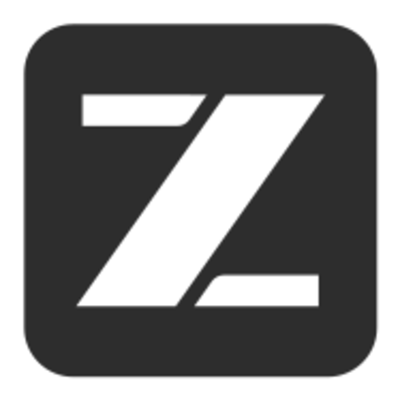}~GLM-5.2 & 753B & \$4.40 & 52.0K & 61.8K & 37.4K & 62.5K & 66.29 & \$0.2550 \\
\logo{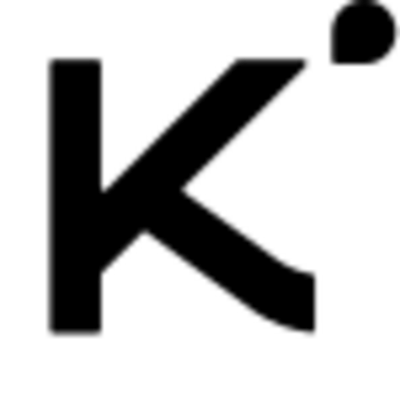}~Kimi-K2.5 & 1T & \$3.00 & 29.1K & 31.6K & 26.5K & 35.2K & 43.07 & \$0.0951 \\
\logo{deepseek.png}~DeepSeek-V4-Pro-Preview & 1.6T & \$1.98 & 32.6K & 37.7K & 29.4K & 42.7K & 51.91 & \$0.0740 \\
\logo{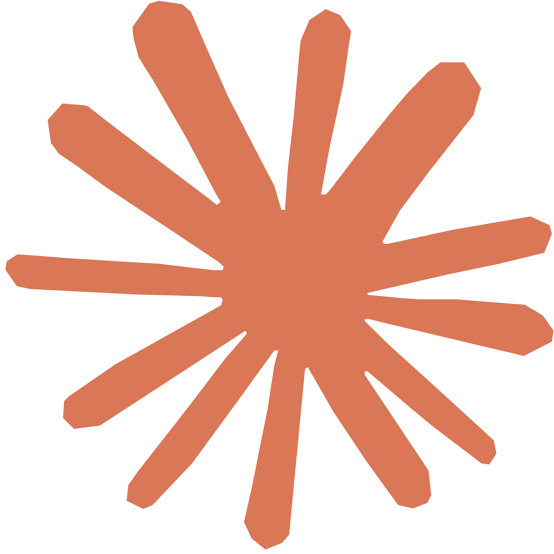}~Claude-Opus-4.8 & -- & \$25.00 & 41.3K & 53.6K & 27.1K & 40.9K & 75.69 & \$1.1316 \\
\logo{openai.png}~GPT-5.5 & -- & \$30.00 & \phantom{0}9.3K & 11.1K & \phantom{0}4.5K & \phantom{0}9.4K & 81.51 & \$0.2907 \\
\midrule
\multicolumn{9}{c}{{\scriptsize\textcolor{gray}{\faCog}}~\textit{Agentic}} \\
\midrule
\logo{qwen.png}~Qwen3.6-35B-A3B & 35B & \$0.10 & 54.1K & 64.6K & 48.0K & 70.0K & 40.61 & \$0.0063 \\
\logo{deepseek.png}~DeepSeek-V4-Flash-Preview & 284B & \$0.66 & 43.2K & 48.9K & 27.8K & 46.3K & 57.76 & \$0.0298 \\
\midrule
\rowcolor{arbRow!6}
\logo{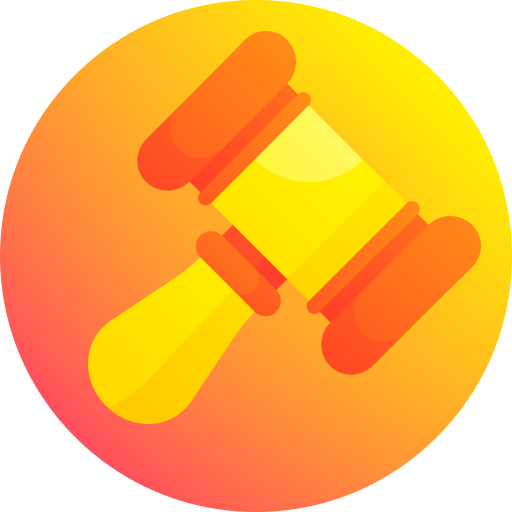}~\textbf{\textsc{Arbiter}} {\small(Ours)} & 35B & \$0.10 & 42.0K & 46.9K & 35.0K & 46.9K & 62.67 & \$0.0045 \\
\bottomrule
\end{tabular}
}
\caption{Mean output tokens per candidate (K), overall hack success, output-token
price, and estimated cost per verification using the reported average output
lengths. Lower output-token usage and cost are better.}
\label{tab:cost}
\end{table}

\input{sections/appendix/verification_funnel}

\begin{table}[H]
\centering
\renewcommand{\arraystretch}{1.1}
\setlength{\tabcolsep}{0pt}
\resizebox{\linewidth}{!}{%
\begin{tabular}{@{}>{\raggedright\arraybackslash}p{4.6cm} !{\color{gray!50}\vrule} >{\centering\arraybackslash}p{1.6cm} !{\color{gray!50}\vrule} *{2}{>{\centering\arraybackslash}p{3.05cm}}!{\color{gray!70}\vrule}*{2}{>{\centering\arraybackslash}p{3.05cm}}@{}}
\toprule
\multirow{2}{*}{\textbf{Model}} & \multirow{2}{*}{\textbf{Params}}
  & \multicolumn{2}{c!{\color{gray!70}\vrule}}{\textbf{\textsc{UOJ-Bench}}}
  & \multicolumn{2}{c}{\textbf{\textsc{USACO-Judge}}} \\
\cmidrule(lr){3-4}\cmidrule(lr){5-6}
& & {\small Easy} & {\small Hard} & {\small Easy} & {\small Hard} \\
\midrule
\logo{qwen.png}~Qwen3.6-35B-A3B & 35B & 16.3 & 17.1 & 12.1 & 16.2 \\
\logo{deepseek.png}~DeepSeek-V4-Flash-Preview & 284B & 21.1 & 21.6 & 15.3 & 22.2 \\
\midrule
\rowcolor{arbRow!6}
\logo{judgement.png}~\textbf{\textsc{Arbiter}} {\small(Ours)} & 35B & 17.5 & 18.7 & 14.2 & 17.8 \\
\bottomrule
\end{tabular}%
}
\caption{Mean interaction turns per candidate in the agentic setting. Non-agentic
baselines use one turn and are omitted.}
\label{tab:turns}
\end{table}

\input{sections/appendix/accepted_trajectory}

\input{sections/appendix/accepted_submission_breaks}

\input{sections/appendix/selection}
\input{sections/appendix/refinement}

\input{sections/appendix/training_ablations}

\input{sections/appendix/length_diagnostics}

\section{Prompts}
\label{app:prompts}

This appendix records the prompt templates used in our experiments: the
non-agentic and agentic system prompts, the per-candidate review request, and
the refinement prompts of \S\ref{sec:exp-refine}. Braced fields, including
\texttt{grading\_language}, are filled per instance; other text is reproduced
verbatim, except where repeated instructions are explicitly marked as elided.

\subsection{Non-agentic System Prompt}
\label{app:nonagentic-prompt}
The non-agentic baselines return the two outputs that can be checked
mechanically: a verdict label and a Python generator whose stdout is the
counterexample input. We use the following system prompt.

\begin{promptbox}{System Prompt --- Non-agentic Baseline}
\scriptsize
You are a \textsc{Hack Agent}. Given a problem statement and a candidate
solution, decide whether the candidate is correct, and if not, produce a
Python~3 program that prints an input that breaks it. The candidate solution
will be graded on the OJ as \{\texttt{grading\_language}\}. You do \textbf{not} need to solve the
problem---only find one counter-example.

\smallskip
Reason step by step. When you are ready, give your final answer as \textbf{exactly
two fenced blocks}, in this order and format:

\smallskip
{\ttfamily
\`{}\`{}\`{}label\\
\# YOUR LABEL\\
\`{}\`{}\`{}\\[4pt]
\`{}\`{}\`{}python\\
\# YOUR CODE HERE\\
\`{}\`{}\`{}
}

\smallskip
The blocks are:
\begin{itemize}\setlength{\itemsep}{2pt}
  \item \texttt{label} --- exactly one of (case-sensitive, no abbreviations):
  \texttt{Accepted} $\vert$ \texttt{Wrong Answer} $\vert$ \texttt{Time Limit
  Exceeded} $\vert$ \texttt{Memory Limit Exceeded} $\vert$ \texttt{Runtime Error}
  $\vert$ \texttt{Compile Error}.
  \item the \texttt{python} block --- a complete deterministic Python~3 program.
  The grader runs \texttt{python3 gen.py} \emph{once}; its stdout becomes the hack
  input fed to the candidate. Stdlib only, no extra stdout chatter, $\leq 30$\,s,
  $\leq 128$\,MB output. The input must satisfy the problem's constraints (a
  validator will reject it otherwise). If you use \texttt{random}, seed it with a
  fixed integer. For \texttt{Accepted} and \texttt{Compile Error} labels---where
  no hack input is required---leave the python block empty.
\end{itemize}

\smallskip
Your final answer must contain only these two fenced blocks.
\end{promptbox}

\subsection{Agentic System Prompt}
\label{app:agentic-prompt}
The agentic \textsc{Arbiter} works in a sandbox with tool access. The
implementation names its final call \texttt{submit\_hack}. The arguments
\texttt{label}, \texttt{verdict}, and \texttt{code} correspond to $v$, $d$,
and $g$, respectively. We use the following system prompt.

\begin{promptbox}{System Prompt --- Agentic \textsc{Arbiter}}
\scriptsize
You have access to these tools:
\begin{itemize}\setlength{\itemsep}{2pt}
  \item \texttt{execute\_code(code)} --- run Python in a sandboxed kernel to
  prototype ideas or generate test cases.
  \item \texttt{execute\_command(command)} --- run shell commands.
  \item \texttt{submit\_hack(label, code, verdict)} --- commit your final
  conclusion. Calling this tool \textsc{ends} the trajectory; you cannot call any
  other tool after it.
\end{itemize}

\smallskip
You are a \textsc{Hack Agent}. Given a problem statement and a candidate solution,
decide whether the candidate is correct, and if not, produce a Python~3 program
that prints an input that breaks it. The candidate solution will be graded on the
OJ as \{\texttt{grading\_language}\}. You do \textbf{not} need to solve the problem---only find one
counter-example.

\smallskip
When you are ready, call \texttt{submit\_hack(label=..., code=..., verdict=...)}.
The arguments are:
\begin{itemize}\setlength{\itemsep}{2pt}
  \item \texttt{label} --- exactly one of (case-sensitive, no abbreviations):
  \texttt{Accepted} $\vert$ \texttt{Wrong Answer} $\vert$ \texttt{Time Limit
  Exceeded} $\vert$ \texttt{Memory Limit Exceeded} $\vert$ \texttt{Runtime Error}
  $\vert$ \texttt{Compile Error}.
  \item \texttt{code} --- a complete deterministic Python~3 program. The grader
  runs \texttt{python3 gen.py} \emph{once}; its stdout becomes the hack input fed
  to the candidate. Stdlib only, no extra stdout chatter, $\leq 30$\,s,
  $\leq 128$\,MB output. The input must satisfy the problem's constraints
  (validator will reject otherwise). If you use \texttt{random}, seed it with a
  fixed integer. For \texttt{Accepted} and \texttt{Compile Error} labels---where no
  hack input is required---pass \texttt{code="{}"}.
  \item \texttt{verdict} --- a short natural-language explanation ($\leq 1000$
  tokens) of the bug and why your input triggers it. For \texttt{Accepted} /
  \texttt{Compile Error}, briefly explain your reasoning.
\end{itemize}

\smallskip
You must judge the candidate solution \textbf{exactly as given}, never a fixed or
rewritten version of it.

\smallskip
\textsc{Important}: you must call \texttt{submit\_hack} to submit your result and
receive reward.
\end{promptbox}

\subsection{Verifier Review Request}
\label{app:review-request}
Within each system-prompt template, only \texttt{grading\_language} varies by benchmark. Each instance
supplies the problem and candidate code through the following request. All reviews use
\texttt{submission\_idx}=1 and \texttt{max\_submissions}=1.

\begin{promptbox}{User Message --- Review Request}
\scriptsize
Here is the problem statement and a candidate solution (submission
\{\texttt{submission\_idx}\} of \{\texttt{max\_submissions}\}) for a
\{\texttt{problem\_type}\} problem. \{\texttt{io\_mode}\}.

\medskip
{\ttfamily \# Problem}

\{\texttt{problem\_description}\}

\medskip
{\ttfamily \# Candidate solution (graded on the OJ as \{grading\_language\})}

{\ttfamily \`{}\`{}\`{}\{language\}}\\
\{\texttt{code}\}\\
{\ttfamily \`{}\`{}\`{}}
\end{promptbox}

\noindent For the refinement experiment, \texttt{problem\_type} is
\texttt{USACO}, \texttt{io\_mode} is \emph{Read input from stdin and write output
to stdout}, and \texttt{grading\_language} and \texttt{language} are both
\texttt{C++17}.

\subsection{Refinement Prompts}
\label{app:refine-prompts}

The refinement experiment of \S\ref{sec:exp-refine} uses the same prompt
skeleton for all five conditions. Round 0 is generated once from the shared
seed prompt. In later rounds, each condition inserts its previous solution as
\{\texttt{code}\} and differs only in the feedback block.

\begin{promptbox}{Round 0 --- Seed Prompt (shared by all five conditions)}
\scriptsize
You are an expert competitive programmer. You will be given a problem statement,
constraints, and example test cases. Please reason step by step about the
solution, then provide a complete implementation in C\texttt{++}17.

\smallskip
Please solve the problem exactly according to the requirements in the problem
statement. Follow the input/output format, function signatures, implementation
details, and any special requirements specified in the problem itself. If the
problem requires standard input/output, include a complete program with
\texttt{main}. If the problem requires implementing specific routines/classes
without \texttt{main}, only implement what is requested.

\smallskip
Assume your code will be run on the OFFICIAL grader, and do not add a main
function, a sample grader, or any other functionality unless it has been
explicitly requested.

\smallskip
Put your final solution within a single code block:

\smallskip
{\ttfamily
\`{}\`{}\`{}cpp\\
// your code here\\
\`{}\`{}\`{}
}

\smallskip
\{\texttt{problem\_statement}\}
\end{promptbox}

\begin{promptbox}{Rounds $1$--$8$ --- Refinement Skeleton (shared by all five conditions)}
\scriptsize
You are an expert competitive programmer. You will be given a problem statement,
constraints, and example test cases, together with a C\texttt{++}17 solution you
wrote earlier for this problem.

\smallskip
Your task is to review and refine that solution. Reason step by step: re-derive
what the problem requires, check the previous solution against it for
correctness, complexity, edge cases, and input/output format, then produce an
improved complete implementation in C\texttt{++}17.

\smallskip
\emph{[The three paragraphs on grader conventions and the fenced-block
instruction are identical to the seed prompt above and are elided here.]}

\smallskip
Output your refined solution in full --- do not abbreviate, and do not refer back
to the previous code with comments like ``unchanged''. Even if you conclude the
previous solution is already correct, restate it in full.

\smallskip
\{\texttt{problem\_statement}\}

\medskip
{\ttfamily \# Your previous solution}

\smallskip
This is the C\texttt{++}17 solution you wrote for this problem last time. It is
not known whether it passes all tests.

\smallskip
{\ttfamily \`{}\`{}\`{}cpp}\\
\{\texttt{previous\_code}\}\\
{\ttfamily \`{}\`{}\`{}}

\smallskip
\fcolorbox{promptframe}{white}{\parbox{0.95\linewidth}{%
\textsc{$\langle$feedback block$\rangle$} --- one of the feedback variants described below.}}
\end{promptbox}

Requiring the complete solution at every round ensures that each response is a
submittable program. An unchanged solution therefore retains its previous
score instead of being treated as missing code.

\begin{promptbox}{Feedback Block --- No Feedback}
\scriptsize
{\ttfamily \# Your task}

\smallskip
This is the code you wrote last time, and it is uncertain whether it passes.
Please refine it and produce the corrected solution.
\end{promptbox}

\begin{promptbox}{Feedback Block --- Online Judge}
\scriptsize
{\ttfamily \# Judge feedback on your previous solution}

\smallskip
Your previous solution was submitted to the online judge and received:

\smallskip
\{\texttt{judge\_summary}\}

\medskip
{\ttfamily \# Your task}

\smallskip
Using the judge feedback above, refine your previous solution and produce the
corrected solution.
\end{promptbox}

\begingroup
\emergencystretch=4em
\noindent\{\texttt{judge\_summary}\} is one line, formatted as
\texttt{"$\langle$verdict$\rangle$ at testcase $\langle$k$\rangle$ (Score:
$\langle$s$\rangle$/100)"}---for example \texttt{"Wrong Answer at testcase 1
(Score: 0/100)"} or \texttt{"Time Limit Exceeded at testcase 6 (Score:
25/100)"}. A compile error is reported as \texttt{"Compile Error"} followed by
the compiler's output in a fenced block. Round 0 is excluded from every arm's
additional-submission budget. A 100-point starting program is retained without
further submissions in every condition. For other trajectories, its existing
result supplies the Online Judge arm's first feedback block. Best@1 then
submits round 1;
Best@8 submits up to eight additional revisions, stops at a 100-point result,
and retains the highest score among additional submissions.
\par
\endgroup

\begin{promptbox}{Feedback Block --- \textsc{Arbiter} ($\mathbf{V}+\mathbf{D}+\mathbf{X}$)}
\scriptsize
{\ttfamily \# Feedback from our verifier on your previous solution}

\smallskip
Your previous solution was reviewed by our verifier, which reported:

\smallskip
\textbf{Label:} \{\texttt{label}\}\hfill{\color{black!55}\scriptsize(V)}

\smallskip
\textbf{Analysis of the bug in your code:}\hfill{\color{black!55}\scriptsize(D)}

\smallskip
\{\texttt{verdict}\}

\smallskip
\textbf{Python program that generates a counterexample
input:}\hfill{\color{black!55}\scriptsize(X)}

\smallskip
{\ttfamily \`{}\`{}\`{}python}\\
\{\texttt{counterexample\_code}\}\\
{\ttfamily \`{}\`{}\`{}}

\medskip
{\ttfamily \# Your task}

\smallskip
Based on the information above, refine your previous solution and produce the
corrected solution.
\end{promptbox}

Each \textsc{Arbiter} condition reviews its own current candidate using the
same verifier and review protocol. The conditions differ in which returned
fields are shown to the solver: $\mathbf{V}+\mathbf{D}+\mathbf{X}$ shows all three,
$\mathbf{V}+\mathbf{D}$ omits the generator, and
$\mathbf{V}+\mathbf{X}$ omits the diagnosis. The verdict is always shown because
it determines the Best@1 stopping decision. When \textsc{Arbiter} returns
\emph{Accepted}, no counterexample is produced, so omitting $\mathbf{X}$
does not remove generator information from that review.
Appendix~\ref{app:refine-x} reports how often $\mathbf{X}$ is non-empty. The
counterexample is marked as Python because it is a generator whose stdout is
the candidate input, not a solution to be repaired.

\smallskip
\noindent If a review exhausts its retry budget without returning a label, that
round uses the no-feedback block and the refinement trajectory continues.

\end{document}

%% file: math_commands.tex
\usepackage{amsmath,amsfonts,bm}

\def\eqref#1{equation~\ref{#1}}

\def\1{\bm{1}}

\DeclareMathAlphabet{\mathsfit}{\encodingdefault}{\sfdefault}{m}{sl}
\SetMathAlphabet{\mathsfit}{bold}{\encodingdefault}{\sfdefault}{bx}{n}

%% file: sections/body/introduction.tex
\section{Introduction}
\label{sec:intro}

\input{sections/figures/overview_ioi_case_study}

Test-time methods allow language models to continue working after an initial
answer. They can generate several candidates for selection or revise an earlier
attempt~\citep{snell2024testtime,shinn2023reflexion,venkatraman2026rsa}. Each use
of additional computation depends on feedback about a candidate already
produced. Selection requires a judgment about which candidate to keep;
revision requires information about a particular failure. Learned verifiers
and reward models have supported candidate selection and guided search across
reasoning tasks~\citep{cobbe2021verifiers,lightman2023process,genrm2024}.
We ask how verification feedback can support both uses while grounding
suspected failures in evidence that can be checked independently.

Competitive programming makes this question concrete. A problem statement
specifies intended behavior, and an official judge tests candidate programs
against hidden inputs and resource limits~\citep{maggiolo2012cms,mares2021security}.
The setting has driven work on code generation and inference-time
computation~\citep{shi2024usaco,o1ioi2025,gencluster2026}. When a solver
compares several candidates or revises an earlier attempt, it needs feedback
before another official submission. Public examples and generated tests can
help with either decision~\citep{codet2023,fu2026refinerl},
but passing them does not establish acceptance on hidden tests. The official
judge determines acceptance under its test suite, but usually provides only a
verdict or score, without a failing input or explanation~\citep{uojbench2026}.
In settings with a finite submission budget, that feedback cannot be requested
freely. A code judge can estimate correctness and describe a suspected fault
for one candidate~\citep{codejudge2024}, but its explanation need not contain
a replayable failing input. Test generators can target known-faulty submissions
or supply tests for debugging and selection~\citep{refute2025,utgen2025,atgen2025,codehacker2026}.
The joint task remains underexplored: given a single candidate of unknown
quality, predict whether it should be accepted and, if a fault is suspected,
produce a candidate-specific counterexample whose legality and effect can be
checked. We formulate this as single-candidate verification with
counterexample-grounded diagnosis.

Given a problem statement and candidate program, \textsc{Arbiter}
uses a generic execution sandbox to return a verdict and diagnosis, plus a
candidate-specific counterexample generator for suspected failures other than
compile errors (Figure~\ref{fig:ioi-climb}, left). It sees neither hidden tests
nor a reference solution. During training and evaluation, judge infrastructure
checks whether generated inputs are legal and break the candidate, providing
executable evidence for suspected failures. We build
\textsc{USACO-Judge} to evaluate acceptance judgments alongside failure
discovery, and train \textsc{Arbiter} with reinforcement learning. On the same
Qwen3.6-35B-A3B backbone, overall counterexample success rises from 16.10\%
without tools to 40.61\% with the agentic scaffold and 62.67\% after training.
On \textsc{USACO-Judge}, mean binary verdict accuracy also rises from 71.41\%
with the agentic scaffold to 87.41\% after training.
The verdict helps rank completed programs. For a rejected candidate, the
diagnosis gives the solver a suspected failure to investigate in its next
attempt. In an IOI 2026 case study, \textsc{Arbiter}-guided refinement brings
GPT-OSS-120B to 385.46 expected points under the official submission limit,
above the gold threshold. It reaches 334.6 expected points with
online-judge feedback (Figure~\ref{fig:ioi-climb}, right). To conclude, our
contributions are:
\begin{itemize}[leftmargin=1.4em,itemsep=3pt,topsep=3pt]

    \item We formulate a task for judging a single candidate program and explaining
    any unaccepted verdict, and use counterexamples that break the program as a verifiable
    proxy for diagnosis quality.

    \item We release \textsc{USACO-Judge}, which evaluates correct and incorrect
    programs from human experts and multiple LLMs against
    official test sets from the USACO 2026 contests.

    \item We introduce \textsc{Arbiter}, a 35B MoE agentic verifier trained with
    reinforcement learning, and show substantial gains over its base model and
    strong results on \textsc{USACO-Judge}, code selection, and refinement.

\end{itemize}

%% file: sections/figures/overview_ioi_case_study.tex
\vspace{-7pt}
\begin{figure}[!ht]
\centering
\definecolor{rkOurs}{HTML}{FF9900}%
\definecolor{tkNeutral}{HTML}{8E8375}
\colorlet{rkNon}{tkNeutral!58}%
\colorlet{tkIn}{tkNeutral}%
\colorlet{tkOJ}{tkNeutral}%
\definecolor{tkDiag}{HTML}{C87A00}%
\definecolor{tkGen}{HTML}{A8321D}%

\newsavebox{\ioitaskpanel}
\savebox{\ioitaskpanel}{%
\begin{tikzpicture}[
    x=1cm, y=1cm, font=\tiny,
    chip/.style={rounded corners=2pt, line width=0.5pt, inner sep=0pt,
                 align=center, minimum height=0.50cm},
    held/.style={chip, dash pattern=on 1.6pt off 1.1pt},
    sys/.style={sharp corners, line width=0.85pt, inner sep=0pt,
                align=center, minimum height=0.62cm},
    otag/.style={rounded corners=5pt, line width=0.7pt, inner sep=0pt,
                 align=center, minimum height=0.34cm, minimum width=4.20cm},
    sbox/.style={rounded corners=1pt, line width=0.4pt, densely dotted,
                 inner sep=0pt, align=center, minimum height=0.30cm,
                 font=\tiny},
    wire/.style={gray!60, line width=0.5pt},
    flow/.style={wire, -{Stealth[length=3.2pt,width=2.8pt]}},
  ]
  \node[chip, draw=tkIn!70, fill=tkIn!8, minimum width=1.95cm] (prob) at (0.975,2.50)
    {{\faFile}~\textbf{Problem}};
  \node[chip, draw=tkIn!70, fill=tkIn!8, minimum width=1.95cm] (cand) at (3.425,2.50)
    {{\faCode}~\textbf{Candidate}};
  \node[held, draw=tkOJ!80, fill=tkOJ!6, minimum width=1.45cm] (td) at (5.625,2.50)
    {{\faDatabase}~\textbf{Test Data}};
  \node[held, draw=tkOJ!80, fill=tkOJ!6, minimum width=1.40cm] (chk) at (7.70,2.50)
    {{\faBalanceScale}~\textbf{Checker}};
  \node[anchor=south east, inner sep=0pt, text=tkOJ!85,
        xshift=-1.3pt, yshift=1.3pt] at (td.south east) {\scalebox{0.62}{\faLock}};
  \node[anchor=south east, inner sep=0pt, text=tkOJ!85,
        xshift=-1.3pt, yshift=1.3pt] at (chk.south east) {\scalebox{0.62}{\faLock}};
  \draw[wire] (0.975,2.25) -- (0.975,2.06) -- (3.20,2.06);
  \draw[wire] (3.20,2.25) -- (3.20,2.06);
  \draw[flow] (2.20,2.06) -- (2.20,1.73);
  \draw[wire] (3.65,2.25) -- (3.65,2.06) -- (7.70,2.06) -- (7.70,2.25);
  \draw[wire] (5.625,2.25) -- (5.625,2.06);
  \draw[flow] (6.625,2.06) -- (6.625,1.73);
  \node[sys, draw=rkOurs, fill=rkOurs!8, minimum width=4.40cm,
        minimum height=0.90cm] (arb) at (2.20,1.25) {};
  \node[anchor=north, inner sep=0pt] at (2.20,1.57)
    {\logo[1.02em]{judgement.png}~\textbf{Agentic Judge}};
  \node[sbox, draw=rkOurs!65, fill=white, minimum width=1.50cm] at (1.10,1.05)
    {{\color{rkOurs!90}\faBrain}~LLM};
  \node[sbox, draw=rkOurs!65, fill=white, minimum width=2.20cm] at (3.24,1.05)
    {{\color{rkOurs!90}\faTerminal}~Sandbox};
  \node[sys, draw=tkOJ!75, fill=tkOJ!7, minimum width=3.55cm,
        minimum height=0.90cm] (oj) at (6.625,1.25) {};
  \node[anchor=north, inner sep=0pt] at (6.625,1.57)
    {\logo[1.02em]{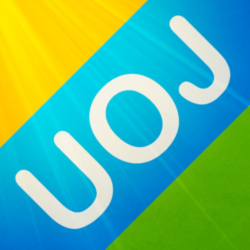}~\textbf{Online Judge}};
  \node[sbox, draw=tkOJ!70, fill=white, minimum width=2.55cm] at (6.625,1.05)
    {{\color{tkOJ}\faTerminal}~Sandbox};
  \draw[flow] (2.20,0.80) -- (2.20,0.50);
  \draw[flow] (6.625,0.80) -- (6.625,0.50);
  \node[draw=rkOurs!55, fill=rkOurs!4, rounded corners=3pt, line width=0.5pt,
        minimum width=4.40cm, minimum height=1.44cm, inner sep=0pt]
        at (2.20,-0.25) {};
  \node[otag, draw=tkOJ!85, fill=tkOJ!9] at (2.20,0.17)
    {{\color{tkOJ}\faGavel}~\textbf{Verdict} \;{\tiny accept, or not}};
  \node[otag, draw=tkDiag, fill=tkDiag!11] at (2.20,-0.25)
    {{\color{tkDiag}\faCommentDots}~{\color{tkDiag}\textbf{Diagnosis}} \;{\tiny why it is wrong}};
  \node[otag, draw=tkGen, fill=tkGen!12] at (2.20,-0.67)
    {{\color{tkGen}\faBug}~{\color{tkGen}\textbf{Counterexample}} \;{\tiny an input it fails on}};
  \node[draw=tkOJ!30, fill=white, rounded corners=3pt, line width=0.5pt,
        minimum width=3.55cm, minimum height=1.44cm, inner sep=0pt]
        at (6.625,-0.25) {};
  \node[otag, draw=tkOJ!85, fill=tkOJ!9, minimum width=3.35cm] at (6.625,0.17)
    {{\color{tkOJ}\faGavel}~\textbf{Verdict} \;{\tiny accept, or not}};
  \node[otag, draw=tkOJ!50, fill=white, minimum width=3.35cm,
        dash pattern=on 1.5pt off 1.1pt] at (6.625,-0.25)
    {{\color{tkOJ!70}\faCommentDots}~{\color{tkOJ!80}\textbf{Diagnosis}}
     \;{\tiny\itshape\color{tkOJ!80}never given}};
  \node[otag, draw=tkOJ!50, fill=white, minimum width=3.35cm,
        dash pattern=on 1.5pt off 1.1pt] at (6.625,-0.67)
    {{\color{tkOJ!70}\faBug}~{\color{tkOJ!80}\textbf{Counterexample}}
     \;{\tiny\itshape\color{tkOJ!80}leaks tests}};
\end{tikzpicture}}

\definecolor{ioOurs}{HTML}{FF9900}%
\definecolor{ioJudge}{HTML}{8E8375}%
\definecolor{ioGold}{HTML}{B9902E}%
\definecolor{ioSilver}{HTML}{9AA0A6}%
\definecolor{ioBronze}{HTML}{A9673A}%
\definecolor{ioGoldZone}{HTML}{F3E3B8}%
\definecolor{ioSilverZone}{HTML}{E7E9EB}%
\definecolor{ioBronzeZone}{HTML}{F0E1D5}%
\definecolor{ioOursZone}{HTML}{FFE3AC}%

\newsavebox{\ioipanel}
\savebox{\ioipanel}{%
\begin{tikzpicture}
\begin{axis}[
    width=3.45cm, height=2.3718cm, scale only axis,
    xmin=-0.25, xmax=5.55, ymin=175, ymax=428,
    xtick={0,1,2,3,4,5},
    xticklabels={0,1,2,3,4,5},
    ytick={200,250,300,350,400},
    tick label style={font=\tiny, /pgf/number format/assume math mode=true},
    xlabel={Refinement Round}, ylabel={\textbf{IOI 2026} Contest Score},
    xlabel style={font=\tiny, yshift=2pt},
    ylabel style={font=\tiny, yshift=-4pt},
    axis lines=left,
    tick align=inside,
    axis line style={gray!55, -{Stealth[length=3.2pt,width=2.8pt]}},
    tick style={gray!55},
    axis on top,
    clip=false,
    legend style={at={(0.5,1.01)}, anchor=south, draw=none, fill=none,
                  font=\tiny, column sep=5pt,
                  /tikz/every even column/.append style={column sep=5pt},
                  legend image post style={line width=1.0pt}},
    legend cell align=left,
    legend columns=-1,
]
  \addplot[ioGoldZone, fill=ioGoldZone, draw=none, forget plot]
    coordinates {(-0.25,361.12) (5.55,361.12) (5.55,428) (-0.25,428)} \closedcycle;
  \addplot[ioSilverZone, fill=ioSilverZone, draw=none, forget plot]
    coordinates {(-0.25,303.28) (5.55,303.28) (5.55,361.12) (-0.25,361.12)} \closedcycle;
  \addplot[ioBronzeZone, fill=ioBronzeZone, draw=none, forget plot]
    coordinates {(-0.25,228.80) (5.55,228.80) (5.55,303.28) (-0.25,303.28)} \closedcycle;

  \addplot[ioGold, densely dashed, line width=0.8pt, mark=none, forget plot]
    coordinates {(-0.25,361.12) (5.55,361.12)};
  \node[font=\tiny, text=ioGold, anchor=south, inner sep=1.5pt]
    at (axis cs:2.5,363) {Gold};

  \addplot[ioSilver, densely dashed, line width=0.7pt, mark=none, forget plot]
    coordinates {(-0.25,303.28) (5.55,303.28)};
  \node[font=\tiny, text=ioSilver, anchor=south, inner sep=1.5pt]
    at (axis cs:1.6,305.5) {Silver};

  \addplot[ioBronze, densely dashed, line width=0.7pt, mark=none, forget plot]
    coordinates {(-0.25,228.80) (5.55,228.80)};
  \node[font=\tiny, text=ioBronze, anchor=south, inner sep=1.5pt]
    at (axis cs:3.7,231) {Bronze};

  \addplot[ioOurs, line width=1.15pt, mark=triangle*, mark size=1.6pt,
           mark options={fill=ioOurs, draw=white, line width=0.3pt, rotate=180}]
    coordinates {(0,240.49) (1,356.91) (2,356.57) (3,356.46) (4,385.46) (5,385.46)};
  \addlegendentry{by \textsc{Arbiter}}
  \addplot[ioOurs, opacity=0.20, mark=*, mark size=4.0pt,
           mark options={fill=ioOurs, draw=none}, forget plot]
    coordinates {(4,385.46)};
  \addplot[ioOurs, only marks, mark=triangle*, mark size=2.6pt,
           mark options={fill=ioOurs, draw=white, line width=0.3pt, rotate=180}, forget plot]
    coordinates {(4,385.46)};
  \node[font=\tiny\bfseries, text=ioOurs, anchor=south east, inner sep=1.5pt]
    at (axis cs:5.15,385.46) {385.46};

  \addplot[ioJudge, line width=0.9pt, mark=triangle*, mark size=1.5pt,
           mark options={fill=ioJudge, draw=white, line width=0.3pt}]
    coordinates {(0,189.0) (1,248.2) (2,262.8) (3,283.2) (4,321.0) (5,334.6)};
  \addlegendentry{by Online Judge}
  \node[font=\tiny, text=ioJudge, anchor=south east, inner sep=1.5pt]
    at (axis cs:4.95,331) {334.60};
\end{axis}
\end{tikzpicture}}

\makebox[0.618\linewidth][c]{\scalebox{1.18}{\raisebox{-0.5\height}{\usebox{\ioitaskpanel}}}}%
\makebox[0.382\linewidth][c]{%
  \scalebox{1.18}{\raisebox{26.53pt}{\raisebox{-0.5\height}{%
    \tikz[baseline]{\node[draw=gray!25, rounded corners=4pt, line width=0.4pt,
      inner sep=3pt] {\usebox{\ioipanel}\hspace{4pt}};}}}}}
\captionsetup{skip=4pt}
\caption{Comparison of \textsc{Arbiter} with an online judge (\textbf{left}) and the IOI 2026 case study (\textbf{right}). Under the same 50-submission cap per problem, the expected \textsc{Arbiter} score crosses the gold threshold in round 4; the expected online-judge-feedback score remains below it.}
\label{fig:ioi-climb}
\end{figure}
\vspace{-7pt}

%% file: sections/body/related_work.tex
\section{Related Work}
\label{sec:related}

\paragraph{Feedback for Test-Time Scaling.}
Test-time scaling can use feedback to select among generated
solutions~\citep{alphacode2022} or revise an earlier
attempt~\citep{venkatraman2026rsa,wan2026deepverifier}.
Outcome and process reward models score complete solutions or intermediate
steps~\citep{cobbe2021verifiers,lightman2023process}. Generative verifiers
provide reasoning for their judgments~\citep{genrm2024}, while agentic
verifiers gather observations through tools~\citep{zhang2026agentvrl,cui2026vagen}.
These approaches differ in both how they produce feedback and what that
feedback establishes about a candidate. Our work connects the two uses of
feedback: \textsc{Arbiter} provides verdicts for selection and
candidate-specific diagnoses and counterexamples for refinement.

\paragraph{Evidence for Code Judgment.}
Code judges can predict correctness and describe faults for a single
program~\citep{codejudge2024}, but such descriptions need not include a
replayable failing input. Execution-based methods use tests to rank candidate
pools or guide revisions~\citep{lever2023,codet2023,agenticverifier2026,fu2026refinerl,fixaudit2026}.
Test generators target known-faulty programs or supply tests for debugging and
selection~\citep{refute2025,utgen2025,atgen2025,codehacker2026}. These outputs
support different conclusions. A predicted verdict is not an executable
witness. An input on which two programs disagree shows that their behavior
differs, but not which program is wrong. A legal input that breaks one
candidate under the task's judging rules supports its rejection. Our task
predicts acceptance for a candidate of unknown quality and, for suspected
non-compilation failures, couples diagnosis with candidate-specific
counterexample search.

\paragraph{Evaluating Code Verifiers.}
Benchmarks assess correctness judgments and fault categories~\citep{zhao2025codejudgeeval},
and study judge behavior across code generation, repair, and unit-test
generation~\citep{jiang2026codejudgebench}. Our task also asks whether a
claimed fault can be exposed by an executable input. Hacking benchmarks
measure counterexample discovery, but the Easy and Hard \textsc{UOJ-Bench}
splits used here contain only faulty programs~\citep{uojbench2026}. They cannot
test whether a verifier recognizes accepted code. We introduce
\textsc{USACO-Judge} to address this gap: it includes accepted and unaccepted
programs from human experts and LLMs, measuring verdict accuracy across both
classes and counterexample success on incorrect candidates under one protocol.

%% file: sections/body/training.tex
\section{Training \textsc{Arbiter}}
\label{sec:verifier}

We train \textsc{Arbiter} (\textbf{A}gentic \textbf{R}L-trained
\textbf{B}ug-finding \textbf{I}nteractive \textbf{T}est
\textbf{E}xamine\textbf{R}) to judge a candidate program, diagnose failures,
and generate counterexamples. Starting from Qwen3.6-35B-A3B~\citep{qwen36_35b_a3b}, we optimize
the mapping $(\mathcal{P},\mathcal{C}) \mapsto (v,d,g)$ with agentic reinforcement
learning. We describe the training data in Section~\ref{sec:verifier-traindata},
the reward design and optimization in Section~\ref{sec:verifier-rl}, and adaptive
length control in Section~\ref{sec:verifier-length}.

\subsection{Training data}
\label{sec:verifier-traindata}

Our training set contains 20,000 candidate programs sampled from
online-judge-graded solver rollouts across 6,145 competitive-programming
problems. We sample at most five solver rollouts per problem and exclude all
evaluation problems. Accepted programs make up 15\% of the set. Among the
remaining programs, 46.8\% score between 76 and 99. These high-scoring failures
are more likely to contain subtle bugs that survive most official tests.
A separate validation set of 400 candidates from 100 hard problems is reserved
for checkpoint selection and shares no problems with the training or test sets.
Appendix~\ref{app:traindata} provides full sampling and construction details.

\subsection{Agentic reinforcement learning}
\label{sec:verifier-rl}
\label{sec:verifier-reward}

For each $(\mathcal{P},\mathcal{C})$ pair, the policy samples a group of multi-turn episodes in the
sandbox of Section~\ref{sec:env}. Each episode ends with one $(v,d,g)$ prediction
and receives a scalar reward. We optimize the policy with GRPO
\citep{shao2024deepseekmath} using the slime RL framework
\citep{slime_github}. Episodes that hit the token budget are truncated and retained with zero reward,
whereas episodes that exceed the wall-clock budget are aborted and excluded
from the policy update. Appendix~\ref{app:training} reports the training and
rollout hyperparameters.

\paragraph{Rewarding executable evidence.}
A verdict-only reward invites a shortcut: with 85\% of training candidates
unaccepted, the policy can favor the majority verdict without testing the
candidate. In a diagnostic run, training reward rises while validation verdict
accuracy stays between 56\% and 58\%. Mean tool calls fall from 4.9 per
rollout to below one (Figure~\ref{fig:label-only-trap}, left). We therefore
give little reward for a correct unaccepted verdict unless the policy produces
a counterexample that survives execution and validation.
Appendix~\ref{app:ablation-replay} gives the full diagnostic.

\input{sections/figures/verification_reward_pipeline}

Figure~\ref{fig:reward-flow} shows the reward allocation. Agreement on an
unaccepted verdict earns $0.05$: it identifies the direction of the decision
but supplies no evidence of a bug. Running $g$ earns another $0.05$, and
producing a legal input adds $0.10$. These small intermediate rewards give
feedback before the harder event of finding a failure. Their total, including
verdict agreement, is $0.20$, while an input that actually breaks
$\mathcal{C}$ earns $0.75$. Matching the failure type
adds the final $0.05$, bringing a verified unaccepted prediction to $1.00$.
When both $v$ and the reference verdict $v^\star$ are \textit{Accepted}, the
rollout receives $1.00$. This evidence gate is necessarily asymmetric in our
setting: a counterexample can establish failure, but not finding one cannot
establish acceptance. The policy can still guess \textit{Accepted} without a
witness. The diagnosis $d$ receives no correctness reward.
A diagnosis over $1000$ tokens incurs a $0.05$ penalty. A valid
counterexample that overturns an \textit{Accepted} reference verdict earns an
additional $0.10$, placing verified evidence above agreement with the official
tests. Appendix~\ref{app:accepted-breaks} gives two such cases.

\paragraph{False-positive penalty.}
An incorrect \textit{Accepted} verdict can stop a refinement trajectory before
the faulty program is repaired. We examine an additional reward penalty
$\lambda$ for predicting \textit{Accepted} when the reference verdict is
unaccepted. Matched runs with $\lambda=0$ and $\lambda=0.5$ show higher
\textit{Accepted}-class precision and lower recall with the larger penalty
(Figure~\ref{fig:label-only-trap}, right). On the hard split of
\textsc{USACO-Judge}, precision rises from 77.78\% to 80.98\%, while recall
falls from 76.67\% to 62.86\%. A false negative can also discard a correct
program during selection. We use $\lambda=0$ for the final \textsc{Arbiter};
the comparison shows how the penalty shifts the precision--recall trade-off. The full
benchmark comparison appears in Appendix~\ref{app:ablation-shaping}.

\begingroup
\setlength{\intextsep}{3pt}
\input{sections/figures/reward_ablations}
\endgroup

\subsection{Efficient training through length control}
\label{sec:verifier-length}

As responses grow longer during training, generation cost
rises and rollout throughput falls. Inspired by length control
\citep{kimik25}, we design an adaptive penalty. When the penalty is active,
we penalize a rollout only if it earns at least $0.9$ raw reward, exceeds a
group-relative length threshold, and at least one-quarter of the group shares
the top reward. These gates spare unsuccessful attempts and rare successes on
difficult problems. For a group $G$, let $S\subseteq G$ be the rollouts tied
for its highest raw reward, and $B$ the upper quartile of their response
lengths, recomputed at each step. For rollout $i$,
let $R_i$ and $L_i$ denote its raw reward and response length. The adjusted
reward is
\begin{equation*}
\small
\widetilde R_i =
\begin{cases}
R_i-0.6, & R_i\geq 0.9,\ L_i>B,\ \frac{|S|}{|G|}\geq\frac{1}{4},\\
R_i, & \text{otherwise}.
\end{cases}
\end{equation*}
Figure~\ref{fig:train-reward} compares two runs continued from the same
checkpoint. Length control shortens responses (middle), reduces rollout
timeouts (right), and avoids the late decline in mean reward seen without it
(left). A continuously applied penalty can curtail exploration:
in a comparison run, median response length falls from 17.2k to 12.9k tokens,
median interaction turns drop from 10 to 7, and hack success falls from 62.3\%
to 59.1\% (Appendix~\ref{app:length-penalty}). We therefore alternate
penalized and unpenalized phases to leave room for longer attempts. The shaded
intervals mark when the penalty is active.

\begingroup
\setlength{\intextsep}{3pt}
\input{sections/figures/adaptive_length_control}
\endgroup

%% file: sections/figures/verification_reward_pipeline.tex
\begin{figure}[H]
\centering
\definecolor{rwOurs}{HTML}{FF9900}%
\definecolor{rwNeutral}{HTML}{7F8790}%
\definecolor{rwGen}{HTML}{FF9900}%
\definecolor{rwDiag}{HTML}{FF9900}%
\definecolor{rwPay}{HTML}{4F7F73}%
\definecolor{rwNil}{HTML}{A5A9AD}%
\definecolor{rwPen}{HTML}{A06B6B}%
\definecolor{rwCmp}{HTML}{8B8174}%
\definecolor{rwBonus}{HTML}{4F7F73}%
\resizebox{\linewidth}{!}{%
\begin{tikzpicture}[
    x=1cm, y=1cm, font=\tiny,
    chip/.style ={rounded corners=2pt, line width=0.5pt, inner sep=2pt,
                  align=center, minimum height=0.46cm, minimum width=1.60cm},
    sys/.style  ={sharp corners, line width=0.85pt, inner sep=2pt,
                  align=center, minimum height=1.05cm, minimum width=1.85cm},
    otag/.style ={rounded corners=5pt, line width=0.7pt, inner sep=2pt,
                  align=center, minimum height=0.40cm, minimum width=2.05cm},
    dec/.style  ={diamond, aspect=2.1, draw=rwCmp!70, fill=rwCmp!10,
                  line width=0.7pt, inner sep=1pt, font=\tiny},
    pay/.style  ={rounded corners=4pt, line width=0.7pt, inner sep=2pt,
                  align=center, minimum height=0.38cm, minimum width=1.30cm},
    wire/.style ={gray!60, line width=0.5pt},
    flow/.style ={wire, -{Stealth[length=3.2pt,width=2.8pt]}},
    inc/.style  ={font=\tiny\bfseries, text=rwPay, inner sep=1pt, fill=white},
    edge/.style ={font=\tiny, text=black!55, inner sep=1pt, fill=white},
  ]
  \def\cA{0}     \def\cB{2.30}  \def\cC{5.25}  \def\cD{8.05}
  \def\cE{10.85} \def\cP{13.72} \def\cQ{11.35}
  \def\rV{2.05}  \def\rX{1.10}  \def\rD{0.15}  \def\dY{0.31667}
  \def\rJ{3.05}  \def\wSys{2.05}  \def\hSys{0.86}

  \node[chip, draw=rwNeutral!70, fill=rwNeutral!8] (cand) at (\cA,\rX+0.31)
    {{\faCode}~\textbf{Candidate}};
  \node[chip, draw=rwNeutral!70, fill=rwNeutral!8] (prob) at (\cA,\rX-0.31)
    {{\faFile}~\textbf{Problem}};
  \node[sys, draw=rwOurs, fill=rwOurs!8, minimum width=\wSys cm,
        minimum height=\hSys cm] (arb) at (\cB,\rX)
    {{\color{rwOurs}\faSearch}\\[1pt]\textbf{\textsc{Arbiter}}};
  \draw[wire] (cand.east) -- ($(cand.east)+(0.28,0)$) -- ($(prob.east)+(0.28,0)$);
  \draw[wire] (prob.east) -- ($(prob.east)+(0.28,0)$);
  \draw[flow] ($(cand.east)!0.5!(prob.east)+(0.28,0)$) -- (arb.west);

  \node[otag, draw=rwOurs, fill=rwOurs!8] (vd) at (\cC,\rV)
    {{\color{rwOurs!70!black}\faGavel}~{\color{rwOurs!70!black}\textbf{Verdict}}};
  \node[otag, draw=rwDiag, fill=rwDiag!8] (dg) at (\cC,\rD)
    {{\color{rwDiag!70!black}\faCommentDots}~{\color{rwDiag!70!black}\textbf{Diagnosis}}};
  \node[otag, draw=rwGen, fill=rwGen!8] (cx) at (\cC,\rX)
    {{\color{rwGen!70!black}\faBug}~{\color{rwGen!70!black}\textbf{Counterexample}}};
  \coordinate (obus) at ($(arb.east)+(0.42,0)$);
  \draw[wire] (arb.east) -- (obus);
  \draw[wire] ($(obus |- vd)$) -- ($(obus |- dg)$);
  \draw[flow] ($(obus |- vd)$) -- (vd.west);
  \draw[flow] ($(obus |- dg)$) -- (dg.west);
  \draw[flow] ($(obus |- cx)$) -- (cx.west);

  \node[sys, draw=rwNeutral!75, fill=rwNeutral!7, minimum width=\wSys cm,
        minimum height=\hSys cm] (oj) at (\cB,\rJ)
    {{\color{rwNeutral}\faServer}\\[1pt]\textbf{Online Judge}};
  \draw[flow] (cand.north) |- (oj.west);
  \node[dec, aspect=1.8] (sw) at (\cC+2.15,\rV+0.50) {agree?};
  \draw[flow] (vd.east) -| (sw.south);
  \node[otag, draw=rwNeutral!60, fill=white] (rf) at (\cC,\rJ)
    {{\color{rwNeutral!75}\faGavel}~\textbf{Reference}};
  \draw[flow] (oj.east) -- (rf.west);
  \draw[flow] (rf.east) -| (sw.north);

  \node[pay, draw=rwPay, fill=rwPay!10] (payAC) at (\cQ,\rV+1.06)
    {\color{rwPay}\textbf{+1.00}};
  \node[pay, draw=rwPay!85, fill=rwPay!7] (payDir) at (\cQ,\rV+0.50)
    {\color{rwPay}\textbf{+0.05}};
  \node[pay, draw=rwNil!70, fill=rwNil!8] (payACno) at (\cQ,\rV)
    {\color{rwNil!130}0};
  \coordinate (sT)  at (\cD+0.55,\rV);
  \coordinate (tAC) at (\cD+0.55,\rV+1.06);
  \coordinate (tNA) at (\cD+0.55,\rV+0.50);
  \draw[wire] (sw.east) -- (tNA);
  \draw[wire] (sT) -- (tAC);
  \draw[flow] (tAC) -- (payAC.west);
  \draw[flow] (tNA) -- (payDir.west);
  \draw[flow] (sT)  -- (payACno.west);
  \node[edge, anchor=south] at ($(tAC)!0.5!(payAC.west)+(0,0.02)$)
    {both \textit{\color{rwPay}Accepted}};
  \node[edge, anchor=south] at ($(tNA)!0.5!(payDir.west)+(0,0.02)$)
    {both \textit{\color{rwOurs!70!black}Unaccepted}};
  \node[edge, anchor=south] at ($(sT)!0.5!(payACno.west)+(0,0.02)$)
    {disagree};

  \node[chip, draw=rwNeutral!70, fill=rwNeutral!8, minimum width=1.75cm] (val)
    at (\cD,\rX) {{\faRuler}~\textbf{Validator}};
  \node[chip, draw=rwNeutral!70, fill=rwNeutral!8, minimum width=1.75cm] (chk)
    at (\cE,\rX) {{\faBalanceScale}~\textbf{Checker}};
  \draw[flow] (cx.east) -- (val.west);
  \draw[flow] (val.east) -- (chk.west);
  \node[inc, anchor=south] at ($(cx.east)!0.5!(val.west)+(0,0.04)$) {+0.05};
  \node[edge, anchor=north] at ($(cx.east)!0.5!(val.west)+(0,-0.04)$) {runs};
  \node[inc, anchor=south] at ($(val.east)!0.5!(chk.west)+(0,0.04)$) {+0.10};
  \node[edge, anchor=north] at ($(val.east)!0.5!(chk.west)+(0,-0.04)$) {legal};

  \node[pay, draw=rwPay, fill=rwPay!10] (payX) at (\cP,\rX+\dY)
    {\color{rwPay}\textbf{+0.75}};
  \node[pay, draw=rwNil!70, fill=rwNil!8] (payXno) at (\cP,\rX-\dY)
    {\color{rwNil!130}0};
  \coordinate (sB) at ($(chk.east)+(0.22,0)$);
  \draw[wire] (chk.east) -- (sB);
  \draw[wire] ($(sB)+(0,-\dY)$) -- ($(sB)+(0,\dY)$);
  \draw[flow] ($(sB)+(0,\dY)$)  -- (payX.west);
  \draw[flow] ($(sB)+(0,-\dY)$) -- (payXno.west);
  \coordinate (lx) at ($(sB)+(0.10,0)$);
  \node[edge, anchor=south west, fill=none] at (lx |- payX)    {breaks};
  \node[edge, anchor=south west, fill=none] at (lx |- payXno)  {survives};

  \node[pay, draw=rwPen!75, fill=rwPen!7, minimum width=1.30cm, anchor=west] (pen)
    at (chk.west |- 0,\rD) {\color{rwPen}\textbf{\textminus0.05}};
  \draw[flow] (dg.east) -- (pen.west)
    node[edge, midway, anchor=south]{over 1000 tokens};

  \coordinate (mid) at ($(vd.south)!0.5!(chk.north)$);
  \coordinate (cUp) at ($(chk.north)+(0.30,0)$);
  \coordinate (mR)  at (cUp |- mid);
  \coordinate (mLL) at (mR -| vd.south);
  \draw[wire, densely dashed] ($(chk.north)+(0.30,0)$) -- (mR) -- (mLL) -- (vd.south);
  \node[inc, fill=white, inner sep=1.5pt] at (val.north |- mR)
    {+0.05 {\mdseries\color{black!55}if matches}};

  \node[pay, draw=rwBonus!85!black, fill=rwBonus!16, minimum width=1.30cm] (payBon)
    at (\cP,\rV) {\color{rwPay!75!black}\textbf{+0.10}};
  \draw[flow] (payACno.east) -- (payBon.west);
  \draw[flow] (payX.north) -- (payBon.south);
  \node[edge, anchor=south] at ($(payACno.east)!0.5!(payBon.west)+(0,0.02)$)
    {and};
\end{tikzpicture}}
\vspace{-14pt}
\caption{The verification pipeline assigns partial rewards according to each outcome and combines them into the final reward used to train \textsc{Arbiter}, with each component tied to a separate check.}
\label{fig:reward-flow}
\end{figure}

%% file: sections/figures/reward_ablations.tex
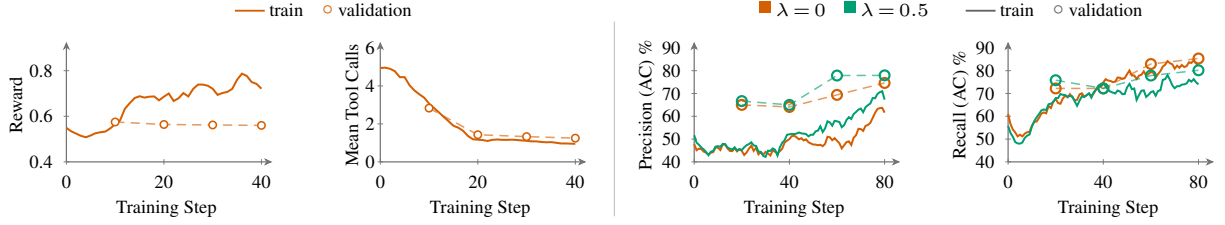
\begin{figure}[H]
\centering
\definecolor{abBlue}{HTML}{009E73}%
\definecolor{abRed}{HTML}{D55E00}%
\colorlet{abReward}{abRed}%
\colorlet{abEval}{abRed}%
\colorlet{abTurns}{abRed}%
\colorlet{abFpp5}{abBlue}%
\colorlet{abFpp0}{abRed}%
\scalebox{1.16}{%
\begin{tikzpicture}
\pgfplotsset{
  abpanel/.style={
    width=2.28cm, height=1.30cm, scale only axis,
    xmin=0, xmax=41, xtick={0,20,40},
    axis lines=left,
    axis line style={-{Stealth[length=2.8pt,width=2.4pt]}, black!55},
    x axis line style={shorten >=-2pt}, y axis line style={shorten >=-2pt},
    enlarge x limits=false, enlarge y limits=false,
    xlabel={Training Step},
    label style={font=\tiny},
    tick label style={font=\tiny, /pgf/number format/assume math mode=true},
    major tick length=1.8pt,
    ylabel style={yshift=-0.16cm}, xlabel style={yshift=0.10cm},
    clip=false,
  },
  abrew/.style={ylabel={Reward}, ymin=0.40, ymax=0.90,
                ytick={0.4,0.6,0.8}},
  abtur/.style={ylabel={Mean Tool Calls},
                yticklabel style={xshift=2.6pt},
                ymin=0, ymax=6, ytick={0,2,4,6}},
}
\newcommand{\abrewplots}{%
  \addplot[abReward, line width=0.65pt] coordinates {(0,0.549) (1,0.534) (2,0.524) (3,0.514) (4,0.508) (5,0.516) (6,0.526) (7,0.530) (8,0.533) (9,0.547) (10,0.565) (11,0.582) (12,0.633) (13,0.658) (14,0.680) (15,0.688) (16,0.682) (17,0.686) (18,0.687) (19,0.670) (20,0.686) (21,0.700) (22,0.667) (23,0.679) (24,0.702) (25,0.687) (26,0.720) (27,0.739) (28,0.739) (29,0.734) (30,0.721) (31,0.694) (32,0.702) (33,0.707) (34,0.720) (35,0.760) (36,0.787) (37,0.777) (38,0.750) (39,0.742) (40,0.721)};
  \addplot[abEval, line width=0.5pt, densely dashed, opacity=0.62, mark=*, mark size=1.2pt,
           mark options={solid, opacity=1, fill=white, draw=abEval, line width=0.5pt}]
    coordinates {(10,0.575) (20,0.564) (30,0.562) (40,0.560)};
}
\newcommand{\abturplots}{\addplot[abTurns, line width=0.65pt] coordinates {(0,4.94) (1,4.96) (2,4.90) (3,4.77) (4,4.46) (5,4.46) (6,4.03) (7,3.76) (8,3.56) (9,3.38) (10,3.07) (11,2.84) (12,2.61) (13,2.37) (14,2.14) (15,1.90) (16,1.80) (17,1.63) (18,1.36) (19,1.19) (20,1.17) (21,1.15) (22,1.10) (23,1.16) (24,1.18) (25,1.16) (26,1.16) (27,1.17) (28,1.15) (29,1.13) (30,1.11) (31,1.08) (32,1.08) (33,1.05) (34,1.03) (35,1.04) (36,1.01) (37,0.97) (38,0.97) (39,0.96) (40,0.95)};
  \addplot[abEval, line width=0.5pt, densely dashed, opacity=0.62, mark=*, mark size=1.2pt,
           mark options={solid, opacity=1, fill=white, draw=abEval, line width=0.5pt}]
    coordinates {(10,2.83) (20,1.43) (30,1.33) (40,1.25)};}
\begin{groupplot}[group style={group name=ab, group size=4 by 1,
                               horizontal sep=1.30cm}, abpanel]
\nextgroupplot[abrew] \abrewplots
\nextgroupplot[abtur] \abturplots
\nextgroupplot[ylabel={Precision (AC) \%}, ymin=40, ymax=90,
               ytick={40,50,60,70,80,90}, xmin=0, xmax=84, xtick={0,40,80}]
  \addplot[abFpp0, line width=0.65pt] coordinates {(0,47.8) (1,45.1) (2,45.8) (3,45.6) (4,44.5) (5,44.0) (6,43.2) (7,44.0) (8,44.3) (9,46.4) (10,46.2) (11,45.2) (12,47.7) (13,47.6) (14,45.9) (15,44.8) (16,44.7) (17,46.0) (18,44.6) (19,45.6) (20,44.8) (21,43.0) (22,44.1) (23,45.8) (24,47.1) (25,44.8) (26,46.2) (27,43.3) (28,44.3) (29,43.2) (30,44.4) (31,43.7) (32,42.8) (33,45.0) (34,44.4) (35,43.7) (36,44.7) (37,45.8) (38,48.5) (39,49.3) (40,50.0) (41,51.3) (42,51.5) (43,50.8) (44,49.9) (45,50.4) (46,49.4) (47,49.9) (48,48.6) (49,48.1) (50,47.6) (51,47.6) (52,47.3) (53,47.5) (54,46.9) (55,47.9) (56,51.0) (57,50.8) (58,49.5) (59,50.4) (60,49.3) (61,48.2) (62,45.9) (63,47.3) (64,48.7) (65,49.0) (66,47.4) (67,50.6) (68,51.7) (69,54.5) (70,54.4) (71,53.6) (72,55.8) (73,56.2) (74,57.1) (75,59.3) (76,61.5) (77,63.2) (78,63.6) (79,63.7) (80,61.6)};
  \addplot[abFpp0, line width=0.5pt, densely dashed, opacity=0.62, mark=o, mark size=1.6pt,
           mark options={solid, opacity=1, draw=abFpp0, line width=0.7pt}]
    coordinates {(20,65.00) (40,64.09) (60,69.39) (80,74.62)};
  \addplot[abFpp5, line width=0.5pt, densely dashed, opacity=0.62, mark=o, mark size=1.6pt,
           mark options={solid, opacity=1, draw=abFpp5, line width=0.7pt}]
    coordinates {(20,66.67) (40,65.00) (60,77.87) (80,77.95)};
  \addplot[abFpp5, line width=0.65pt] coordinates {(0,51.7) (1,48.4) (2,47.5) (3,46.3) (4,45.5) (5,44.3) (6,42.9) (7,44.5) (8,44.8) (9,46.7) (10,46.9) (11,46.6) (12,47.9) (13,47.8) (14,46.3) (15,45.8) (16,45.4) (17,46.4) (18,45.0) (19,45.6) (20,45.2) (21,46.4) (22,47.3) (23,47.8) (24,48.6) (25,47.2) (26,47.0) (27,44.5) (28,44.4) (29,42.6) (30,42.2) (31,44.9) (32,44.8) (33,45.9) (34,45.3) (35,42.8) (36,46.4) (37,48.4) (38,49.5) (39,51.9) (40,52.1) (41,52.3) (42,52.2) (43,52.4) (44,52.9) (45,52.5) (46,52.4) (47,52.2) (48,51.2) (49,51.9) (50,51.9) (51,53.4) (52,53.7) (53,55.6) (54,53.1) (55,54.7) (56,57.8) (57,57.3) (58,55.9) (59,58.0) (60,57.8) (61,57.5) (62,55.2) (63,56.1) (64,58.4) (65,59.6) (66,58.8) (67,61.0) (68,61.9) (69,62.9) (70,63.4) (71,61.8) (72,64.0) (73,64.9) (74,65.6) (75,68.0) (76,67.9) (77,69.8) (78,70.6) (79,71.0) (80,67.3)};
\nextgroupplot[ylabel={Recall (AC) \%}, ymin=40, ymax=90,
               ytick={40,50,60,70,80,90}, xmin=0, xmax=84, xtick={0,40,80}]
  \addplot[abFpp0, line width=0.65pt] coordinates {(0,61.0) (1,56.6) (2,54.2) (3,52.9) (4,51.2) (5,51.9) (6,51.2) (7,52.4) (8,53.0) (9,53.2) (10,55.9) (11,57.6) (12,57.7) (13,59.3) (14,61.2) (15,61.7) (16,61.3) (17,64.5) (18,63.6) (19,66.5) (20,67.6) (21,65.8) (22,65.5) (23,67.8) (24,69.0) (25,68.3) (26,69.7) (27,68.4) (28,66.9) (29,67.2) (30,69.6) (31,70.1) (32,68.5) (33,69.2) (34,69.9) (35,69.0) (36,70.0) (37,71.8) (38,73.3) (39,72.8) (40,74.9) (41,76.3) (42,75.7) (43,75.5) (44,75.0) (45,75.2) (46,77.0) (47,76.9) (48,76.4) (49,76.7) (50,78.5) (51,80.1) (52,77.9) (53,77.9) (54,76.4) (55,77.8) (56,78.2) (57,79.7) (58,77.9) (59,78.6) (60,79.4) (61,79.2) (62,76.9) (63,78.5) (64,79.8) (65,81.8) (66,82.6) (67,84.2) (68,83.3) (69,82.8) (70,80.9) (71,82.2) (72,81.3) (73,82.5) (74,81.6) (75,82.0) (76,83.7) (77,83.4) (78,84.6) (79,83.1) (80,82.4)};
  \addplot[abFpp0, line width=0.5pt, densely dashed, opacity=0.62, mark=o, mark size=1.6pt,
           mark options={solid, opacity=1, draw=abFpp0, line width=0.7pt}]
    coordinates {(20,72.22) (40,72.22) (60,82.93) (80,85.35)};
  \addplot[abFpp5, line width=0.5pt, densely dashed, opacity=0.62, mark=o, mark size=1.6pt,
           mark options={solid, opacity=1, draw=abFpp5, line width=0.7pt}]
    coordinates {(20,75.81) (40,72.22) (60,77.87) (80,80.17)};
  \addplot[abFpp5, line width=0.65pt] coordinates {(0,55.8) (1,52.7) (2,51.0) (3,48.7) (4,48.1) (5,48.0) (6,48.5) (7,50.5) (8,51.5) (9,52.0) (10,55.4) (11,57.9) (12,59.6) (13,61.0) (14,62.6) (15,63.0) (16,62.7) (17,65.3) (18,65.5) (19,66.3) (20,68.0) (21,68.0) (22,69.8) (23,69.4) (24,69.3) (25,68.6) (26,68.9) (27,68.7) (28,66.5) (29,64.6) (30,66.8) (31,68.8) (32,68.5) (33,69.3) (34,70.1) (35,68.2) (36,70.6) (37,71.8) (38,73.5) (39,73.3) (40,74.3) (41,73.1) (42,71.6) (43,70.8) (44,70.4) (45,70.0) (46,71.3) (47,70.9) (48,69.7) (49,69.8) (50,72.7) (51,74.2) (52,71.9) (53,72.2) (54,67.2) (55,69.3) (56,71.2) (57,71.4) (58,68.5) (59,69.4) (60,70.6) (61,71.0) (62,69.0) (63,71.1) (64,71.4) (65,76.0) (66,76.3) (67,78.0) (68,76.0) (69,73.6) (70,72.3) (71,73.2) (72,73.2) (73,73.9) (74,74.0) (75,74.4) (76,76.0) (77,75.2) (78,76.3) (79,74.9) (80,74.0)};
\end{groupplot}

\path let \p1=(ab c2r1.north east), \p2=(ab c3r1.north west) in
  coordinate (abmidT) at ($(\x1,\y1)!0.5!(\x2,\y1)$);
\path let \p1=(ab c2r1.south east), \p2=(ab c3r1.south west) in
  coordinate (abmidB) at ($(\x1,\y1)!0.5!(\x2,\y1)$);
\draw[gray!45, line width=0.4pt]
  ($(abmidT)+(-0.26,0.30)$) -- ($(abmidB)+(-0.26,-0.62)$);

\node[anchor=south, yshift=0.24cm]
  at ($(ab c1r1.north)!0.5!(ab c2r1.north)$) {\tiny
    \textcolor{abReward}{\rule[0.22em]{0.34cm}{0.6pt}}~train\quad
    \textcolor{abEval}{\raisebox{0.15ex}{$\circ$}}~validation};
\node[anchor=south, yshift=0.24cm]
  at ($(ab c3r1.north)!0.5!(ab c4r1.north)$) {\tiny
    \textcolor{abFpp0}{\rule[0.16em]{0.13cm}{0.13cm}}~$\lambda\,{=}\,0$\quad
    \textcolor{abFpp5}{\rule[0.16em]{0.13cm}{0.13cm}}~$\lambda\,{=}\,0.5$\qquad
    \textcolor{black!62}{\rule[0.22em]{0.34cm}{0.6pt}}~train\quad
    \textcolor{black!62}{\raisebox{0.15ex}{$\circ$}}~validation};
\end{tikzpicture}}
\vspace{-11pt}
\caption{Reward ablations. \textbf{Left}: reward, verdict accuracy, and tool use
under verdict-only training. \textbf{Right}: \emph{Accepted} precision and recall for
false-positive penalties $\lambda=0$ and $0.5$.}
\label{fig:label-only-trap}
\end{figure}

%% file: sections/body/benchmarks.tex
\section{Evaluation Benchmarks}
\label{sec:verifier-data}

\subsection{\textsc{UOJ-Bench}}
\label{sec:uoj-bench}

\textsc{UOJ-Bench}~\citep{uojbench2026} evaluates code generation, hacking,
and repair using real submissions to the Universal Online Judge
(UOJ)\footnote{\url{https://uoj.ac/}}. We use its
code-hacking splits: 479 Easy submissions with faults exposed by standard tests,
and 1,046 Hard submissions that passed those tests before a community hack
exposed a fault. All are faulty. We re-host the released candidates on our UOJ
instance and reproduce their original outcomes
(Appendices~\ref{app:infra} and~\ref{app:uojbench}). These splits test
counterexample discovery on
overt and covert faults, but leave acceptance untested. A verifier could predict
\emph{unaccepted} for every candidate and obtain perfect binary verdict
accuracy without finding a counterexample.

\subsection{\textsc{USACO-Judge}}
\label{sec:usaco-hack}

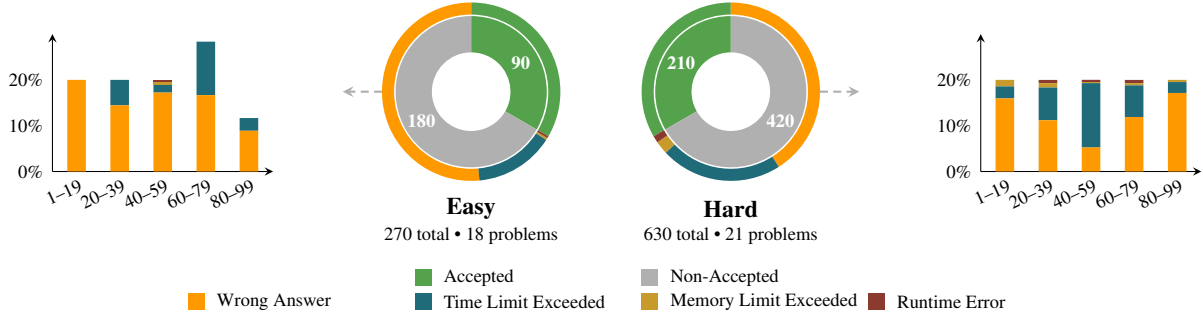
\begin{figure}[H]
  \centering
  \definecolor{cAC}{HTML}{54A24B}%
  \definecolor{cNonAC}{HTML}{B0B0B0}%
  \definecolor{cWA}{HTML}{FF9900}%
  \definecolor{cTLE}{HTML}{206B7A}%
  \definecolor{cMLE}{HTML}{C79A2B}%
  \definecolor{cRE}{HTML}{8B3A2F}%
  \newcommand{\dseg}[8]{%
    \fill[#7] ([shift={(#5:#4)}] #1,#2) arc[start angle=#5, end angle=#6, radius=#4]
      -- ([shift={(#6:#3)}] #1,#2) arc[start angle=#6, end angle=#5, radius=#3]
      -- cycle;
    \pgfmathsetmacro{\segmid}{(#5+#6)/2}%
    \pgfmathsetmacro{\seglr}{(#3+#4)/2}%
    \node[white,font=\scriptsize\bfseries]
      at ([shift={(\segmid:\seglr)}] #1,#2) {#8};}
  \pgfplotsset{
    tier bar/.style={
      ybar stacked, bar width=7pt,
      width=4.5cm, height=3.4cm,
      symbolic x coords={1--19,20--39,40--59,60--79,80--99},
      xtick=data, x tick label style={font=\scriptsize, rotate=25, anchor=north east, xshift=3pt, yshift=1pt},
      ymin=0, ymax=30,
      ytick={0,10,20}, yticklabels={0\%,10\%,20\%},
      y tick label style={font=\scriptsize},
      ytick style={draw=none},
      xtick style={draw=none},
      every axis plot/.append style={draw=none},
      axis lines*=left,
      axis line style={draw=black, -stealth},
      enlarge x limits=0.14,
    },
  }

  \newsavebox{\easybar}
  \savebox{\easybar}{\begin{tikzpicture}
    \begin{axis}[tier bar]
      \addplot[fill=cWA]  coordinates {(1--19,20.00)(20--39,14.44)(40--59,17.22)(60--79,16.67)(80--99,8.89)};
      \addplot[fill=cTLE] coordinates {(1--19,0.00)(20--39,5.56)(40--59,1.67)(60--79,11.67)(80--99,2.78)};
      \addplot[fill=cMLE] coordinates {(1--19,0.00)(20--39,0.00)(40--59,0.56)(60--79,0.00)(80--99,0.00)};
      \addplot[fill=cRE]  coordinates {(1--19,0.00)(20--39,0.00)(40--59,0.56)(60--79,0.00)(80--99,0.00)};
    \end{axis}
  \end{tikzpicture}}
  \newsavebox{\easydonut}
  \savebox{\easydonut}{\begin{tikzpicture}
    \dseg{0}{0}{1.0}{0.52}{-30}{90}{cAC}{90}
    \dseg{0}{0}{1.0}{0.52}{90}{330}{cNonAC}{180}
    \dseg{0}{0}{1.18}{1.02}{-30}{90}{cAC}{}
    \dseg{0}{0}{1.18}{1.02}{90}{275.333}{cWA}{}
    \dseg{0}{0}{1.18}{1.02}{275.333}{327.333}{cTLE}{}
    \dseg{0}{0}{1.18}{1.02}{327.333}{328.667}{cMLE}{}
    \dseg{0}{0}{1.18}{1.02}{328.667}{330}{cRE}{}
    \node[font=\small\bfseries] at (0,-1.55) {Easy};
    \node[font=\scriptsize]    at (0,-1.9)  {270 total \textbullet{} 18 problems};
    \draw[cNonAC,thick,densely dashed,->,>=stealth] (-1.18,0) -- (-1.7,0);
  \end{tikzpicture}}
  \newsavebox{\harddonut}
  \savebox{\harddonut}{\begin{tikzpicture}
    \dseg{0}{0}{1.0}{0.52}{210}{90}{cAC}{210}
    \dseg{0}{0}{1.0}{0.52}{90}{-150}{cNonAC}{420}
    \dseg{0}{0}{1.18}{1.02}{210}{90}{cAC}{}
    \dseg{0}{0}{1.18}{1.02}{90}{-57.429}{cWA}{}
    \dseg{0}{0}{1.18}{1.02}{-57.429}{-136.857}{cTLE}{}
    \dseg{0}{0}{1.18}{1.02}{-136.857}{-145.429}{cMLE}{}
    \dseg{0}{0}{1.18}{1.02}{-145.429}{-150}{cRE}{}
    \node[font=\small\bfseries] at (0,-1.55) {Hard};
    \node[font=\scriptsize]    at (0,-1.9)  {630 total \textbullet{} 21 problems};
    \draw[cNonAC,thick,densely dashed,->,>=stealth] (1.18,0) -- (1.7,0);
  \end{tikzpicture}}
  \newsavebox{\hardbar}
  \savebox{\hardbar}{\begin{tikzpicture}
    \begin{axis}[tier bar]
      \addplot[fill=cWA]  coordinates {(1--19,15.95)(20--39,11.19)(40--59,5.24)(60--79,11.90)(80--99,17.14)};
      \addplot[fill=cTLE] coordinates {(1--19,2.62)(20--39,7.14)(40--59,14.05)(60--79,6.90)(80--99,2.38)};
      \addplot[fill=cMLE] coordinates {(1--19,1.43)(20--39,0.95)(40--59,0.24)(60--79,0.48)(80--99,0.48)};
      \addplot[fill=cRE]  coordinates {(1--19,0.00)(20--39,0.71)(40--59,0.48)(60--79,0.71)(80--99,0.00)};
    \end{axis}
  \end{tikzpicture}}
  \begin{minipage}{\linewidth}
    \centering
    \scalebox{1.00}{%
      \raisebox{-0.5\height}{\usebox{\easybar}}\hspace{0.9cm}%
      \raisebox{-0.5\height}{\usebox{\easydonut}}\hspace{0.9cm}%
      \raisebox{-0.5\height}{\usebox{\harddonut}}\hspace{0.9cm}%
      \raisebox{-0.5\height}{\usebox{\hardbar}}}%
  \end{minipage}
  \\[2pt]
  \begin{tikzpicture}
    \foreach \c/\t/\k [count=\i] in {%
      cAC/{Accepted}/1,
      cNonAC/{Non-Accepted}/2%
    }{%
      \pgfmathsetmacro{\xpos}{\k*3.0}%
      \fill[\c] (\xpos,0.32) rectangle ++(0.22,0.22);
      \node[anchor=west,font=\scriptsize] at (\xpos+0.26,0.43) {\t};%
    }
    \foreach \c/\t/\k [count=\i] in {%
      cWA/{Wrong Answer}/0,
      cTLE/{Time Limit Exceeded}/1,
      cMLE/{Memory Limit Exceeded}/2,
      cRE/{Runtime Error}/3%
    }{%
      \pgfmathsetmacro{\xpos}{\k*3.0}%
      \fill[\c] (\xpos,0) rectangle ++(0.22,0.22);
      \node[anchor=west,font=\scriptsize] at (\xpos+0.26,0.11) {\t};%
    }
  \end{tikzpicture}
  \caption{\textsc{USACO-Judge} composition by difficulty, verdict, failure type,
  and official score band.}
  \label{fig:composition}
\end{figure}

To evaluate acceptance judgments alongside counterexample discovery, we build
\textsc{USACO-Judge} with both accepted and unaccepted programs. The benchmark
contains 900 C++17 candidates for 39 held-out problems
from the USACO 2026 contests, spanning Bronze through
Platinum.\footnote{\url{https://usaco.org/index.php?page=training}} The
easy tier contains 270 candidates over 18 problems from the Bronze and Silver divisions, and
the hard tier contains 630 candidates over 21 problems from the Gold and Platinum divisions,
including problems from the USACO Open. Candidates come from five LLMs:
Claude Opus 4.8~\citep{anthropic2026claudeopus48},
GPT-5.5~\citep{openai2026gpt55},
GPT-OSS-120B~\citep{openai2025gptoss120bgptoss20bmodel},
DeepSeek-V4~\citep{deepseekai2026deepseekv4highlyefficientmilliontoken}, and
Qwen3.6-35B-A3B~\citep{qwen36_35b_a3b}, as well as five human competitive programmers. Both tiers use a 1:2 ratio of accepted to unaccepted candidates. Within each tier, we sample the same number of candidates for every problem and
distribute the unaccepted candidates across five official score bands, from programs that fail most tests to near-misses. The
hard tier is balanced across bands. We also select for distinct code structures
and runtime profiles within each problem and score band, reducing repeated
variants of the same implementation. High-scoring unaccepted submissions are
scarce in the easy tier, so its unfilled 80--99 quota is assigned to the 60--79 band,
which therefore exceeds 20\%. Figure~\ref{fig:composition} summarizes the
resulting distribution.

The USACO website provides the problem statements, test data, and official
solutions, but not the validators and checkers required to deploy the problems
on our own online judge. We employ experienced competitive programmers to implement validators
compatible with UOJ for every problem, allowing the judge
to verify that each generated input satisfies the problem constraints. For
output judging, we use UOJ's native checker whenever possible and write custom
checkers for the six problems that allow multiple valid outputs. Every
validator and custom checker is verified end to end on the deployed online
judge before benchmark construction. We then rejudge every candidate on the
official test data, so every label is an actual judge verdict, and retain it
only if its original score is reproduced. Multiple runs ensure that judge
variability does not change a candidate's accepted or unaccepted status. See
Appendix~\ref{app:usaco-judge-construction} for candidate construction and
Appendix~\ref{app:validators-checkers} for validator and checker details.

\FloatBarrier

%% file: sections/body/experiments.tex
\section{Experiments}
\label{sec:verifier-results}

We evaluate \textsc{Arbiter}'s verification quality on the two benchmarks of
Section~\ref{sec:verifier-data} (Section~\ref{sec:exp-verification}), then
test its use for code refinement (Section~\ref{sec:exp-refine}) and selection
(Section~\ref{sec:exp-selection}).

\subsection{Verification Quality}
\label{sec:exp-verification}

Although \textsc{Arbiter} predicts a detailed verdict, the failure type from an
online judge depends on the first failed test and need not describe every fault.
We map predicted and reference verdicts to \emph{Accepted} or \emph{unaccepted}.
\textbf{Verdict accuracy} (Verdict Acc.) measures this binary decision. \textbf{Hack
success} (Hack Succ.) is the fraction of truly \emph{unaccepted} candidates
broken by a valid generated input under the checker, providing an executable
measure of counterexample quality. On the \textsc{UOJ-Bench} hacking splits,
verdict accuracy is
unaccepted-class recall because every target is faulty; \textsc{USACO-Judge}
tests both sides of the decision. The benchmark's AC/non-AC split follows the
official tests; for evaluation, an officially \emph{Accepted} candidate is
treated as unaccepted when a validated counterexample exposes a failure.
Table~\ref{tab:results} reports multiple-run averages for all baselines and
\textsc{Arbiter}.

\begin{table}[H]
\centering
\renewcommand{\arraystretch}{1.2}
\newcolumntype{R}{>{\centering\arraybackslash}p{0.85cm}}
\newcolumntype{O}{>{\centering\arraybackslash}p{1.8cm}}
\newcolumntype{P}{>{\color{black!48}}r}
\resizebox{\linewidth}{!}{%
\begin{tabular}{@{}l !{\color{gray!50}\vrule} P !{\color{gray!50}\vrule} RR!{\color{gray!50}\vrule}RR!{\color{gray!70}\vrule}RR!{\color{gray!50}\vrule}RR!{\color{gray!70}\vrule}O@{}}
\toprule
\multirow{3}{*}[-1ex]{\textbf{Model}} & \multirow{3}{*}[-1ex]{\textcolor{black!48}{\textbf{Params}}} & \multicolumn{4}{c!{\color{gray!70}\vrule}}{\textbf{\textsc{UOJ-Bench} (Hacking)}} & \multicolumn{4}{c!{\color{gray!70}\vrule}}{\textbf{\textsc{USACO-Judge}}} & \multirow{3}{=}[-1ex]{\centering\textbf{Overall}\\\textbf{Hack Succ.}} \\
\cmidrule(lr){3-6}\cmidrule(lr){7-10}
& & \multicolumn{2}{c!{\color{gray!50}\vrule}}{\logo[0.9em]{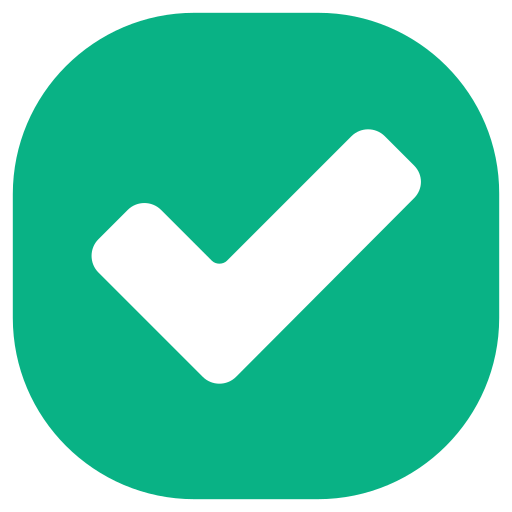}~Verdict Acc.} & \multicolumn{2}{c!{\color{gray!70}\vrule}}{\logo[0.9em]{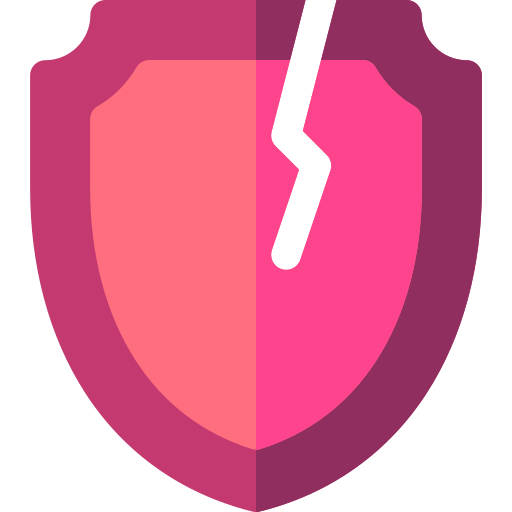}~Hack Succ.}
& \multicolumn{2}{c!{\color{gray!50}\vrule}}{\logo[0.9em]{check.png}~Verdict Acc.} & \multicolumn{2}{c!{\color{gray!70}\vrule}}{\logo[0.9em]{broken-shield.png}~Hack Succ.} & \\
\cmidrule(lr){3-4}\cmidrule(lr){5-6}\cmidrule(lr){7-8}\cmidrule(lr){9-10}
& & {\small Easy} & {\small Hard} & {\small Easy} & {\small Hard} & {\small Easy} & {\small Hard} & {\small Easy} & {\small Hard} & \\
\midrule
\multicolumn{11}{c}{{\scriptsize\textcolor{gray}{\faBolt}}~\textit{Non-agentic}} \\
\midrule
\logo{qwen.png}~Qwen3.6-35B-A3B    & 35B & 86.85 & 87.57 & 16.70 & 10.85 & 66.42 & 64.55 & 32.04 & 21.65 & 16.10~ \\
\logo{openai.png}~GPT-OSS-120B & 120B & 79.12 & 77.06 & 29.09 & 20.07 & 60.74 & 66.88 & 33.89 & 36.84 & 26.59~ \\
\logo{deepseek.png}~DeepSeek-V4-Flash-Preview & 284B & 73.49 & 73.10 & 33.49 & 22.03 & 74.94 & 68.68 & 55.00 & 41.52 & 31.26~ \\
\logo{qwen.png}~Qwen3.5-397B-A17B & 397B & \su{88.87} & \su{88.34} & 41.68 & 30.78 & 85.31 & 74.23 & 77.31 & 58.88 & 42.73~ \\
\logo{zai.png}~GLM-5.2 & 753B & 80.17 & 76.77 & 65.34 & 57.33 & 94.07 & 80.63 & 91.16 & \hi{79.03} & 66.29~ \\
\logo{kimi.png}~Kimi-K2.5 & 1T & 72.44 & 68.67 & 46.76 & 30.95 & 83.33 & 72.86 & 70.56 & 57.24 & 43.07~ \\
\logo{deepseek.png}~DeepSeek-V4-Pro-Preview & 1.6T & 76.06 & 75.46 & 51.15 & 38.94 & 90.74 & 82.86 & 84.85 & 70.95 & 51.91~ \\
\logo{claude.png}~Claude-Opus-4.8 & -- & 84.97 & 75.72 & \su{77.04} & \su{67.06} & \hi{\su{94.81}} & 82.38 & \hi{\su{94.48}} & \su{87.61} & \su{75.69}~ \\
\logo{openai.png}~GPT-5.5 & -- & \hi{\tb{91.86}} & \hi{\tb{89.87}} & \hi{\tb{78.71}} & \hi{\tb{77.01}} & \hi{94.44} & \hi{\tb{90.79}} & \hi{90.66} & \hi{\tb{91.97}} & \hi{\tb{81.51}~} \\
\midrule
\multicolumn{11}{c}{{\scriptsize\textcolor{gray}{\faCog}}~\textit{Agentic}} \\
\midrule
\logo{qwen.png}~Qwen3.6-35B-A3B    & 35B & 71.82 & 67.43 & 41.34 & 31.47 & 85.56 & 65.34 & 79.44 & 45.88 & 40.61~ \\
\logo{deepseek.png}~DeepSeek-V4-Flash-Preview & 284B & 65.55 & 63.19 & 56.37 & \hi{49.35} & \hi{\tb{96.30}} & 77.62 & 92.78 & 65.27 & 57.76~ \\
\midrule
\rowcolor{arbRow!6}
\logo{judgement.png}~\textsc{Arbiter} {\small(Ours)} & 35B & \hi{87.61} & \hi{83.40} & \hi{61.48} & \hi{50.66} & \hi{94.20} & \su{84.50} & \hi{\tb{96.64}} & \hi{79.38} & \hi{62.67~} \\
\bottomrule
\end{tabular}%
}
\caption{Results on \textsc{UOJ-Bench} (Hacking) and \textsc{USACO-Judge}, reported by difficulty split
(Easy / Hard) and grouped by inference mode: non-agentic ({\textcolor{gray}{\faBolt}},
single-shot) versus agentic ({\textcolor{gray}{\faCog}}, tool-using).}
\label{tab:results}
\end{table}

\begingroup
\setlength{\floatsep}{5pt plus 1pt minus 1pt}
\input{sections/figures/hack_test_time_scaling}
\endgroup

\paragraph{Verification performance.}
Training raises overall hack success from 40.61\% to 62.67\%. On the hard
splits, verdict accuracy rises from 67.43\% to 83.40\% on \textsc{UOJ-Bench}
and from 65.34\% to 84.50\% on \textsc{USACO-Judge}, a 26.5\% relative
improvement on average. The resulting 35B \textsc{Arbiter} outperforms
non-agentic models as large as 1.6T parameters. On \textsc{USACO-Judge},
\textsc{Arbiter} achieves the highest reported hack success among the evaluated
open-weight models on both Easy and Hard, and the highest reported verdict
accuracy on Hard.

\paragraph{Test-time scaling.}
All systems in Table~\ref{tab:results} use one verifier sample per candidate. To
isolate backbone scaling, Figure~\ref{fig:hack-at-k} varies the sample budget for
Qwen3.6-35B-A3B without tools, with the agentic scaffold, and after
\textsc{Arbiter} training. Repeated sampling improves all three variants, while
one \textsc{Arbiter} sample already exceeds eight no-tools samples on every split.

\paragraph{Counterexamples track diagnosis quality, and correct diagnoses are
specific and actionable.}
We manually audit 400 \textsc{Arbiter} diagnoses on USACO candidates, with three
competitive programmers comparing each diagnosis against the problem, candidate,
and official judge evidence. Their independent initial labels agree on all 400
cases. Among compilable candidates,
counterexamples succeed for 71.6\% of correct diagnoses and 10.3\% of incorrect
or vacuous ones (Figure~\ref{fig:diagnosis-audit-main}, left). Among the 163
correct diagnoses that identify a real bug, 82.2\% locate its cause and 54.0\%
suggest a fix (Figure~\ref{fig:diagnosis-audit-main}, right).
Appendix~\ref{app:diagnosis-proxy} gives the full audit and sampling protocol.

\input{sections/figures/diagnosis_audit}
\vspace{-4pt}
\subsection{Code Refinement}
\label{sec:exp-refine}

For each of the 39 held-out USACO 2026 problems, we test Qwen3.6-35B-A3B and
GPT-OSS-120B using eight shared, already judged starting programs and up to
eight revisions. We compare previous code alone, online-judge feedback
(verdict and subtask score), and three combinations of \textsc{Arbiter}'s verdict
($\mathbf{V}$), diagnosis ($\mathbf{D}$), and counterexample generator
($\mathbf{X}$). Round 0 is excluded from the additional-submission budget;
every arm retains a 100-point seed without further submissions. Otherwise,
only Online Judge uses the round-0 result as feedback, while \textsc{Arbiter} reviews
each version. Best@1 retains round 8 for previous code, round 1 for Online
Judge, and the first \textsc{Arbiter}-predicted \emph{Accepted} (or round 8).
Best@8 retains the highest score among up to eight additional submissions;
Online Judge stops at 100 points. Table~\ref{tab:refine} reports the 30
non-Bronze problems; Appendix~\ref{app:refine-detail} gives the full protocol.

\input{sections/tables/refinement_results}

At Best@8, the $\mathbf{V}+\mathbf{D}$ condition scores above Online Judge
for both solvers at every difficulty tier. Both $\mathbf{V}+\mathbf{D}$ and
$\mathbf{V}+\mathbf{X}$ expose the verdict and use the same acceptance-based
stopping rule; their comparison favors diagnosis as repair guidance. Adding
counterexample-generator source $\mathbf{X}$
to $\mathbf{V}+\mathbf{D}$ brings no consistent gain
(Appendix~\ref{app:refine-x}). At Best@1, \textsc{Arbiter}'s predicted
\emph{Accepted} verdict determines which version is retained; false acceptances
remain in the reported scores (Appendix~\ref{app:refine-verifier}).

\paragraph{Case study: IOI 2026.}
On all six IOI 2026 problems, GPT-OSS-120B starts
50 internal trajectories per refinement target and makes up to five
\textsc{Arbiter}-guided revisions. Submissions are counted per problem, not per
target: four tasks combine programs by trajectory index, while \emph{Classroom
Game} and \emph{Partition} divide the official 50-slot cap among separate
programs. \textsc{Arbiter} reviews use no slots; each online-judge feedback
query uses one. Under this cap, expected scores are 385.46 with
\textsc{Arbiter} (above the 361.12 gold threshold) and 334.6 with
online-judge feedback
(Figure~\ref{fig:ioi-climb}). Appendix~\ref{app:ioi-protocol} gives the budget
and score accounting.

\subsection{Code Selection}
\label{sec:exp-selection}

Following \citet{agenticverifier2026}, we regenerate 8- and 64-program pools
from Qwen3-30B-A3B-Thinking and Qwen3-235B-A22B-Thinking~\citep{yang2025qwen3technicalreport}
on LiveCodeBench v6~\citep{jain2025livecodebench} and OJBench~\citep{ojbench2025}.
Repeated uniform selection closely recovers published pass@$1$
(Appendix~\ref{app:selection}). We run one or eight \textsc{Arbiter} calls per
candidate, then select uniformly among those with the most \emph{Accepted}
verdicts, or from the full pool if none is accepted. Table~\ref{tab:selection}
places our results and perfect-selection upper bounds on these pools beside
published baselines.

\input{sections/tables/selection_results}

Across all eight settings, one \textsc{Arbiter} call per candidate has higher
reported pass@$1$ than the strongest published baseline. With the 30B solver
and a pool of 64, \textsc{Arbiter} scores 82.6 versus the published 73.7 on
LiveCodeBench and 43.6 versus 34.9 on OJBench. Eight calls improve every
setting further, moving selection consistently closer to the perfect-selection
upper bound.

\FloatBarrier

%% file: sections/figures/hack_test_time_scaling.tex
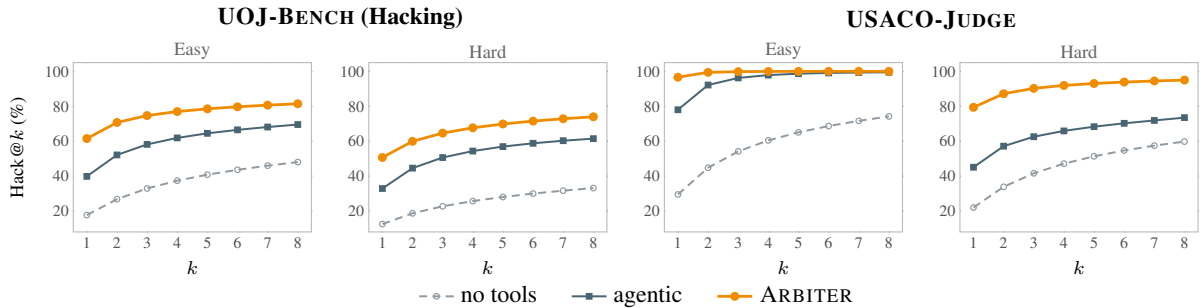
\begin{figure}[H]
\centering
\setlength{\abovecaptionskip}{3pt}
\begin{tikzpicture}
\definecolor{hkNon}{HTML}{929AA1}%
\definecolor{hkAg}{HTML}{496574}%
\definecolor{hkOurs}{HTML}{F08C00}%
\tikzset{
  hkline/.style={line width=0.75pt},
  hkstyleNon/.style={hkline, hkNon, densely dashed, mark=o, mark size=1.05pt,
                     mark options={solid, fill=white, draw=hkNon, line width=0.45pt}},
  hkstyleAg/.style={hkline, hkAg, mark=square*, mark size=0.9pt,
                    mark options={solid, fill=hkAg, draw=none}},
  hkstyleOurs/.style={line width=1.05pt, hkOurs, mark=*, mark size=1.15pt,
                      mark options={solid, fill=hkOurs, draw=none}},
}
\pgfplotsset{
  hkpanel/.style={
    width=3.15cm, height=2.24cm, scale only axis,
    xmin=0.55, xmax=8.45,
    xtick={1,2,3,4,5,6,7,8},
    ymin=8, ymax=105, ytick={20,40,60,80,100},
    axis lines=box,
    axis line style={black!28, line width=0.4pt},
    enlarge x limits=false, enlarge y limits=false,
    major tick length=0pt,
    tick label style={font=\tiny, text=black!68,
                      /pgf/number format/assume math mode=true},
    xticklabel style={yshift=1.6pt},
    yticklabel style={xshift=1.8pt},
    label style={font=\scriptsize},
    xlabel={$k$},
    xlabel style={font=\scriptsize, yshift=2pt},
    title style={font=\scriptsize, text=black!68, align=center, yshift=-8pt},
    grid=none,
    axis background/.style={fill=white},
  },
}
\begin{groupplot}[
  group style={group name=hk, group size=4 by 1, horizontal sep=0.76cm},
  hkpanel,
]

\nextgroupplot[title={\textcolor{black!60}{Easy}}, ylabel={Hack@$k$ (\%)}]
  \addplot[hkstyleNon] coordinates
    {(1,17.64) (2,26.77) (3,32.90) (4,37.37) (5,40.81) (6,43.60) (7,45.96) (8,48.02)};
  \addplot[hkstyleAg] coordinates
    {(1,39.82) (2,52.11) (3,58.15) (4,61.87) (5,64.48) (6,66.48) (7,68.11) (8,69.52)};
  \addplot[hkstyleOurs] coordinates
    {(1,61.48) (2,70.71) (3,74.66) (4,76.96) (5,78.52) (6,79.70) (7,80.64) (8,81.42)};

\nextgroupplot[title={\textcolor{black!60}{Hard}}]
  \addplot[hkstyleNon] coordinates
    {(1,12.51) (2,18.68) (3,22.70) (4,25.67) (5,28.02) (6,29.98) (7,31.65) (8,33.13)};
  \addplot[hkstyleAg] coordinates
    {(1,32.86) (2,44.55) (3,50.61) (4,54.33) (5,56.89) (6,58.78) (7,60.27) (8,61.45)};
  \addplot[hkstyleOurs] coordinates
    {(1,50.66) (2,59.91) (3,64.63) (4,67.66) (5,69.84) (6,71.50) (7,72.84) (8,73.97)};

\nextgroupplot[title={\textcolor{black!60}{Easy}}]
  \addplot[hkstyleNon] coordinates
    {(1,29.46) (2,44.76) (3,54.13) (4,60.46) (5,65.06) (6,68.66) (7,71.63) (8,74.18)};
  \addplot[hkstyleAg] coordinates
    {(1,78.01) (2,92.17) (3,96.23) (4,97.86) (5,98.66) (6,99.08) (7,99.32) (8,99.45)};
  \addplot[hkstyleOurs] coordinates
    {(1,96.64) (2,99.42) (3,99.83) (4,99.95) (5,99.99) (6,100.00) (7,100.00) (8,100.00)};

\nextgroupplot[title={\textcolor{black!60}{Hard}}]
  \addplot[hkstyleNon] coordinates
    {(1,21.98) (2,33.84) (3,41.59) (4,47.11) (5,51.32) (6,54.68) (7,57.44) (8,59.77)};
  \addplot[hkstyleAg] coordinates
    {(1,45.07) (2,57.13) (3,62.55) (4,65.88) (5,68.30) (6,70.23) (7,71.87) (8,73.49)};
  \addplot[hkstyleOurs] coordinates
    {(1,79.38) (2,87.17) (3,90.26) (4,91.94) (5,93.04) (6,93.86) (7,94.49) (8,95.00)};

\end{groupplot}

\node[anchor=south, font=\small\bfseries, yshift=0.32cm]
  at ($(hk c1r1.north)!0.5!(hk c2r1.north)$) {\textsc{UOJ-Bench} (Hacking)};
\node[anchor=south, font=\small\bfseries, yshift=0.32cm]
  at ($(hk c3r1.north)!0.5!(hk c4r1.north)$) {\textsc{USACO-Judge}};

\newcommand{\hkswatch}[1]{%
  \tikz[baseline=-0.55ex]{%
    \draw[#1] (0,0) -- (0.46,0);
    \draw[#1, only marks] plot coordinates {(0.23,0)};}}
\node[anchor=north, yshift=-0.56cm] at ($(hk c1r1.south)!0.5!(hk c4r1.south)$) {%
  \footnotesize
  \begin{tabular}{@{}c@{\hspace{1.4em}}c@{\hspace{1.4em}}c@{}}
    \hkswatch{hkstyleNon}~no tools &
    \hkswatch{hkstyleAg}~agentic &
    \hkswatch{hkstyleOurs}~\textsc{Arbiter} \\
  \end{tabular}};

\end{tikzpicture}
\caption{Test-time scaling on the Qwen3.6-35B-A3B backbone. We draw eight samples
per candidate and report Hack@$k$, the expected success rate of a uniformly chosen
subset of $k$ samples.}
\label{fig:hack-at-k}
\end{figure}

%% file: sections/figures/diagnosis_audit.tex
\begin{figure}[t]
  \centering
  \definecolor{diagOrange}{HTML}{C9823D}
  \definecolor{diagSlate}{HTML}{7D939B}
  \begin{tikzpicture}
    \begin{groupplot}[
      group style={
        group size=2 by 1,
        horizontal sep=3.05cm,
      },
      width=0.36\linewidth,
      height=3.10cm,
      xbar,
      /pgf/bar shift=0pt,
      axis x line*=bottom,
      axis y line*=left,
      axis line style={black!55},
      clip=false,
      tick style={draw=none},
      tick label style={font=\scriptsize},
      label style={font=\scriptsize},
      title style={font=\small, yshift=-4pt},
      enlarge y limits=0.24,
    ]

      \nextgroupplot[
        title={\textbf{(a)} Counterexample proxy},
        title style={font=\small, yshift=-4pt, xshift=-1.12cm},
        xlabel={Success (\%)},
        xmin=0,
        xmax=88,
        xtick={0,20,40,60,80},
        xticklabels={0,20,40,60,80},
        ymin=0,
        ymax=1,
        ytick={0.33,0.67},
        yticklabels={Incorrect Diagnosis,Correct Diagnosis},
        y tick label style={font=\scriptsize, text width=2.25cm, align=left},
        bar width=7pt,
      ]
        \addplot[fill=diagOrange, draw=none]
          coordinates {(71.6,0.67)};
        \addplot[fill=diagSlate, draw=none]
          coordinates {(10.3,0.33)};
        \node[anchor=west, xshift=2pt, font=\scriptsize]
          at (axis cs:71.6,0.67) {71.6\%};
        \node[anchor=west, xshift=2pt, font=\scriptsize]
          at (axis cs:10.3,0.33) {10.3\%};

      \nextgroupplot[
        title={\textbf{(b)} Diagnosis depth},
        title style={font=\small, yshift=-4pt, xshift=-0.85cm},
        xlabel={Share (\%)},
        xmin=0,
        xmax=64,
        xtick={0,20,40,60},
        xticklabels={0,20,40,60},
        symbolic y coords={Symptom,Mechanism,Located,Actionable},
        ytick={Symptom,Mechanism,Located,Actionable},
        yticklabels={Symptom only,Mechanism only,Located cause,Actionable fix},
        y tick label style={font=\scriptsize, text width=1.70cm, align=left},
        bar width=7pt,
      ]
        \addplot[fill=diagOrange, draw=none]
          coordinates {
            (54.0,Actionable)
            (28.2,Located)
            (10.4,Mechanism)
            (7.4,Symptom)
          };
        \node[anchor=west, xshift=2pt, font=\scriptsize]
          at (axis cs:54.0,Actionable) {54.0\%};
        \node[anchor=west, xshift=2pt, font=\scriptsize]
          at (axis cs:28.2,Located) {28.2\%};
        \node[anchor=west, xshift=2pt, font=\scriptsize]
          at (axis cs:10.4,Mechanism) {10.4\%};
        \node[anchor=west, xshift=2pt, font=\scriptsize]
          at (axis cs:7.4,Symptom) {7.4\%};
    \end{groupplot}
  \end{tikzpicture}
  \setlength{\abovecaptionskip}{2pt}
  \caption{Human audit of \textsc{Arbiter} diagnoses. Counterexample success
  tracks correctness (\textbf{Left}), while correct diagnoses usually locate the bug and
  over half suggest a fix (\textbf{Right}).}
  \label{fig:diagnosis-audit-main}
\end{figure}
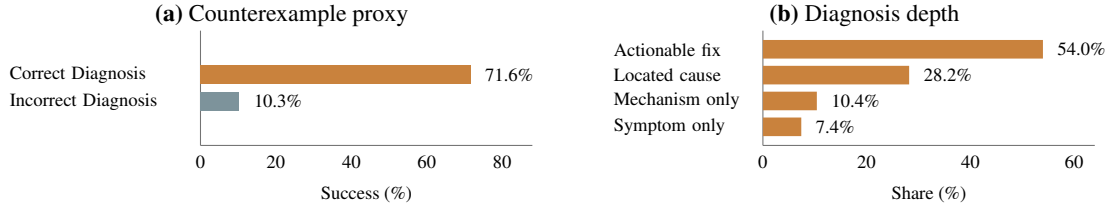

%% file: sections/tables/refinement_results.tex
\begin{table}[H]
\centering
\renewcommand{\arraystretch}{1.1}
\setlength{\tabcolsep}{2.5pt}
\newcolumntype{W}{>{\centering\arraybackslash}p{1.06cm}}
\providecommand{\z}{\phantom{0}}
\providecommand{\tb}[1]{\mbox{\textbf{#1}}}
\providecommand{\su}[1]{\mbox{\underline{#1}}}
\resizebox{\linewidth}{!}{%
\begin{tabular}{@{}l !{\color{gray!70}\vrule} WW!{\color{gray!50}\vrule}WW!{\color{gray!50}\vrule}WW !{\color{gray!70}\vrule} WW!{\color{gray!50}\vrule}WW!{\color{gray!50}\vrule}WW@{}}
\toprule
\multirow{3}{*}[-1.2ex]{\textbf{Condition}}
  & \multicolumn{6}{c!{\color{gray!70}\vrule}}{\logo{qwen.png}~\textbf{Qwen3.6-35B-A3B}}
  & \multicolumn{6}{c}{\logo{openai.png}~\textbf{GPT-OSS-120B}} \\
\cmidrule(lr){2-7}\cmidrule(lr){8-13}
  & \multicolumn{2}{c!{\color{gray!50}\vrule}}{\small\textbf{Easy} {\scriptsize(9)}}
  & \multicolumn{2}{c!{\color{gray!50}\vrule}}{\small\textbf{Medium} {\scriptsize(9)}}
  & \multicolumn{2}{c!{\color{gray!70}\vrule}}{\small\textbf{Hard} {\scriptsize(12)}}
  & \multicolumn{2}{c!{\color{gray!50}\vrule}}{\small\textbf{Easy} {\scriptsize(9)}}
  & \multicolumn{2}{c!{\color{gray!50}\vrule}}{\small\textbf{Medium} {\scriptsize(9)}}
  & \multicolumn{2}{c}{\small\textbf{Hard} {\scriptsize(12)}} \\
\cmidrule(lr){2-3}\cmidrule(lr){4-5}\cmidrule(lr){6-7}\cmidrule(lr){8-9}\cmidrule(lr){10-11}\cmidrule(lr){12-13}
  & {\small @1} & {\small @8} & {\small @1} & {\small @8} & {\small @1} & {\small @8}
  & {\small @1} & {\small @8} & {\small @1} & {\small @8} & {\small @1} & {\small @8} \\
\midrule
Starting code & 69.5 & 69.5 & 52.4 & 52.4 & \z9.5 & \z9.5 & 73.8 & 73.8 & 62.8 & 62.8 & 12.8 & 12.8 \\
\midrule
Previous code only & 52.4 & 93.6 & 38.9 & 59.8 & \z5.8 & 16.6 & 88.1 & 89.6 & 75.9 & 81.6 & 21.8 & 28.5 \\
Online Judge & 68.9 & \su{98.1} & 49.2 & 64.7 & 10.2 & 17.4 & 80.7 & 94.3 & 73.0 & 86.0 & 20.5 & 35.5 \\
\midrule
\rowcolor{arbRow!6}
\logo{judgement.png}~\textsc{Arbiter}: $\mathbf{V}+\mathbf{X}$ & 90.1 & 92.5 & \su{62.4} & 65.9 & \z6.4 & 13.4 & \tb{95.8} & \tb{96.8} & 84.0 & 84.1 & 26.3 & 35.8 \\
\rowcolor{arbRow!6}
\logo{judgement.png}~\textsc{Arbiter}: $\mathbf{V}+\mathbf{D}$ & \tb{95.6} & \tb{100.0\z} & \tb{66.4} & \tb{69.0} & \tb{12.3} & \tb{20.1} & 93.3 & 95.4 & \su{85.0} & \tb{89.2} & \tb{32.8} & \tb{39.1} \\
\rowcolor{arbRow!6}
\logo{judgement.png}~\textsc{Arbiter}: $\mathbf{V}+\mathbf{D}+\mathbf{X}$ & \su{93.1} & 95.8 & \su{62.4} & \su{67.7} & \su{11.0} & \su{19.3} & \su{94.7} & \su{96.0} & \tb{86.6} & \su{88.6} & \su{30.4} & \su{38.3} \\
\bottomrule
\end{tabular}%
}
\caption{Refinement from shared, already judged seeds on 30 non-Bronze USACO
2026 problems. The tiers are Silver, Gold, and Platinum$+$Open. Budgets count
one or eight additional judge submissions; inference compute is unmatched.
\textbf{Bold} and \underline{underlined} entries mark the top two conditions.}
\label{tab:refine}
\end{table}

%% file: sections/tables/selection_results.tex
\begin{table}[H]
\centering
\renewcommand{\arraystretch}{1.1}
\newcolumntype{Y}{>{\centering\arraybackslash}p{1.34cm}}
\definecolor{dsup}{HTML}{2E7D32}
\definecolor{dsdn}{HTML}{B3261E}
\providecommand{\up}[2]{#1}
\providecommand{\upb}[2]{\textbf{#1}}
\providecommand{\dn}[2]{#1}
\providecommand{\nv}[1]{#1}
\providecommand{\selul}[1]{\mbox{\underline{#1}}}
\providecommand{\gv}[2]{\mbox{\color{black!58}#1}}
\providecommand{\mcite}[1]{{\scriptsize\color{black!55}\citep{#1}}}
\resizebox{\linewidth}{!}{%
\begin{tabular}{@{}l !{\color{gray!50}\vrule} YY!{\color{gray!50}\vrule}YY!{\color{gray!70}\vrule}YY!{\color{gray!50}\vrule}YY@{}}
\toprule
\multirow{3}{*}[-0.5ex]{\textbf{Selection method}}
  & \multicolumn{4}{c!{\color{gray!70}\vrule}}{\logo{qwen.png}~\textbf{Qwen3-30B-A3B-Thinking}}
  & \multicolumn{4}{c}{\logo{qwen.png}~\textbf{Qwen3-235B-A22B-Thinking}} \\
\cmidrule(lr){2-5}\cmidrule(lr){6-9}
  & \multicolumn{2}{c!{\color{gray!50}\vrule}}{\small\textbf{LiveCodeBench v6}} & \multicolumn{2}{c!{\color{gray!70}\vrule}}{\small\textbf{OJBench}}
  & \multicolumn{2}{c!{\color{gray!50}\vrule}}{\small\textbf{LiveCodeBench v6}} & \multicolumn{2}{c}{\small\textbf{OJBench}} \\
\cmidrule(lr){2-3}\cmidrule(lr){4-5}\cmidrule(lr){6-7}\cmidrule(lr){8-9}
  & {\small Pool=8} & {\small Pool=64} & {\small Pool=8} & {\small Pool=64}
  & {\small Pool=8} & {\small Pool=64} & {\small Pool=8} & {\small Pool=64} \\
\midrule
Uniform selection & \nv{66.2} & \nv{66.2} & \nv{28.9} & \nv{28.9} & \nv{74.7} & \nv{74.7} & \nv{37.1} & \nv{37.1} \\
\midrule
Grading RM~\mcite{skyworkrewardv2} & \up{67.1}{0.9} & \up{68.0}{1.8} & \dn{28.7}{0.2} & \up{29.3}{0.4} & \up{75.9}{1.2} & \up{74.9}{0.2} & \up{37.6}{0.5} & \dn{34.9}{2.2} \\
MBR-Exec (hard)~\mcite{shi-etal-2022-natural} & \up{68.5}{2.3} & \up{69.2}{3.0} & \up{29.6}{0.7} & \up{32.4}{3.5} & \up{77.9}{3.2} & \up{77.7}{3.0} & \up{37.7}{0.6} & \up{39.6}{2.5} \\
MBR-Exec (soft)~\mcite{shi-etal-2022-natural} & \up{70.0}{3.8} & \up{70.0}{3.8} & \up{31.4}{2.5} & \up{31.8}{2.9} & \up{78.5}{3.8} & \up{78.5}{3.8} & \up{40.5}{3.4} & \up{41.4}{4.3} \\
CodeT~\mcite{codet2023} & \up{72.7}{6.5} & \up{70.5}{4.3} & \up{31.6}{2.7} & \up{32.1}{3.2} & \up{78.8}{4.1} & \up{79.2}{4.5} & \up{41.0}{3.9} & \up{40.6}{3.5} \\
CodeRM~\mcite{ma-etal-2025-dynamic} & \up{71.2}{5.0} & \up{71.4}{5.2} & \up{32.0}{3.1} & \up{31.3}{2.4} & \up{79.3}{4.6} & \up{80.3}{5.6} & \up{40.4}{3.3} & \up{39.5}{2.4} \\
Random Generator~\mcite{alphacode2022} & \up{70.7}{4.5} & \up{70.4}{4.2} & \up{33.4}{4.5} & \up{34.1}{5.2} & \up{78.8}{4.1} & \up{78.9}{4.2} & \up{43.8}{6.7} & \up{44.2}{7.1} \\
Agentic Verifier~\mcite{agenticverifier2026} & \up{73.6}{7.4} & \up{73.7}{7.5} & \up{33.5}{4.6} & \up{34.9}{6.0} & \up{81.6}{6.9} & \up{82.9}{8.2} & \up{45.9}{8.8} & \up{48.6}{11.5} \\
\midrule
\rowcolor{arbRow!6}
\logo{judgement.png}~\textsc{Arbiter} {\small(1 call)} & \up{76.8}{10.6} & \up{82.6}{16.4} & \up{37.6}{8.7} & \up{43.6}{14.7} & \up{85.0}{10.3} & \up{87.7}{13.0} & \up{50.0}{12.9} & \up{53.3}{16.2} \\
\rowcolor{arbRow!6}
\logo{judgement.png}~\textsc{Arbiter} {\small(8 calls)} & \upb{79.5}{13.3} & \upb{84.6}{18.4} & \upb{40.0}{11.1} & \upb{46.3}{17.4} & \upb{85.9}{11.2} & \upb{89.3}{14.6} & \upb{51.7}{14.6} & \upb{56.1}{19.0} \\
\midrule
Perfect selection\textsuperscript{\textdagger} & \selul{81.1} & \selul{86.3} & \selul{42.2} & \selul{50.0} & \selul{88.0} & \selul{92.6} & \selul{55.6} & \selul{62.1} \\
\bottomrule
\end{tabular}%
}
\vspace{1pt}
{\scriptsize\raggedright \textsuperscript{\textdagger}Perfect selection uses online-judge verdicts for our evaluated candidate pools and serves as their upper bound, unavailable at inference time.\par}
\caption{Pass@$1$ following~\citet{agenticverifier2026}. Baselines are
published; \textsc{Arbiter} and perfect selection use regenerated pools with
unmatched inference compute.}
\label{tab:selection}
\end{table}

%% file: sections/body/conclusion.tex
\section{Conclusion}
\label{sec:conclusion}

\textsc{Arbiter} combines candidate verification and counterexample-grounded
diagnosis. On \textsc{USACO-Judge}, its candidate-weighted verdict accuracy rises
from 71.41\% to 87.41\%, and overall counterexample success from 40.61\% to
62.67\%. Under the reproduced protocol, it leads reported code-selection
results and improves refinement over the evaluated baselines. In our IOI 2026
case study, \textsc{Arbiter}-guided GPT-OSS-120B reaches 385.46 expected points
under the official 50-submission cap per problem, above the 361.12 gold
threshold.

%% file: sections/appendix/verification_funnel.tex
\subsection{Verification Funnel}
\label{app:verification-funnel}

For each split, \textbf{Attempt} is the fraction of unaccepted candidates for
which the model submits a counterexample generator. \textbf{Exec.},
\textbf{Valid}, and \textbf{Succ.} measure generator execution, input validity,
and counterexample success, each conditional on reaching the preceding stage.
The columns therefore show where an attempted verification succeeds or is lost.
\textsc{UOJ-Bench} contains only unaccepted candidates, so \emph{Accepted}
precision and recall are reported for \textsc{USACO-Judge}.

\begin{table}[h]
\centering
\renewcommand{\arraystretch}{1.2}
\setlength{\tabcolsep}{0pt}
\textbf{(a) \textsc{UOJ-Bench} (Hacking)}\par\vspace{2pt}
\resizebox{\linewidth}{!}{%
\begin{tabular}{@{}>{\raggedright\arraybackslash}p{4.6cm} !{\color{gray!50}\vrule} *{4}{S[table-format=3.2, table-column-width=1.725cm]} !{\color{gray!50}\vrule} *{4}{S[table-format=3.2, table-column-width=1.725cm]}@{}}
\toprule
\multirow{2}{*}{\textbf{Model}}
& \multicolumn{4}{c !{\color{gray!50}\vrule}}{\textbf{Easy}}
& \multicolumn{4}{c}{\textbf{Hard}} \\
\cmidrule(lr){2-5}\cmidrule(l){6-9}
& {Att.} & {Exec.} & {Valid} & {Succ.}
& {Att.} & {Exec.} & {Valid} & {Succ.} \\
\midrule
\multicolumn{9}{c}{{\scriptsize\textcolor{gray}{\faBolt}}~\textit{Non-agentic}} \\
\midrule
\logo{qwen.png}~Qwen3.6-35B-A3B
& 86.85 & 96.49 & 92.03 & 22.89
& 88.45 & 96.95 & 89.92 & 16.23 \\
\logo{openai.png}~GPT-OSS-120B
& 83.23 & 91.82 & 92.80 & 41.02
& 83.12 & 89.34 & 90.04 & 30.01 \\
\logo{deepseek.png}~DeepSeek-V4-Flash-Preview
& 80.04 & 90.40 & 95.38 & 48.51
& 79.60 & 90.03 & 94.84 & 32.41 \\
\logo{qwen.png}~Qwen3.5-397B-A17B
& 98.47 & 89.26 & 95.02 & 49.91
& 98.14 & 88.75 & 95.05 & 37.18 \\
\logo{zai.png}~GLM-5.2
& 87.47 & 89.02 & 95.98 & 87.43
& 84.35 & 87.83 & 97.71 & 79.20 \\
\logo{kimi.png}~Kimi-K2.5
& 75.16 & 95.28 & 95.34 & 68.50
& 70.65 & 95.21 & 92.60 & 49.68 \\
\logo{deepseek.png}~DeepSeek-V4-Pro-Preview
& 73.49 & 98.96 & 94.63 & 74.39
& 71.74 & 97.52 & 96.49 & 57.70 \\
\logo{claude.png}~Claude-Opus-4.8
& 84.97 & 100.00 & 98.53 & 92.02
& 75.32 & 100.00 & 99.08 & 89.87 \\
\logo{openai.png}~GPT-5.5
& 91.65 & 99.54 & 96.80 & 89.13
& 89.05 & 99.44 & 98.88 & 87.95 \\
\midrule
\multicolumn{9}{c}{{\scriptsize\textcolor{gray}{\faCog}}~\textit{Agentic}} \\
\midrule
\logo{qwen.png}~Qwen3.6-35B-A3B
& 65.43 & 97.13 & 90.45 & 68.98
& 64.03 & 96.56 & 89.00 & 59.29 \\
\logo{deepseek.png}~DeepSeek-V4-Flash-Preview
& 64.30 & 100.00 & 97.73 & 89.70
& 59.80 & 99.83 & 96.17 & 85.96 \\
\midrule
\rowcolor{arbRow!6}
\logo{judgement.png}~\textbf{\textsc{Arbiter}} {\small(Ours)}
& \bfseries 83.87 & \bfseries 99.66 & \bfseries 96.07 & \bfseries 76.57
& \bfseries 75.69 & \bfseries 99.11 & \bfseries 94.79 & \bfseries 70.35 \\
\bottomrule
\end{tabular}%
}

\vspace{5pt}
\textbf{(b) \textsc{USACO-Judge}}\par\vspace{2pt}
\resizebox{\linewidth}{!}{%
\begin{tabular}{@{}>{\raggedright\arraybackslash}p{4.6cm} !{\color{gray!50}\vrule} *{4}{S[table-format=3.2, table-column-width=1.15cm]} !{\color{gray!30}\vrule} *{2}{S[table-format=3.2, table-column-width=1.15cm]} !{\color{gray!50}\vrule} *{4}{S[table-format=3.2, table-column-width=1.15cm]} !{\color{gray!30}\vrule} *{2}{S[table-format=3.2, table-column-width=1.15cm]}@{}}
\toprule
\multirow{3}{*}{\textbf{Model}}
& \multicolumn{6}{c !{\color{gray!50}\vrule}}{\textbf{Easy}}
& \multicolumn{6}{c}{\textbf{Hard}} \\
\cmidrule(lr){2-7}\cmidrule(l){8-13}
& \multicolumn{4}{c !{\color{gray!30}\vrule}}{\textbf{Counterexample}}
& \multicolumn{2}{c !{\color{gray!50}\vrule}}{\textbf{\emph{Accepted}}}
& \multicolumn{4}{c !{\color{gray!30}\vrule}}{\textbf{Counterexample}}
& \multicolumn{2}{c}{\textbf{\emph{Accepted}}} \\
\cmidrule(lr){2-5}\cmidrule(lr){6-7}\cmidrule(lr){8-11}\cmidrule(l){12-13}
& {Att.} & {Exec.} & {Valid} & {Succ.}
& {Prec.} & {Rec.}
& {Att.} & {Exec.} & {Valid} & {Succ.}
& {Prec.} & {Rec.} \\
\midrule
\multicolumn{13}{c}{{\scriptsize\textcolor{gray}{\faBolt}}~\textit{Non-agentic}} \\
\midrule
\logo{qwen.png}~Qwen3.6-35B-A3B
& 77.06 & 94.09 & 94.59 & 43.23 & 51.06 & 51.11
& 87.34 & 95.91 & 94.68 & 28.21 & 47.68 & 24.58 \\
\logo{openai.png}~GPT-OSS-120B
& 63.33 & 83.94 & 93.03 & 68.67 & 44.64 & 73.70
& 82.29 & 88.11 & 95.78 & 53.08 & 50.39 & 50.32 \\
\logo{deepseek.png}~DeepSeek-V4-Flash-Preview
& 77.67 & 90.52 & 98.73 & 79.19 & 58.86 & 83.11
& 83.46 & 92.51 & 97.09 & 55.38 & 53.34 & 50.86 \\
\logo{qwen.png}~Qwen3.5-397B-A17B
& 96.68 & 89.52 & 98.94 & 90.30 & 75.98 & 81.85
& 96.90 & 87.51 & 99.18 & 70.05 & 63.87 & 53.02 \\
\logo{zai.png}~GLM-5.2
& 97.24 & 94.89 & 100.00 & 98.80 & 88.54 & 94.44
& 90.55 & 95.42 & 98.40 & 92.95 & 81.77 & 70.48 \\
\logo{kimi.png}~Kimi-K2.5
& 80.00 & 95.83 & 98.55 & 93.38 & 68.50 & 96.67
& 77.80 & 96.40 & 97.82 & 78.03 & 58.06 & 68.57 \\
\logo{deepseek.png}~DeepSeek-V4-Pro-Preview
& 90.20 & 99.18 & 100.00 & 94.85 & 83.21 & 90.74
& 85.98 & 98.47 & 98.17 & 85.36 & 73.61 & 75.71 \\
\logo{claude.png}~Claude-Opus-4.8
& 94.48 & 100.00 & 100.00 & 100.00 & 91.40 & 94.44
& 89.91 & 100.00 & 100.00 & 97.45 & 92.21 & 67.62 \\
\logo{openai.png}~GPT-5.5
& 92.31 & 98.81 & 100.00 & 99.40 & 87.88 & 96.67
& 96.10 & 99.52 & 98.80 & 97.33 & 91.35 & 80.48 \\
\midrule
\multicolumn{13}{c}{{\scriptsize\textcolor{gray}{\faCog}}~\textit{Agentic}} \\
\midrule
\logo{qwen.png}~Qwen3.6-35B-A3B
& 90.79 & 97.85 & 97.69 & 91.06 & 86.80 & 81.67
& 69.85 & 96.59 & 92.03 & 73.44 & 59.18 & 61.43 \\
\logo{deepseek.png}~DeepSeek-V4-Flash-Preview
& 93.33 & 100.00 & 100.00 & 99.40 & 90.00 & 100.00
& 72.49 & 100.00 & 98.71 & 91.21 & 63.07 & 86.19 \\
\midrule
\rowcolor{arbRow!6}
\logo{judgement.png}~\textbf{\textsc{Arbiter}} {\small(Ours)}
& \bfseries 98.70 & \bfseries 100.00 & \bfseries 99.51 & \bfseries 98.40 & \bfseries 97.30 & \bfseries 85.00
& \bfseries 92.48 & \bfseries 99.81 & \bfseries 99.16 & \bfseries 86.74 & \bfseries 83.37 & \bfseries 67.32 \\
\bottomrule
\end{tabular}%
}
\caption{Verification funnel and acceptance quality by benchmark split. Each funnel
stage is conditional on the preceding stage. Prec. and Rec. treat \emph{Accepted} as the positive class.}
\label{tab:verification-funnel}
\end{table}

Training improves both the decision to attack and the quality of the attack.
Relative to its agentic Qwen3.6-35B-A3B baseline, \textsc{Arbiter} raises
\textbf{Attempt} by 8--23 percentage points across the four splits. Since this
rate is measured only on truly unaccepted candidates, the increase means that
more buggy programs are identified and taken forward to counterexample search.
Execution and validity exceed 94.7\% on every split, while conditional
counterexample success rises by another 7--13 points. The final gain therefore
comes from more attempts and a higher conversion rate for those attempts.

The \textsc{USACO-Judge} acceptance columns measure the corresponding behavior
on correct programs. On the hard split, \emph{Accepted} precision rises from
59.18\% to 83.37\%, and recall rises from 61.43\% to 67.32\%. These results pair
broader bug coverage with more reliable acceptance decisions.

\paragraph{Counterexample characteristics.}
Table~\ref{tab:counterexample-size} examines generated input sizes and the
failure types of successful hacks. Among validator-approved inputs from \textsc{Arbiter},
65.6--89.0\% are smaller than 1 KB, while 7.0--23.9\% are at least 1 MB.
The large-input share exceeds that of the agentic Qwen3.6-35B-A3B
baseline on every split, showing that training adds constraint-scale
stress tests while compact inputs remain the majority. Successful hacks are
also varied. \emph{Wrong Answer} accounts for 65.1--77.1\% and
\emph{Time Limit Exceeded} for 11.8--29.9\%, with other failure types making up
the remainder.

\input{sections/tables/counterexample_sizes}

\clearpage
\subsection{Interaction Turns}

Table~\ref{tab:turns} places these results against the amount of interaction
used by each agentic verifier. \textsc{Arbiter} averages 14.2--18.7 turns per
candidate, only 1.2--2.1 more than its Qwen3.6-35B-A3B baseline and fewer
than agentic DeepSeek-V4-Flash-Preview on every split. All three systems operate
at a similar interaction scale.

%% file: sections/tables/counterexample_sizes.tex
\definecolor{tripLeft}{HTML}{5F7D91}
\definecolor{tripMiddle}{HTML}{E7ECEF}
\definecolor{tripRight}{HTML}{9BB0A5}
\newcommand{\tripbar}[2]{%
  \begin{tikzpicture}[baseline=-0.4ex,x=0.031cm,y=1cm]
    \useasboundingbox (0,0) rectangle (100,0.26);
    \pgfmathsetmacro{\tripleftend}{17.5+0.65*#1}
    \pgfmathsetmacro{\triprightstart}{17.5+0.65*(100-#2)}
    \fill[tripLeft] (17.5,0.02) rectangle (\tripleftend,0.24);
    \fill[tripMiddle] (\tripleftend,0.02) rectangle (\triprightstart,0.24);
    \fill[tripRight] (\triprightstart,0.02) rectangle (82.5,0.24);
    \draw[black!18,line width=0.2pt] (17.5,0.02) rectangle (82.5,0.24);
    \node[anchor=west,font=\scriptsize,text=black!85,inner sep=0pt] at (2,0.13) {#1};
    \node[anchor=east,font=\scriptsize,text=black!85,inner sep=0pt] at (98,0.13) {#2};
  \end{tikzpicture}%
}

\begin{table}[t]
\centering
\renewcommand{\arraystretch}{1.05}
\setlength{\tabcolsep}{0pt}
\textbf{(a) \textsc{UOJ-Bench} (Hacking)}\par\vspace{2pt}
\resizebox{\linewidth}{!}{%
\begin{tabular}{@{}>{\raggedright\arraybackslash}p{4.6cm} !{\color{gray!50}\vrule} *{2}{S[table-format=3.1, table-column-width=1.725cm]} !{\color{gray!30}\vrule} *{2}{S[table-format=3.1, table-column-width=1.725cm]} !{\color{gray!50}\vrule} *{2}{S[table-format=3.1, table-column-width=1.725cm]} !{\color{gray!30}\vrule} *{2}{S[table-format=3.1, table-column-width=1.725cm]}@{}}
\toprule
\multirow{2}{*}{\textbf{Model}}
& \multicolumn{4}{c !{\color{gray!50}\vrule}}{\textbf{Easy}}
& \multicolumn{4}{c}{\textbf{Hard}} \\
\cmidrule(lr){2-5}\cmidrule(l){6-9}
& \multicolumn{2}{c !{\color{gray!30}\vrule}}{\shortstack{\textbf{Input Size} (\%)\\[-1pt]\scriptsize \ensuremath{<1}KB \enspace\textbar\enspace other \enspace\textbar\enspace \ensuremath{\geq1}MB}}
& \multicolumn{2}{c !{\color{gray!50}\vrule}}{\shortstack{\textbf{Hack Type} (\%)\\[-1pt]\scriptsize WA \enspace\textbar\enspace Other \enspace\textbar\enspace TLE}}
& \multicolumn{2}{c !{\color{gray!30}\vrule}}{\shortstack{\textbf{Input Size} (\%)\\[-1pt]\scriptsize \ensuremath{<1}KB \enspace\textbar\enspace other \enspace\textbar\enspace \ensuremath{\geq1}MB}}
& \multicolumn{2}{c}{\shortstack{\textbf{Hack Type} (\%)\\[-1pt]\scriptsize WA \enspace\textbar\enspace Other \enspace\textbar\enspace TLE}} \\
\midrule
\multicolumn{9}{c}{{\scriptsize\textcolor{gray}{\faBolt}}~\textit{Non-agentic}} \\
\midrule
\logo{qwen.png}~Qwen3.6-35B-A3B
& \multicolumn{2}{c !{\color{gray!35}\vrule}}{\tripbar{83.1}{7.7}} & \multicolumn{2}{c !{\color{gray!60}\vrule}}{\tripbar{78.4}{9.2}}
& \multicolumn{2}{c !{\color{gray!35}\vrule}}{\tripbar{80.3}{7.3}} & \multicolumn{2}{c}{\tripbar{70.2}{13.9}} \\
\logo{openai.png}~GPT-OSS-120B
& \multicolumn{2}{c !{\color{gray!35}\vrule}}{\tripbar{67.0}{18.3}} & \multicolumn{2}{c !{\color{gray!60}\vrule}}{\tripbar{73.9}{10.8}}
& \multicolumn{2}{c !{\color{gray!35}\vrule}}{\tripbar{67.7}{18.1}} & \multicolumn{2}{c}{\tripbar{65.8}{13.6}} \\
\logo{deepseek.png}~DeepSeek-V4-Flash-Preview
& \multicolumn{2}{c !{\color{gray!35}\vrule}}{\tripbar{79.8}{11.7}} & \multicolumn{2}{c !{\color{gray!60}\vrule}}{\tripbar{79.3}{9.1}}
& \multicolumn{2}{c !{\color{gray!35}\vrule}}{\tripbar{81.3}{8.7}} & \multicolumn{2}{c}{\tripbar{79.1}{11.0}} \\
\logo{qwen.png}~Qwen3.5-397B-A17B
& \multicolumn{2}{c !{\color{gray!35}\vrule}}{\tripbar{72.7}{13.2}} & \multicolumn{2}{c !{\color{gray!60}\vrule}}{\tripbar{78.3}{10.9}}
& \multicolumn{2}{c !{\color{gray!35}\vrule}}{\tripbar{72.9}{11.3}} & \multicolumn{2}{c}{\tripbar{73.0}{16.8}} \\
\logo{zai.png}~GLM-5.2
& \multicolumn{2}{c !{\color{gray!35}\vrule}}{\tripbar{67.3}{17.9}} & \multicolumn{2}{c !{\color{gray!60}\vrule}}{\tripbar{77.0}{10.5}}
& \multicolumn{2}{c !{\color{gray!35}\vrule}}{\tripbar{71.6}{14.7}} & \multicolumn{2}{c}{\tripbar{77.6}{16.3}} \\
\logo{kimi.png}~Kimi-K2.5
& \multicolumn{2}{c !{\color{gray!35}\vrule}}{\tripbar{74.3}{14.4}} & \multicolumn{2}{c !{\color{gray!60}\vrule}}{\tripbar{77.7}{10.3}}
& \multicolumn{2}{c !{\color{gray!35}\vrule}}{\tripbar{71.9}{12.3}} & \multicolumn{2}{c}{\tripbar{76.8}{12.9}} \\
\logo{deepseek.png}~DeepSeek-V4-Pro-Preview
& \multicolumn{2}{c !{\color{gray!35}\vrule}}{\tripbar{71.6}{15.5}} & \multicolumn{2}{c !{\color{gray!60}\vrule}}{\tripbar{76.7}{11.2}}
& \multicolumn{2}{c !{\color{gray!35}\vrule}}{\tripbar{73.9}{12.2}} & \multicolumn{2}{c}{\tripbar{76.5}{15.3}} \\
\logo{claude.png}~Claude-Opus-4.8
& \multicolumn{2}{c !{\color{gray!35}\vrule}}{\tripbar{61.6}{22.7}} & \multicolumn{2}{c !{\color{gray!60}\vrule}}{\tripbar{73.2}{13.6}}
& \multicolumn{2}{c !{\color{gray!35}\vrule}}{\tripbar{62.0}{18.7}} & \multicolumn{2}{c}{\tripbar{70.8}{22.6}} \\
\logo{openai.png}~GPT-5.5
& \multicolumn{2}{c !{\color{gray!35}\vrule}}{\tripbar{64.1}{18.4}} & \multicolumn{2}{c !{\color{gray!60}\vrule}}{\tripbar{75.9}{13.8}}
& \multicolumn{2}{c !{\color{gray!35}\vrule}}{\tripbar{65.7}{17.0}} & \multicolumn{2}{c}{\tripbar{74.9}{19.1}} \\
\midrule
\multicolumn{9}{c}{{\scriptsize\textcolor{gray}{\faCog}}~\textit{Agentic}} \\
\midrule
\logo{qwen.png}~Qwen3.6-35B-A3B
& \multicolumn{2}{c !{\color{gray!35}\vrule}}{\tripbar{81.3}{6.2}} & \multicolumn{2}{c !{\color{gray!60}\vrule}}{\tripbar{86.9}{4.5}}
& \multicolumn{2}{c !{\color{gray!35}\vrule}}{\tripbar{87.8}{4.7}} & \multicolumn{2}{c}{\tripbar{85.4}{6.9}} \\
\logo{deepseek.png}~DeepSeek-V4-Flash-Preview
& \multicolumn{2}{c !{\color{gray!35}\vrule}}{\tripbar{76.1}{11.3}} & \multicolumn{2}{c !{\color{gray!60}\vrule}}{\tripbar{80.7}{9.3}}
& \multicolumn{2}{c !{\color{gray!35}\vrule}}{\tripbar{84.9}{6.2}} & \multicolumn{2}{c}{\tripbar{85.9}{8.7}} \\
\midrule
\rowcolor{arbRow!6}
\logo{judgement.png}~\textbf{\textsc{Arbiter}} {\small(Ours)}
& \multicolumn{2}{c !{\color{gray!35}\vrule}}{\tripbar{65.9}{21.0}} & \multicolumn{2}{c !{\color{gray!60}\vrule}}{\tripbar{75.0}{11.8}}
& \multicolumn{2}{c !{\color{gray!35}\vrule}}{\tripbar{69.3}{17.4}} & \multicolumn{2}{c}{\tripbar{77.1}{15.3}} \\
\bottomrule
\end{tabular}%
}

\vspace{5pt}
\textbf{(b) \textsc{USACO-Judge}}\par\vspace{2pt}
\resizebox{\linewidth}{!}{%
\begin{tabular}{@{}>{\raggedright\arraybackslash}p{4.6cm} !{\color{gray!50}\vrule} *{2}{S[table-format=3.1, table-column-width=1.725cm]} !{\color{gray!30}\vrule} *{2}{S[table-format=3.1, table-column-width=1.725cm]} !{\color{gray!50}\vrule} *{2}{S[table-format=3.1, table-column-width=1.725cm]} !{\color{gray!30}\vrule} *{2}{S[table-format=3.1, table-column-width=1.725cm]}@{}}
\toprule
\multirow{2}{*}{\textbf{Model}}
& \multicolumn{4}{c !{\color{gray!50}\vrule}}{\textbf{Easy}}
& \multicolumn{4}{c}{\textbf{Hard}} \\
\cmidrule(lr){2-5}\cmidrule(l){6-9}
& \multicolumn{2}{c !{\color{gray!30}\vrule}}{\shortstack{\textbf{Input Size} (\%)\\[-1pt]\scriptsize \ensuremath{<1}KB \enspace\textbar\enspace other \enspace\textbar\enspace \ensuremath{\geq1}MB}}
& \multicolumn{2}{c !{\color{gray!50}\vrule}}{\shortstack{\textbf{Hack Type} (\%)\\[-1pt]\scriptsize WA \enspace\textbar\enspace Other \enspace\textbar\enspace TLE}}
& \multicolumn{2}{c !{\color{gray!30}\vrule}}{\shortstack{\textbf{Input Size} (\%)\\[-1pt]\scriptsize \ensuremath{<1}KB \enspace\textbar\enspace other \enspace\textbar\enspace \ensuremath{\geq1}MB}}
& \multicolumn{2}{c}{\shortstack{\textbf{Hack Type} (\%)\\[-1pt]\scriptsize WA \enspace\textbar\enspace Other \enspace\textbar\enspace TLE}} \\
\midrule
\multicolumn{9}{c}{{\scriptsize\textcolor{gray}{\faBolt}}~\textit{Non-agentic}} \\
\midrule
\logo{qwen.png}~Qwen3.6-35B-A3B
& \multicolumn{2}{c !{\color{gray!35}\vrule}}{\tripbar{90.1}{4.1}} & \multicolumn{2}{c !{\color{gray!60}\vrule}}{\tripbar{73.0}{26.3}}
& \multicolumn{2}{c !{\color{gray!35}\vrule}}{\tripbar{86.8}{5.1}} & \multicolumn{2}{c}{\tripbar{62.4}{29.5}} \\
\logo{openai.png}~GPT-OSS-120B
& \multicolumn{2}{c !{\color{gray!35}\vrule}}{\tripbar{85.9}{6.6}} & \multicolumn{2}{c !{\color{gray!60}\vrule}}{\tripbar{62.3}{37.7}}
& \multicolumn{2}{c !{\color{gray!35}\vrule}}{\tripbar{63.0}{21.7}} & \multicolumn{2}{c}{\tripbar{50.0}{41.7}} \\
\logo{deepseek.png}~DeepSeek-V4-Flash-Preview
& \multicolumn{2}{c !{\color{gray!35}\vrule}}{\tripbar{88.2}{7.3}} & \multicolumn{2}{c !{\color{gray!60}\vrule}}{\tripbar{78.2}{21.0}}
& \multicolumn{2}{c !{\color{gray!35}\vrule}}{\tripbar{85.6}{8.4}} & \multicolumn{2}{c}{\tripbar{73.4}{20.4}} \\
\logo{qwen.png}~Qwen3.5-397B-A17B
& \multicolumn{2}{c !{\color{gray!35}\vrule}}{\tripbar{83.1}{8.7}} & \multicolumn{2}{c !{\color{gray!60}\vrule}}{\tripbar{74.5}{25.5}}
& \multicolumn{2}{c !{\color{gray!35}\vrule}}{\tripbar{69.0}{16.7}} & \multicolumn{2}{c}{\tripbar{59.8}{35.3}} \\
\logo{zai.png}~GLM-5.2
& \multicolumn{2}{c !{\color{gray!35}\vrule}}{\tripbar{91.7}{5.3}} & \multicolumn{2}{c !{\color{gray!60}\vrule}}{\tripbar{77.0}{22.4}}
& \multicolumn{2}{c !{\color{gray!35}\vrule}}{\tripbar{67.3}{19.4}} & \multicolumn{2}{c}{\tripbar{67.1}{29.2}} \\
\logo{kimi.png}~Kimi-K2.5
& \multicolumn{2}{c !{\color{gray!35}\vrule}}{\tripbar{87.8}{5.0}} & \multicolumn{2}{c !{\color{gray!60}\vrule}}{\tripbar{76.4}{23.6}}
& \multicolumn{2}{c !{\color{gray!35}\vrule}}{\tripbar{72.3}{14.1}} & \multicolumn{2}{c}{\tripbar{67.3}{25.7}} \\
\logo{deepseek.png}~DeepSeek-V4-Pro-Preview
& \multicolumn{2}{c !{\color{gray!35}\vrule}}{\tripbar{90.2}{6.0}} & \multicolumn{2}{c !{\color{gray!60}\vrule}}{\tripbar{76.8}{23.1}}
& \multicolumn{2}{c !{\color{gray!35}\vrule}}{\tripbar{69.0}{18.3}} & \multicolumn{2}{c}{\tripbar{67.1}{27.5}} \\
\logo{claude.png}~Claude-Opus-4.8
& \multicolumn{2}{c !{\color{gray!35}\vrule}}{\tripbar{89.5}{5.2}} & \multicolumn{2}{c !{\color{gray!60}\vrule}}{\tripbar{76.6}{22.8}}
& \multicolumn{2}{c !{\color{gray!35}\vrule}}{\tripbar{57.1}{27.5}} & \multicolumn{2}{c}{\tripbar{58.9}{37.4}} \\
\logo{openai.png}~GPT-5.5
& \multicolumn{2}{c !{\color{gray!35}\vrule}}{\tripbar{91.6}{5.4}} & \multicolumn{2}{c !{\color{gray!60}\vrule}}{\tripbar{77.6}{21.8}}
& \multicolumn{2}{c !{\color{gray!35}\vrule}}{\tripbar{59.8}{23.2}} & \multicolumn{2}{c}{\tripbar{64.6}{32.4}} \\
\midrule
\multicolumn{9}{c}{{\scriptsize\textcolor{gray}{\faCog}}~\textit{Agentic}} \\
\midrule
\logo{qwen.png}~Qwen3.6-35B-A3B
& \multicolumn{2}{c !{\color{gray!35}\vrule}}{\tripbar{94.7}{2.4}} & \multicolumn{2}{c !{\color{gray!60}\vrule}}{\tripbar{80.3}{19.0}}
& \multicolumn{2}{c !{\color{gray!35}\vrule}}{\tripbar{93.2}{2.6}} & \multicolumn{2}{c}{\tripbar{87.6}{9.7}} \\
\logo{deepseek.png}~DeepSeek-V4-Flash-Preview
& \multicolumn{2}{c !{\color{gray!35}\vrule}}{\tripbar{94.0}{3.6}} & \multicolumn{2}{c !{\color{gray!60}\vrule}}{\tripbar{79.0}{20.4}}
& \multicolumn{2}{c !{\color{gray!35}\vrule}}{\tripbar{85.0}{9.1}} & \multicolumn{2}{c}{\tripbar{82.1}{12.9}} \\
\midrule
\rowcolor{arbRow!6}
\logo{judgement.png}~\textbf{\textsc{Arbiter}} {\small(Ours)}
& \multicolumn{2}{c !{\color{gray!35}\vrule}}{\tripbar{89.0}{7.0}} & \multicolumn{2}{c !{\color{gray!60}\vrule}}{\tripbar{76.3}{22.7}}
& \multicolumn{2}{c !{\color{gray!35}\vrule}}{\tripbar{65.6}{23.9}} & \multicolumn{2}{c}{\tripbar{65.1}{29.9}} \\
\bottomrule
\end{tabular}%
}
\caption{Percentage composition of validator-approved input sizes and successful-hack verdicts. Each bar runs from \ensuremath{<1}KB or WA (left), through other cases, to \ensuremath{\geq1}MB or TLE (right).}
\label{tab:counterexample-size}
\end{table}

%% file: sections/appendix/accepted_trajectory.tex
\subsection{Example Accepted Trajectory}
\label{app:arbiter-trajectory}

We trace an \textit{Accepted} prediction from \textsc{USACO-Judge}. Submission
913421 solves the Bronze problem \emph{Photoshoot} and scores 100 on the
official tests. The task maintains an $N\times N$ grid under increasing point
updates and asks for the maximum sum of any $K\times K$ window after each
update. \textsc{Arbiter} returns \textit{Accepted} after four turns.

\begin{table}[h]
\centering
\small
\renewcommand{\arraystretch}{1.2}
\begin{tabularx}{\linewidth}{@{}>{\centering\arraybackslash}m{2.4em}X@{}}
\toprule
\textbf{Turn} & \textbf{Reasoning and observation} \\
\midrule
\ding{172} & Identifies that \texttt{S[i][j]} stores the sum of the window whose
top-left corner is $(i,j)$. It derives the exact range of windows affected by
an update and compares the candidate with brute force on 100,000 random tests,
finding no mismatch. \\
\hline
\ding{173} & Checks the worst-case cost. Each update touches at most $K^2\leq625$
windows, giving at most 18.75 million updates over the full input. It then
generates a constraint-scale timing instance. \\
\hline
\ding{174} & Repairs the timing input so repeated updates to a cell remain strictly
increasing, compiles the candidate, and measures 0.088 seconds for 30,000
updates. \\
\hline
\ding{175} & Rechecks boundary cases, integer range, memory use, and input format, then
returns \textit{Accepted} with no counterexample generator. \\
\bottomrule
\end{tabularx}
\caption{An \textsc{Arbiter} trajectory for \textsc{USACO-Judge} submission 913421. Four turns of analysis and testing correctly support its \textit{Accepted} verdict.}
\label{tab:arbiter-trajectory}
\end{table}

The candidate updates exactly those window origins in
{\small
\[
  i\in[\max(1,r-K+1),\min(N-K+1,r)],\qquad
  j\in[\max(1,c-K+1),\min(N-K+1,c)].
\]
}
Since all cell values only increase, the running maximum never needs to be
recomputed from unaffected windows. The final diagnosis states this invariant,
the $O(QK^2)$ time bound, and the agreement with differential testing. The
submitted verdict is \textit{Accepted}, and $g$ is empty.

%% file: sections/appendix/accepted_submission_breaks.tex
\subsection{Breaking Officially Accepted Submissions}
\label{app:accepted-breaks}

The preceding section shows the ordinary \textit{Accepted} outcome. Here we
study the rarer setting in which \textsc{Arbiter} hacks a submission that
received an \textit{Accepted} verdict from the official USACO tests. We show
one time-limit failure and one wrong answer, each confirmed by replay on our
UOJ deployment.

\paragraph{Time Limit Exceeded from a quadratic cleanup loop.}
Submission 912773 for \emph{Hoof, Paper, Scissors Triples} (Contest 1
Platinum P1) scored 100 on the official tests. It solves the main geometric
counting problem, then executes the following loop over groups of equal
strategy vectors:

\begin{codebox}
for (auto &pr : g) {
    long long c = pr.second, it = c * c;
    for (long long i = 0; i < it; ++i) t += i;
}
\end{codebox}

The official tests accepted the submission. \textsc{Arbiter} noticed that the
postprocessing cost is quadratic in the largest group size. It placed all
200,000 cows in one group by assigning every cow the same strategy. The correct
answer is zero because all pairwise games tie, but the candidate executes
$200{,}000^2 = 4\times10^{10}$ loop iterations and exceeds the time limit.
Its counterexample generator is:

\begin{codebox}
print(1)
print(200000)
for _ in range(200000):
    print("1 0 0")
\end{codebox}

Table~\ref{tab:accepted-tle-trace} condenses all seven turns. It preserves the
reasoning decision and tool result at each turn while omitting repeated source
code and problem text.

\begin{table}[H]
\centering
\small
\renewcommand{\arraystretch}{1.2}
\begin{tabularx}{\linewidth}{@{}>{\centering\arraybackslash}m{2.4em}X@{}}
\toprule
\textbf{Turn} & \textbf{Reasoning and observation} \\
\midrule
\ding{172} & Derives the geometric interpretation of the task, compiles the candidate,
and confirms that it passes the supplied examples. \\
\hline
\ding{173} & Checks the angular grouping and two-pointer count, then begins randomized
differential testing against a brute-force solver. \\
\hline
\ding{174} & Repairs an error in the brute-force harness. The process-per-case version
is too slow to complete the planned search. \\
\hline
\ding{175} & Batches 2,000 random instances into one candidate invocation. No wrong
answer is found, so the search shifts from functional correctness to runtime. \\
\hline
\ding{176} & Spots the $c^2$ loop after grouping and constructs one group of size
200,000. Running the candidate on this input exceeds the local timeout. \\
\hline
\ding{177} & Checks a small all-identical instance by enumeration and confirms that its
correct answer is zero. \\
\hline
\ding{178} & Returns \textit{Time Limit Exceeded}, explains the quadratic loop, and
submits the deterministic generator. UOJ validates the input and confirms the
time-limit failure. \\
\bottomrule
\end{tabularx}
\caption{A successful \textsc{Arbiter} trajectory for \textsc{USACO-Judge} submission 912773, identifying and triggering a \textit{Time Limit Exceeded} failure.}
\label{tab:accepted-tle-trace}
\end{table}

\paragraph{Wrong Answer from using a valid value as a sentinel.}
Submission 913195 for \emph{Min Max Subarrays II} (Contest 3 Platinum P3)
scored 100 on the official tests. The task asks for an array satisfying interval
minimum and maximum constraints. The candidate stores lower bounds in
\texttt{K\_min}, initialized
to zero, and later uses the same value to mean that no minimum constraint was
recorded:

\begin{codebox}
vector<long long> K_min(N + 1, 0);
...
if (K_min[j] != 0) {
    // record the minimum constraint covering position j
}
\end{codebox}

Zero is also a legal constraint value. For a one-element array with the
constraint $\min(a_1)=0$, the candidate drops the constraint and prints
\texttt{-1}, although $a_1=0$ is a valid construction. \textsc{Arbiter}
isolates this collision between data and sentinel state with a 14-byte input:

\begin{codebox}
print(1)
print("1 1")
print("1 1 1 0")
\end{codebox}

\begin{table}[H]
\centering
\small
\renewcommand{\arraystretch}{1.2}
\begin{tabularx}{\linewidth}{@{}>{\centering\arraybackslash}m{2.4em}X@{}}
\toprule
\textbf{Turn} & \textbf{Reasoning and observation} \\
\midrule
\ding{172} & Traces the DSU-based construction and notices that zero denotes both a
legal lower bound and the initial state. It compiles the candidate and observes
\texttt{-1} on the minimal instance. \\
\hline
\ding{173} & Connects the failure to the \texttt{K\_min[j] != 0} guard, then checks the
candidate on the first supplied examples. \\
\hline
\ding{174} & Runs the second supplied example and confirms that alternative printed
arrays still satisfy its constraints. \\
\hline
\ding{175} & Verifies that $N=Q=1$ and $k=0$ are legal, then checks directly that the
array $[0]$ satisfies the constraint. \\
\hline
\ding{176} & Replays the three-line input and again observes \texttt{-1}, fixing the
failure mechanism and expected output before submission. \\
\hline
\ding{177} & Returns \textit{Wrong Answer}, identifies the sentinel collision, and
submits the generator. UOJ validates the input and confirms the wrong answer. \\
\bottomrule
\end{tabularx}
\caption{A successful \textsc{Arbiter} trajectory for \textsc{USACO-Judge} submission 913195, identifying and triggering a \textit{Wrong Answer} failure.}
\label{tab:accepted-wa-trace}
\end{table}

%% file: sections/appendix/selection.tex
\section{Selection Details}
\label{app:selection}

\paragraph{Protocol.}
The candidate-generation, judging, and pool-size settings follow
\citet{agenticverifier2026}, but we independently regenerate the candidate
pools. Repeated uniform-selection runs closely recover the published pass@$1$
rates, checking aggregate candidate success; they do not establish identical
programs or matched selection difficulty. The baseline values in
Table~\ref{tab:selection} are from that study. An \textsc{Arbiter} call is a
complete agentic review and
may contain multiple sandbox tool invocations, so call counts are not matched
compute budgets across methods.
Input-based baselines receive 512 generated inputs per problem for pools of 8
and 64 candidates. \textsc{Arbiter} reviews every candidate independently,
using between one and eight calls per candidate. We count the
\emph{Accepted} verdicts for each candidate and sample uniformly among those
with the highest count. If no call returns \emph{Accepted}, selection falls
back to a uniform draw from the full pool. Candidate programs are written in
Python for LiveCodeBench and C++ for OJBench, following each benchmark's
convention. The prompt's \texttt{grading\_language} is \texttt{Python 3} for
LiveCodeBench and \texttt{C++17} for OJBench; \texttt{language} separately
specifies the code-fence tag. Our results for pools of 64 are averaged over
four independent repeats. Depending on the setting, the range across repeats is
0.79--1.25 pass@$1$ points.

\paragraph{Binary verdict quality.}
We evaluate the filtering decision directly through binary verdict accuracy.
Table~\ref{tab:selection-confusion} pools the complete evaluations for
candidate pools of 8 and 64.

\begin{table}[h]
\centering
\small
\renewcommand{\arraystretch}{1.12}
\setlength{\tabcolsep}{9pt}
\resizebox{\linewidth}{!}{%
\begin{tabular}{@{}>{\raggedright\arraybackslash}p{6.4cm} !{\color{gray!45}\vrule}
S[table-format=2.2] S[table-format=2.2]
!{\color{gray!45}\vrule}
S[table-format=2.2] S[table-format=2.2]@{}}
\toprule
& \multicolumn{2}{!{\color{gray!45}\vrule}c}{\textbf{Qwen3-30B-A3B-Thinking}}
& \multicolumn{2}{!{\color{gray!45}\vrule}c}{\textbf{Qwen3-235B-A22B-Thinking}} \\
\cmidrule(lr){2-3}\cmidrule(lr){4-5}
\textbf{Reference Verdict \textbackslash{} \textsc{Arbiter} Prediction}
& {Accepted} & {Unaccepted}
& {Accepted} & {Unaccepted} \\
\midrule
Accepted   & 88.81 & 11.19 & 88.28 & 11.72 \\
Unaccepted &  3.64 & 96.36 &  6.33 & 93.67 \\
\bottomrule
\end{tabular}%
}
\caption{Row-normalized binary verdict confusion matrices for code selection.
Rows are reference verdicts, columns are \textsc{Arbiter} predictions, and values are percentages.}
\label{tab:selection-confusion}
\end{table}

\textsc{Arbiter} accepts 88.81\% and 88.28\% of correct programs from the
30B and 235B solvers, respectively. It rejects 96.36\% and 93.67\% of
their unaccepted programs. Selection ranks candidates by their
\emph{Accepted} counts, so a false acceptance can place an incorrect program
in the top group. This occurs for 3.64\% of unaccepted 30B candidates and
6.33\% of unaccepted 235B candidates.

\paragraph{Call scaling.}
Table~\ref{tab:selection} reports one and eight calls per candidate, while
Figure~\ref{fig:attack-scaling} shows the intermediate budgets. One call
has higher reported pass@$1$ than the published Agentic Verifier baseline in
all eight settings. Calls two through eight add 0.9--3.4 pass@$1$ points. For pools of
64, OJBench continues to improve through the eighth call, while most of the
LiveCodeBench gain arrives within the first four calls. The smaller pools
plateau earlier and show minor fluctuations after the first few calls.

\input{sections/figures/selection_call_scaling}

%% file: sections/figures/selection_call_scaling.tex
\begin{figure}[t]
\centering
\begin{tikzpicture}
\definecolor{atkOurs}{HTML}{F08C00}%
\definecolor{atkRival}{HTML}{6B7280}%
\definecolor{atkFloor}{HTML}{93B8D6}%
\definecolor{atkCeil}{HTML}{1F4E79}%
\definecolor{atkBand}{HTML}{EDF1F5}%
\pgfplotsset{
  atkpanel/.style={
    width=2.95cm, height=1.86cm, scale only axis,
    xmin=0.55, xmax=8.45,
    xtick={1,2,3,4,5,6,7,8},
    tick align=outside, tick pos=left,
    major tick length=1.4pt,
    tick label style={font=\tiny, /pgf/number format/assume math mode=true},
    xticklabel style={yshift=1.6pt},
    yticklabel style={xshift=1.8pt, /pgf/number format/assume math mode=true,
      anchor=east, text width=1.05em, align=right},
    label style={font=\scriptsize},
    title style={font=\scriptsize, align=center, yshift=-8pt},
    ytick align=outside,
    grid=none,
    axis line style={black!55},
  },
}
\begin{groupplot}[
  group style={
    group name=atk,
    group size=4 by 2,
    horizontal sep=0.92cm,
    vertical sep=0.62cm,
  },
  atkpanel,
]

\nextgroupplot[title={\textcolor{black!60}{Qwen3-30B-A3B}},
               ylabel={pass@1 (\%)}, ymin=25, ymax=50, ytick={30,35,40,45}]
  \path[fill=atkBand, fill opacity=0.55] (axis cs:0.55,28.90) rectangle (axis cs:8.45,42.24);
  \addplot[atkCeil, line width=0.7pt] coordinates {(0.55,42.24) (8.45,42.24)};
  \addplot[atkFloor, line width=0.7pt] coordinates {(0.55,28.90) (8.45,28.90)};
  \addplot[atkRival, densely dashed, thick] coordinates {(1,33.5) (8,33.5)};
  \addplot[atkOurs, thick, mark=*, mark size=1.25pt, mark options={fill=white, draw=atkOurs, line width=0.6pt}] coordinates
    {(1,37.61) (2,38.70) (3,39.05) (4,39.66) (5,39.83) (6,40.21) (7,40.05) (8,40.02)};

\nextgroupplot[title={\textcolor{black!60}{Qwen3-235B-A22B}},
               ymin=35, ymax=60, ytick={40,45,50,55}]
  \path[fill=atkBand, fill opacity=0.55] (axis cs:0.55,37.10) rectangle (axis cs:8.45,55.60);
  \addplot[atkCeil, line width=0.7pt] coordinates {(0.55,55.60) (8.45,55.60)};
  \addplot[atkFloor, line width=0.7pt] coordinates {(0.55,37.10) (8.45,37.10)};
  \addplot[atkRival, densely dashed, thick] coordinates {(1,45.9) (8,45.9)};
  \addplot[atkOurs, thick, mark=*, mark size=1.25pt, mark options={fill=white, draw=atkOurs, line width=0.6pt}] coordinates
    {(1,50.04) (2,50.90) (3,51.63) (4,51.56) (5,51.30) (6,51.41) (7,51.73) (8,51.67)};

\nextgroupplot[title={\textcolor{black!60}{Qwen3-30B-A3B}},
               ymin=59, ymax=84, ytick={60,65,70,75,80}]
  \path[fill=atkBand, fill opacity=0.55] (axis cs:0.55,66.20) rectangle (axis cs:8.45,81.14);
  \addplot[atkCeil, line width=0.7pt] coordinates {(0.55,81.14) (8.45,81.14)};
  \addplot[atkFloor, line width=0.7pt] coordinates {(0.55,66.20) (8.45,66.20)};
  \addplot[atkRival, densely dashed, thick] coordinates {(1,73.6) (8,73.6)};
  \addplot[atkOurs, thick, mark=*, mark size=1.25pt, mark options={fill=white, draw=atkOurs, line width=0.6pt}] coordinates
    {(1,76.76) (2,77.53) (3,78.14) (4,78.29) (5,78.29) (6,79.14) (7,79.21) (8,79.50)};

\nextgroupplot[title={\textcolor{black!60}{Qwen3-235B-A22B}},
               ymin=68, ymax=93, ytick={70,75,80,85,90}]
  \path[fill=atkBand, fill opacity=0.55] (axis cs:0.55,74.70) rectangle (axis cs:8.45,88.00);
  \addplot[atkCeil, line width=0.7pt] coordinates {(0.55,88.00) (8.45,88.00)};
  \addplot[atkFloor, line width=0.7pt] coordinates {(0.55,74.70) (8.45,74.70)};
  \addplot[atkRival, densely dashed, thick] coordinates {(1,81.6) (8,81.6)};
  \addplot[atkOurs, thick, mark=*, mark size=1.25pt, mark options={fill=white, draw=atkOurs, line width=0.6pt}] coordinates
    {(1,85.00) (2,85.33) (3,85.71) (4,85.71) (5,85.71) (6,85.71) (7,85.93) (8,85.93)};

\nextgroupplot[ylabel={pass@1 (\%)}, ymin=28, ymax=53, ytick={30,35,40,45,50}]
  \path[fill=atkBand, fill opacity=0.55] (axis cs:0.55,28.90) rectangle (axis cs:8.45,50.00);
  \addplot[atkCeil, line width=0.7pt] coordinates {(0.55,50.00) (8.45,50.00)};
  \addplot[atkFloor, line width=0.7pt] coordinates {(0.55,28.90) (8.45,28.90)};
  \addplot[atkRival, densely dashed, thick] coordinates {(1,34.9) (8,34.9)};
  \addplot[atkOurs, thick, mark=*, mark size=1.25pt, mark options={fill=white, draw=atkOurs, line width=0.6pt}] coordinates
    {(1,43.60) (2,44.36) (3,44.74) (4,45.51) (5,45.68) (6,45.58) (7,45.75) (8,46.30)};

\nextgroupplot[ymin=35, ymax=65, ytick={40,45,50,55,60}]
  \path[fill=atkBand, fill opacity=0.55] (axis cs:0.55,37.10) rectangle (axis cs:8.45,62.07);
  \addplot[atkCeil, line width=0.7pt] coordinates {(0.55,62.07) (8.45,62.07)};
  \addplot[atkFloor, line width=0.7pt] coordinates {(0.55,37.10) (8.45,37.10)};
  \addplot[atkRival, densely dashed, thick] coordinates {(1,48.6) (8,48.6)};
  \addplot[atkOurs, thick, mark=*, mark size=1.25pt, mark options={fill=white, draw=atkOurs, line width=0.6pt}] coordinates
    {(1,53.30) (2,55.19) (3,54.57) (4,55.13) (5,55.24) (6,55.86) (7,56.07) (8,56.11)};

\nextgroupplot[ymin=62, ymax=88, ytick={65,70,75,80,85}]
  \path[fill=atkBand, fill opacity=0.55] (axis cs:0.55,66.20) rectangle (axis cs:8.45,86.29);
  \addplot[atkCeil, line width=0.7pt] coordinates {(0.55,86.29) (8.45,86.29)};
  \addplot[atkFloor, line width=0.7pt] coordinates {(0.55,66.20) (8.45,66.20)};
  \addplot[atkRival, densely dashed, thick] coordinates {(1,73.7) (8,73.7)};
  \addplot[atkOurs, thick, mark=*, mark size=1.25pt, mark options={fill=white, draw=atkOurs, line width=0.6pt}] coordinates
    {(1,82.60) (2,83.87) (3,84.00) (4,84.40) (5,84.43) (6,84.45) (7,84.55) (8,84.56)};

\nextgroupplot[ymin=72, ymax=95, ytick={75,80,85,90}]
  \path[fill=atkBand, fill opacity=0.55] (axis cs:0.55,74.70) rectangle (axis cs:8.45,92.57);
  \addplot[atkCeil, line width=0.7pt] coordinates {(0.55,92.57) (8.45,92.57)};
  \addplot[atkFloor, line width=0.7pt] coordinates {(0.55,74.70) (8.45,74.70)};
  \addplot[atkRival, densely dashed, thick] coordinates {(1,82.9) (8,82.9)};
  \addplot[atkOurs, thick, mark=*, mark size=1.25pt, mark options={fill=white, draw=atkOurs, line width=0.6pt}] coordinates
    {(1,87.70) (2,88.31) (3,88.93) (4,89.03) (5,89.10) (6,89.18) (7,89.26) (8,89.26)};

\end{groupplot}

\node[anchor=south, font=\small\bfseries, yshift=0.34cm]
  at ($(atk c1r1.north)!0.5!(atk c2r1.north)$) {OJBench};
\node[anchor=south, font=\small\bfseries, yshift=0.34cm]
  at ($(atk c3r1.north)!0.5!(atk c4r1.north)$) {LiveCodeBench};

\node[anchor=south, rotate=-90, font=\scriptsize\bfseries, xshift=0.30cm]
  at (atk c4r1.east) {Pool=8};
\node[anchor=south, rotate=-90, font=\scriptsize\bfseries, xshift=0.30cm]
  at (atk c4r2.east) {Pool=64};

\path let \p1=(atk c1r2.south west), \p2=(atk c4r2.south east) in
  coordinate (lgL) at (\x1,\y1) coordinate (lgR) at (\x2,\y1);
\node[anchor=north, yshift=-0.34cm, font=\scriptsize] at ($(lgL)!0.5!(lgR)$)
  {\textsc{Arbiter} calls per candidate};

\newcommand{\atksw}[1]{\tikz[baseline=-0.55ex]{\draw[#1] (0,0) -- (4.6mm,0);}}
\node[anchor=north, yshift=-0.62cm] at ($(lgL)!0.5!(lgR)$) {%
  \scriptsize
  \begin{tabular}{@{}c@{~}l@{\hspace{2.4mm}}c@{~}l@{\hspace{2.4mm}}c@{~}l@{\hspace{2.4mm}}c@{~}l@{}}
    \tikz[baseline=-0.55ex]{\draw[atkOurs, thick] (0,0) -- (4.6mm,0);
      \draw[atkOurs, fill=white, line width=0.6pt] (2.3mm,0) circle (0.52mm);}
      & \textsc{Arbiter} (Ours)
    & \atksw{atkRival, densely dashed, thick} & Agentic Verifier~\citep{agenticverifier2026}
    & \atksw{atkCeil, line width=0.7pt} & Perfect selection (Upper)
    & \atksw{atkFloor, line width=0.7pt} & Uniform selection (Lower) \\
  \end{tabular}};
\end{tikzpicture}
\caption{Pass@$1$ as \textsc{Arbiter} calls per candidate increase from one to
eight. Rows show candidate pools of 8 and 64. Uniform and perfect selection mark
the lower and upper reference levels.}
\label{fig:attack-scaling}
\end{figure}
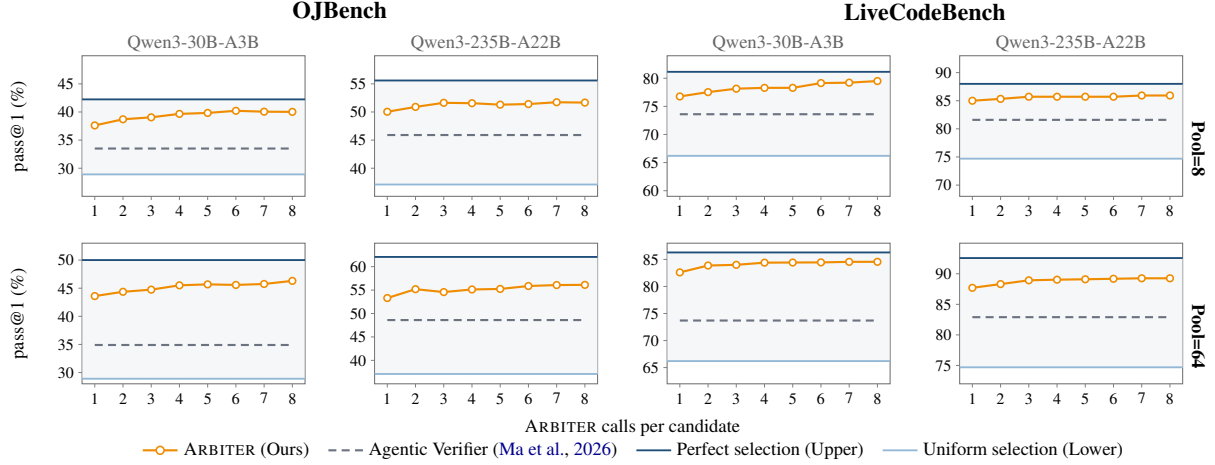

%% file: sections/appendix/refinement.tex
\section{Refinement and Diagnosis Analysis}
\label{app:refine}

\subsection{Submission Protocol}
\label{app:refine-detail}

Round $0$ is the shared, already judged starting program. It is excluded from
the budget of additional submissions. Every condition retains a 100-point
starting program without further submissions; only the Online Judge condition
receives its result as solver feedback. For trajectories still active under
Best@1, the previous-code condition refines without external feedback and
submits round $8$, while the Online Judge condition submits round $1$. Each
\textsc{Arbiter} condition reviews the shared round $0$ program and its own
later revisions, then retains
the first version labelled \emph{Accepted}, or round $8$ if none is. Under
Best@8, up to eight additional revisions are submitted; the Online Judge
condition stops at a 100-point score, and the highest score among additional
submissions is retained. These budgets count judge submissions, not solver
generations or \textsc{Arbiter} reviews. All results are macro-averaged over problems.

\subsection{IOI 2026 Protocol and Score Accounting}
\label{app:ioi-protocol}

\paragraph{Contest rules.}
IOI 2026 had six problems worth 100 points each and allowed at most 50
submissions per problem. A problem score sums the best score obtained on each
subtask, even when those scores come from different submissions. Four problems
require one program. \emph{Classroom Game} and \emph{Partition} require several,
and a submitted program cannot always identify its subtask at runtime. The 50
submissions for these two problems must therefore be divided among the required
programs.

\paragraph{Refinement protocol.}
GPT-OSS-120B generates 50 initial solutions for each refinement target. At every
round, \textsc{Arbiter} returns a verdict and feedback for each active trajectory.
A trajectory stops when \textsc{Arbiter} returns \emph{Accepted} and retains that
program. Otherwise, it runs for five refinement rounds and keeps the last
version. For the four single-program problems, we align retained programs by
trajectory index and combine the target-specific implementations at each index
into one task-level submission, yielding 50 budgeted candidates. For
\emph{Classroom Game} and \emph{Partition}, the separately required programs
instead share the same 50-slot per-problem budget. \emph{Classroom Game} uses
17, 17, and 16 slots for its three programs. \emph{Partition} uses 31 and 1
slots for its two programs, leaving 18 unused once no further score gain is
available.

One budgeted submission is one task-level program evaluated by the judge.
The 50 initial trajectories per refinement target are internal candidates.
Generation and \textsc{Arbiter} review use no
submission slots. The Online Judge condition starts ten trajectories per
problem, including the two problems requiring multiple programs, and allows
at most five judged submissions per trajectory. Each feedback query uses one
of these slots, so the condition uses at most $10\times5=50$ submissions per
problem. Both policies allow at most 50 judged submissions on each of the six
problems. The two curves in
Figure~\ref{fig:ioi-climb} report expected contest scores under this same cap
and the same per-subtask-maximum scoring rule, taking the expectation over
each policy's candidate draws. Inference compute is not matched.

Only seven trajectories remain active on average after early stopping. Each
refinement target uses fewer than 100 solver generations, with fewer than two
revision rounds on average. The expected score rises from 240.49 before
refinement to 356.91 after one round and 385.46 after round 4, above the gold
cutoff of 361.12. The expected score with online-judge feedback reaches 334.6
under the same submission limit.

\paragraph{Expected and realized scores.}
The 385.46-point total averages over draws without replacement from retained
\emph{Classroom Game} candidates under its fixed 17/17/16 allocation. Its
expected contribution is 30.46 points; the other five problems contribute
355.00. Separately, the selected programs scored
390.20 out of 600 in a QOJ\footnote{\url{https://qoj.ac/}} evaluation:
35.00 on \emph{Ball Machine}, 58.00 on \emph{Monuments}, 100.00 on \emph{Tiling
Game}, 35.20 on \emph{Classroom Game}, 100.00 on \emph{Magic City}, and 62.00 on
\emph{Partition}. The 4.74-point difference is entirely due to the realized
\emph{Classroom Game} score. To control inference cost for \emph{Magic City}, we retain
separate refinement targets for official subtasks 1--5, then group subtasks
6--10, 11--12, and 13--50 into three more targets. QOJ still scores all 50
subtasks separately.

\subsection{Binary Verdict Quality During Refinement}
\label{app:refine-verifier}

Three arms of Table~\ref{tab:refine} use \textsc{Arbiter} feedback. At Best@1,
each retains the first version predicted \emph{Accepted}, or round $8$ if none
is. We compare every predicted verdict with the corresponding online-judge result,
separately for the two solver models.

\begin{table}[h]
\centering
\small
\renewcommand{\arraystretch}{1.12}
\setlength{\tabcolsep}{9pt}
\resizebox{\linewidth}{!}{%
\begin{tabular}{@{}>{\raggedright\arraybackslash}p{6.4cm} !{\color{gray!45}\vrule}
S[table-format=2.2] S[table-format=2.2]
!{\color{gray!45}\vrule}
S[table-format=2.2] S[table-format=2.2]@{}}
\toprule
& \multicolumn{2}{!{\color{gray!45}\vrule}c}{\textbf{GPT-OSS-120B}}
& \multicolumn{2}{!{\color{gray!45}\vrule}c}{\textbf{Qwen3.6-35B-A3B}} \\
\cmidrule(lr){2-3}\cmidrule(lr){4-5}
\textbf{Reference Verdict \textbackslash{} \textsc{Arbiter} Prediction}
& {Accepted} & {Unaccepted}
& {Accepted} & {Unaccepted} \\
\midrule
Accepted   & 97.20 &  2.80 & 96.37 &  3.63 \\
Unaccepted &  9.68 & 90.32 &  2.65 & 97.35 \\
\bottomrule
\end{tabular}%
}
\caption{Row-normalized binary verdict confusion matrices during refinement.
Rows are reference verdicts, columns are \textsc{Arbiter} predictions, and values are percentages.}
\label{tab:refine-verifier}
\end{table}

The reference \emph{Accepted} rate differs between the two solvers, at 68.6\%
for GPT-OSS-120B and 46.2\% for Qwen3.6-35B-A3B. The main refinement results
use the programs selected under these predictions, without correcting false
acceptances or false rejections after evaluation.

\subsection{Diagnosis Quality and the Counterexample Proxy}
\label{app:diagnosis-proxy}

\paragraph{Audit protocol.}
Free-form diagnoses cannot be checked with the deterministic procedure used for
counterexamples, so we audit 400 \textsc{Arbiter} diagnoses of GPT-OSS-120B
candidates from the refinement experiment. We randomly sample 100 diagnoses
within each USACO division, without stratifying by counterexample outcome.
Three competitive programmers independently assess
each diagnosis against the problem statement, candidate program, official score
or compile error, and the
first failing official test when available. They assign one of three labels:
correct, incorrect, or vacuous, where vacuous means too generic to guide a repair.
Their initial labels agree on all 400 diagnoses. For a bug claim about an
officially \emph{Accepted} candidate, the auditors label it correct if a
validated counterexample exposes the claimed failure, and incorrect otherwise.
Counterexample replay is also recorded for the proxy analysis below.
The audit takes place after refinement
and does not affect any reported score.

\begin{table}[h]
\centering
\small
\renewcommand{\arraystretch}{1.05}
\setlength{\tabcolsep}{10pt}
\begin{tabular}{@{}l r r@{}}
\toprule
\textbf{Diagnosis audit} & \textbf{Count} & \textbf{Share} \\
\midrule
Correct & 332 & 83.0\% \\
Incorrect & 65 & 16.2\% \\
Vacuous & 3 & 0.8\% \\
\bottomrule
\end{tabular}
\caption{Human labels for the division-balanced random audit. Shares are computed
within the audit sample.}
\label{tab:diagnosis-audit}
\end{table}

Of the 332 correct diagnoses, 169 concern programs judged \emph{Accepted},
while 163 identify a bug in an \emph{unaccepted} program. Among the
65 incorrect diagnoses, 37 claim a bug in a candidate awarded full official
credit without a validated counterexample substantiating the claim, and 21
reason about runtime behavior in code that does not compile.
Four name the wrong failure type, and three mark an \emph{unaccepted} candidate
\emph{Accepted}. The three vacuous diagnoses merely restate the counterexample.
Table~\ref{tab:diagnosis-depth} measures how much repair guidance appears in the
163 correct diagnoses of a bug.

\begin{table}[H]
\centering
\renewcommand{\arraystretch}{1.05}
\setlength{\tabcolsep}{6pt}
\begin{tabularx}{\linewidth}{@{}p{0.20\linewidth} r r >{\raggedright\arraybackslash}X@{}}
\toprule
\textbf{Depth} & \textbf{Count} & \textbf{Share} & \textbf{Operational rule} \\
\midrule
Actionable fix & 88 & 54.0\% & Cites candidate code and states the correct behavior. \\
Located cause & 46 & 28.2\% & Cites candidate code and explains the failure, but gives no fix. \\
Mechanism only & 17 & 10.4\% & Explains the failure without locating it in the code. \\
Symptom only & 12 & 7.4\% & Reports the failure without locating or explaining it. \\
\bottomrule
\end{tabularx}
\caption{Depth of the 163 correct diagnoses that identify a bug. Each level
follows the operational rule shown in the table.}
\label{tab:diagnosis-depth}
\end{table}

Most correct bug diagnoses contain concrete repair information. Of the 163
outputs, 134 cite a specific identifier, expression, or branch in the candidate
program, and 88 also state the behavior needed to fix it. Only 12 stop at the
observed symptom.

\paragraph{Representative cases.}
Table~\ref{tab:diagnosis-cases} gives one audited example for each depth level.
Every counterexample in the table passes the validator and breaks its candidate.
The diagnoses are shortened for space without changing their technical claims.

\begin{table}[H]
\centering
\small
\renewcommand{\arraystretch}{1.12}
\setlength{\tabcolsep}{5pt}
\resizebox{\linewidth}{!}{%
\begin{tabular}{@{}p{2.6cm}p{13cm}@{}}
\toprule
\textbf{Depth and CPID} & \textbf{Audited case} \\
\midrule
\textbf{Actionable fix}
\newline \href{https://usaco.org/index.php?page=viewproblem2\&cpid=1545}{CPID 1545}
& On the generated five-node instance, cow 2 hosts a C party on the first
night. The diagnosis identifies that \texttt{range\_chmax} updates the queried
depth before \texttt{oldOwner} is recovered, so the program mistakes the
updated node for the previous owner and prints \mbox{\texttt{0 0 0}} instead of
\mbox{\texttt{2 0 0}}. It gives the precise repair: call \texttt{point\_query} before
\texttt{range\_chmax}. \\
\addlinespace[2pt]
\textbf{Located cause}
\newline \href{https://usaco.org/index.php?page=viewproblem2\&cpid=1545}{CPID 1545}
& The generated instance is a two-node cycle. Cow 2 first hosts C and cow 1
then hosts O. The diagnosis traces the failure to the second cycle host: when
\texttt{prevIdx+1 >= Lsize}, \texttt{process\_interval} reads past the cycle
boundary. The candidate consequently prints \mbox{\texttt{0 2 0}} instead of
\mbox{\texttt{1 1 0}}. The diagnosis locates the failing condition and explains its
effect, but does not prescribe a repair. \\
\addlinespace[2pt]
\textbf{Mechanism only}
\newline \href{https://usaco.org/index.php?page=viewproblem2\&cpid=1590}{CPID 1590}
& For cyclic costs \texttt{[7,10,2]} and a time budget of 34, the diagnosis
explains that the candidate incorrectly buys later wins in cheapest-first
order. The fixed queue requires cost 36 for the third win, whereas the
candidate assumes cost 28 and reports three wins instead of two. This explains
the failure mechanism and the correct behavior without pointing to a particular
expression or branch. \\
\addlinespace[2pt]
\textbf{Symptom only}
\newline \href{https://usaco.org/index.php?page=viewproblem2\&cpid=1573}{CPID 1573}
& On a two-position instance with equal input probabilities, the diagnosis
reports that the candidate assigns probability \texttt{3/4} to each position
instead of \texttt{1/2}, producing a total of \texttt{3/2}. It only speculates
about the DP sweep and residue contributions, without identifying a responsible
operation or establishing a causal account of the failure. \\
\bottomrule
\end{tabular}%
}
\caption{Representative audited diagnoses at each depth. Each generated input
is accepted by the validator and breaks the corresponding candidate.}
\label{tab:diagnosis-cases}
\end{table}

\paragraph{Counterexample success tracks diagnosis quality.}
We evaluate the proxy among audited outputs that report a counterexample and
exclude candidates that do not compile, since no generated input can execute
against them. A counterexample breaks the candidate for 71.6\% of correct
diagnoses, compared with 10.3\% of incorrect or vacuous diagnoses. The clear
separation supports counterexample success as a mechanically verifiable signal
of diagnosis quality.

\paragraph{Scope.}
Counterexample replay covers both solver models, while the diagnosis audit
covers GPT-OSS-120B. We therefore treat the diagnosis results as evidence for
this refinement setting rather than as a cross-solver estimate. We restrict the proxy analysis to compilable candidates to prevent
compile errors from inflating the association. No generated input can break a program
that does not compile.

\subsection{Diagnosis for Refinement and Counterexamples for Verification}
\label{app:refine-x}

Table~\ref{tab:refine} separates the three fields exposed to the refinement
model: verdict $\mathbf{V}$, diagnosis $\mathbf{D}$, and counterexample-generator source $\mathbf{X}$. $\mathbf{V}+\mathbf{D}$ and
$\mathbf{V}+\mathbf{D}+\mathbf{X}$ perform similarly across the twelve solver, difficulty, and
submission-budget settings. $\mathbf{V}+\mathbf{D}$ scores higher in nine, while $\mathbf{V}+\mathbf{D}+\mathbf{X}$ leads
in the other three by 0.6--1.6 points.

\paragraph{Counterexample availability.}
$\mathbf{X}$ is requested only when \textsc{Arbiter} predicts a failure that can be exposed
by an input. The field is non-empty in 27.4\% of GPT-OSS-120B reviews and 50.8\% of
Qwen3.6-35B-A3B reviews. When $\mathbf{X}$ is empty, omitting it adds no
feedback content within that review. Reviews can differ once the conditions'
revision trajectories diverge.

\paragraph{Diagnosis as repair guidance.}
The audit above shows that $\mathbf{D}$ usually contains concrete information for repair.
Among the 163 correct diagnoses of a bug, 134 (82.2\%) point to specific
candidate code and 88 (54.0\%) state the behavior needed to fix it. The
downstream ablation
agrees with this finding: $\mathbf{V}+\mathbf{D}$ outperforms $\mathbf{V}+\mathbf{X}$ in ten of twelve settings.
On the Hard split, $\mathbf{D}$ improves over $\mathbf{X}$ by 5.9 and 6.7 points for
Qwen3.6-35B-A3B and by 6.5 and 3.3 points for GPT-OSS-120B under Best@1
and Best@8, respectively.

In the refinement study, $\mathbf{D}$ provides code-level repair guidance, while
adding counterexample-generator source $\mathbf{X}$ to $\mathbf{V}+\mathbf{D}$ shows no
consistent additional benefit. Separately, success of replayed counterexamples
serves as a mechanically checkable proxy for diagnosis quality in verifier
training and offline evaluation.

%% file: sections/appendix/training_ablations.tex
\section{Training Ablations and Length Diagnostics}
\label{app:ablation}

\subsection{Verdict-Only Rewards Encourage Majority-Class Guessing}
\label{app:ablation-replay}

We train a diagnostic run in which reward depends only on agreement between the
predicted and reference binary verdict. Training reward rises,
while validation verdict accuracy remains between 56\% and 58\%. Over the same
run, mean tool calls fall from 4.9 to below one, and the share of rollouts that
use no tools rises from 45\% to above 70\%. The model learns to improve its
reward without becoming a better verifier.

The training pool is 85\% \emph{unaccepted}. A policy that always guesses that
verdict would therefore be correct on 85\% of this pool without checking any
candidate. The rising training reward, flat validation accuracy, and declining
tool use are consistent with this majority-class shortcut. In the full reward,
most credit for a rejection other than \emph{Compile Error} instead requires a
legal input that breaks the candidate.

\subsection{False-Positive Penalty}
\label{app:ablation-shaping}

We compare two otherwise matched runs with $\lambda=0$ and $\lambda=0.5$,
where we subtract $\lambda$ from the reward for a false-positive
\emph{Accepted} prediction.
Figure~\ref{fig:label-only-trap} (right) follows precision and recall during
training and on the validation set. Table~\ref{tab:false-accept-ablation}
evaluates the same step-80 checkpoints on the easy and hard splits of
\textsc{USACO-Judge}.

\begin{table}[h]
\centering
\small
\renewcommand{\arraystretch}{1.15}
\setlength{\tabcolsep}{9pt}
\begin{tabular}{@{}l r r r r@{}}
\toprule
& \multicolumn{2}{c}{Easy} & \multicolumn{2}{c}{Hard} \\
\cmidrule(lr){2-3}\cmidrule(lr){4-5}
$\lambda$ & \textbf{Prec.} & \textbf{Rec.} & \textbf{Prec.} & \textbf{Rec.} \\
\midrule
0   & 97.56  & 88.89 & 77.78 & 76.67 \\
0.5 & 100.00 & 83.33 & 80.98 & 62.86 \\
\bottomrule
\end{tabular}
\caption{\emph{Accepted}-class precision and recall on \textsc{USACO-Judge}
for the two $\lambda$ settings. All values are percentages.}
\label{tab:false-accept-ablation}
\end{table}

The larger penalty raises precision on both splits and lowers recall. On hard
problems, precision rises from 77.78\% to 80.98\%, while recall falls from
76.67\% to 62.86\%. The candidate-count-weighted mean of per-split precision
rises from 83.7\% to 86.7\%, while the corresponding mean of per-split recall
falls from 80.3\% to 69.0\%. These are averages of split-level rates, not metrics
pooled over all predictions. The matched runs establish the direction of the
trade-off. We do not treat either setting as universally optimal.

%% file: sections/appendix/length_diagnostics.tex
\subsection{Length Diagnostics}
\label{app:length-penalty}

This appendix gives the evidence behind the alternating length control used in
training. Its schedule and hyperparameters are reported in
Appendix~\ref{app:training}.

\paragraph{Why token length matters.}
Without length control, median response length grows from 22.6k to 29.5k tokens
over 50 training steps, while the abort rate rises from 13.6\% to 26.6\%.
Nearly all aborts are wall-clock timeouts. The median number of interaction
turns remains nine, showing that the growth occurs within turns and would not be
addressed by a turn limit alone.

\paragraph{Continuous penalty comparison.}
With the upper-quartile length penalty active at every training step,
median response length falls from 25.9k to 17.2k and then to 12.9k tokens. Mean
reward and hack success peak at 0.777 and 62.3\% when the median length is 17.2k,
then decline to 0.755 and 59.1\% at 12.9k. Over the same interval, the abort rate
falls to 1.1\% and median interaction turns drop from 10 to 7. Once timeout
waste has largely been removed, further compression coincides with weaker
verification. We therefore alternate penalty-on phases with phases that restore
the original reward, preserving room for longer reasoning when it is useful.